%% file: main.tex
\documentclass[table]{stellaredge}

\usepackage{times}          
\usepackage[utf8]{inputenc}
\usepackage{amsmath,amssymb,amsfonts}
\usepackage{graphicx}
\usepackage{booktabs}
\usepackage{pifont}          
\usepackage{xspace}
\usepackage{needspace}
\usepackage{longtable}
\usepackage{fontawesome5}
\input{sec5_macros}       
\input{toc}               

\crefname{figure}{Fig.}{Figs.}\Crefname{figure}{Fig.}{Figs.}
\crefname{table}{Tab.}{Tabs.}\Crefname{table}{Tab.}{Tabs.}
\crefname{section}{Sec.}{Secs.}\Crefname{section}{Sec.}{Secs.}
\crefname{appendix}{App.}{Apps.}\Crefname{appendix}{App.}{Apps.}


\title{%
  \raisebox{-0.20\height}{\includegraphics[height=1.6em]{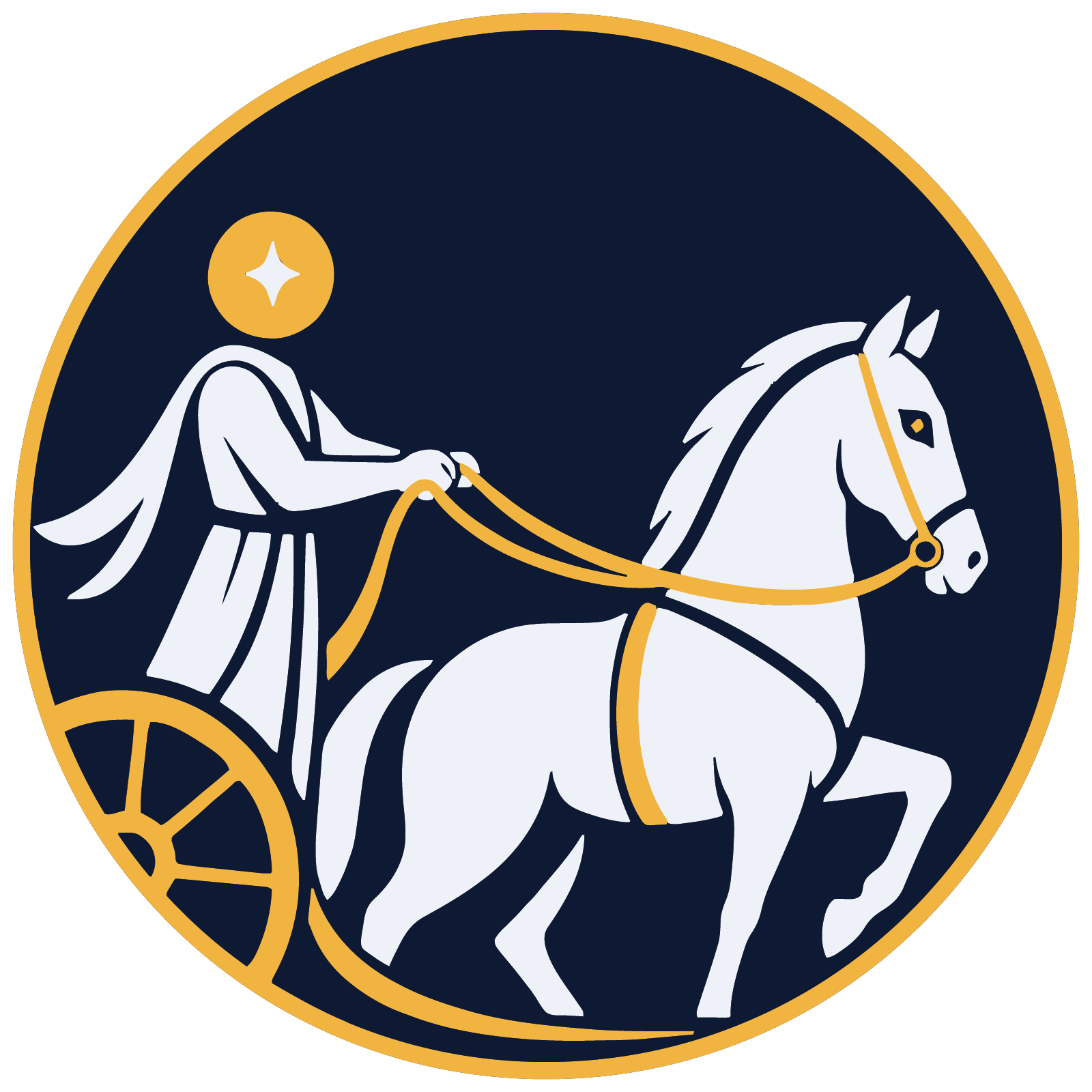}}\hspace{0.32em}%
  Auriga Technical Report: A Task-Agnostic Agent Harness for Zero-Shot Embodied Manipulation}

\affiliation{StellarEdge AI Technical Team}

\date{Nov. 1, 2026}
\metadata{%
  \href{https://auriga.stelledge.com/}{\faGlobe\ Project Page}
  \quad\textbar\quad
  \href{https://github.com/lose4578/Auriga}{\faGithub\ GitHub}%
}
\abstract{%
Vision-language models can interpret instructions and plan manipulation tasks, yet struggle when acting directly as robot policies.
On RoboDojo, a large \emph{planning--execution gap} separates high-level task decisions from physical execution: removing arm execution raises GPT-6 Astra's success rate from 22.5\% under the reference control interface to 76.2\%.
We present \textbf{Auriga}, a task-agnostic agent harness that closes 68.3\% of this gap in zero-shot manipulation without target-task trials or resets.
In Auriga, the model makes all task decisions over an explicit task belief and acts through either direct tool calls or synthesized multi-step Python programs, while the harness supplies calibrated 3D measurements, finger-referenced dual-arm control, and a single admission gateway to the robot.
Before motion, the model inspects a visual trajectory preview while simulation is paused; during execution, the harness checks declared stage tolerances and grasp bounds, stopping the robot on divergence and returning the measured discrepancy for in-place recovery.
Developed across 13 cross-task capability iterations and frozen for evaluation, Auriga ranks first on the RoboDojo leaderboard with a 57.7\% success rate (64.0 Score), more than doubling the reference harness on the same model, and remains first under scene randomization while learned policies drop by up to 33.9\%.
On RoboCasa365, the same frozen harness reaches a 73.2\% success rate, outperforming VLA-assisted agents by up to 23.1\% on composite tasks.
}

\begin{document}
\maketitle

\begin{figure}[t]
  \centering
  \includegraphics[width=\linewidth]{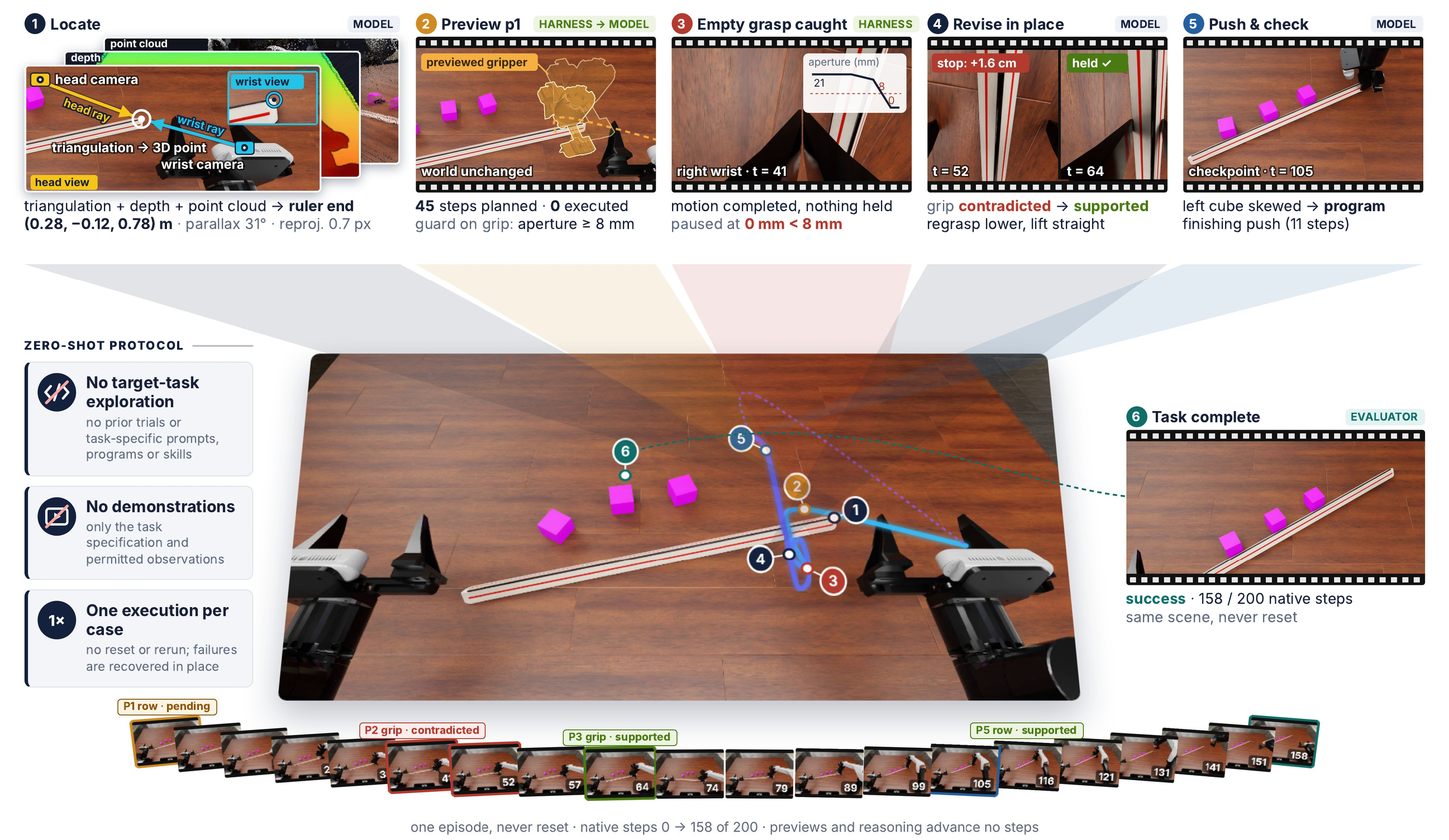}
  \caption{\textbf{Auriga completes manipulation tasks zero-shot within a single episode.} A general-purpose model plans and acts through a shared harness, with no target-task exploration, demonstrations, or resets. The model localizes targets by multi-view triangulation, complemented by depth and point-cloud estimates (1), and reviews candidate motions before they are executed (2). The harness verifies each execution and stops when an action fails (3); the model then revises its plan and recovers from the current state (4--5) until the task succeeds (6).}
  \label{fig:teaser}
\end{figure}

\input{introduction}
\input{problem_setting}

\input{sec_rsi}
\input{methodology}
\input{sec5_execution}    
\input{sec6_augment}      
\input{experiments}
\input{related_work}
\input{conclusion}
\clearpage
\bibliographystyle{plainnat}
\bibliography{references}

\input{appendix}

\end{document}

%% file: sec5_macros.tex
\providecommand{\method}{Auriga\xspace}
\providecommand{\code}[1]{\texttt{\detokenize{#1}}}



%% file: toc.tex
\makeatletter
\appto\appendix{%
  \phantomsection
  \addcontentsline{toc}{section}{Appendix}%
  \addtocontents{toc}{\protect\let\protect\l@section\protect\l@subsection}%
}
\makeatother

%% file: introduction.tex
\section{Introduction}

\begin{quote}
\small\itshape\color{seforeground}
``Plans are worthless, but planning is everything.'' \upshape{--- Dwight D.\ Eisenhower}
\end{quote}
\vspace{-0.2em}

In physical manipulation, a failed attempt changes the scene, often irreversibly.
Unlike simulated settings where an agent can call \code{env.reset} repeatedly on a target task to tune coordinate offsets, control scripts, or failure notes before evaluation~\citep{hexashell2026,embodiedevo2026,physicalrsi2026,sear2026}, a deployed robot must complete an unseen task within a single episode.
Tuning procedures across repeated trials on a fixed scene also fits to that scene's layout, and performance drops when object positions are randomized or the task changes.
Generalization in manipulation therefore separates two problems: across tasks during development, using failures to improve the shared perception, control, and verification tools rather than task-specific scripts; and during execution on an unseen task, completing the task in one episode without prior trials, demonstrations, or resets (\cref{sec:problem_setting}).

On a 210-episode diagnostic across all 42 tasks of the RoboDojo benchmark~\citep{robodojo2026}, the main bottleneck is physical execution rather than task reasoning (\cref{sec:gap,tab:gap}).
When GPT-6 Astra is given a privileged action that moves objects along chosen waypoints without the robot arm, it succeeds in \textbf{76.2\%} of episodes; when it drives the arm through the benchmark's reference end-effector interface (RoboProbe~\citep{unexpected2026}), it succeeds in \textbf{22.5\%}.
The model usually identifies the right object and destination, but the arm misses a grasp by a few millimeters, loses the object in flight, or hits a container rim during transport.
Common control interfaces do not catch these errors in time.
Step-wise and chunked action interfaces~\citep{pi052025,xvla2025,unexpected2026} issue motions from the latest frames without showing the full route in advance, so collisions appear only after the arm has moved.
Program generation~\citep{liang2023code,capx2026} hands execution to a script that cannot see the scene, where normal program exit does not mean the intended contact succeeded.

\begin{figure}[!htb]
  \centering
  \includegraphics[width=\linewidth]{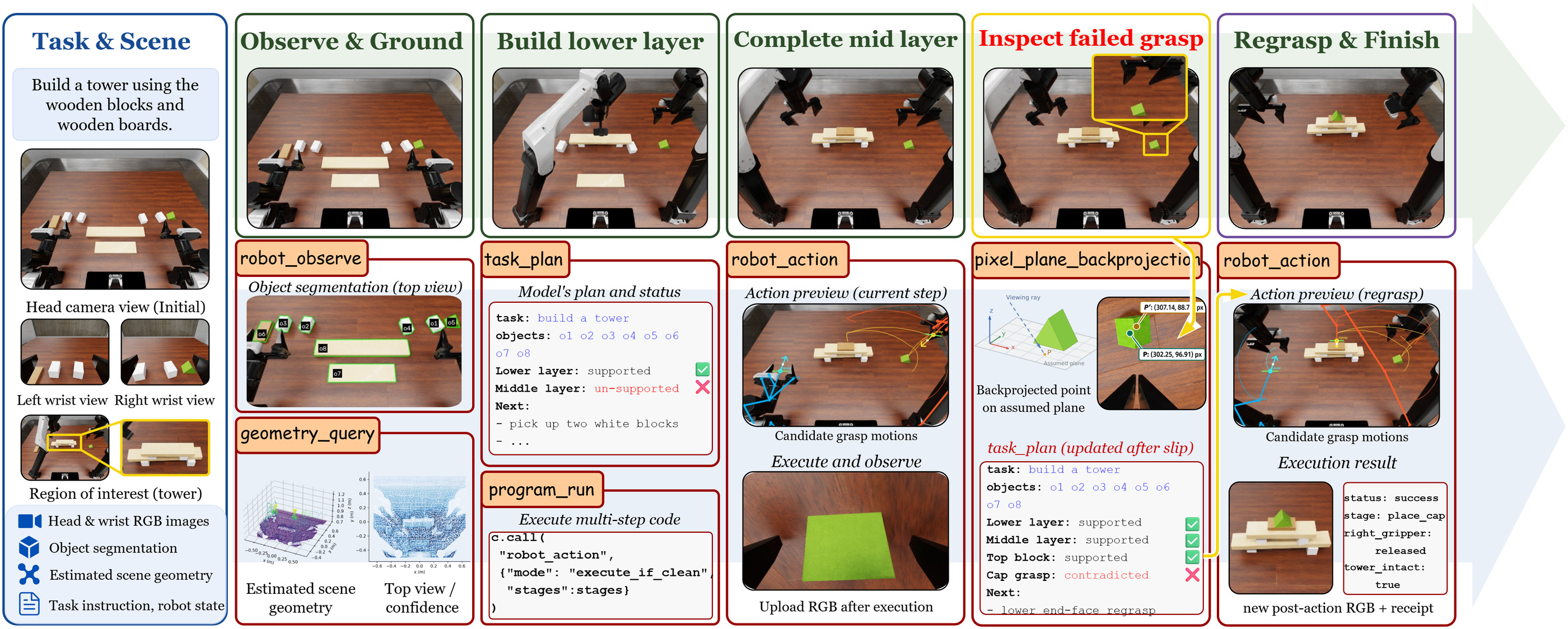}
  \caption{\textbf{Planning, execution, and recovery within a single tower-building episode.} From left to right, the model grounds the task in visual and geometric observations, constructs a plan, and builds the tower through direct tool calls and generated programs. Action previews and execution feedback support a preview--execute--verify loop. After a failed grasp, the model inspects the scene, updates its plan, and regrasps the object to complete the tower. The harness provides measurements and execution checks; the model selects actions and recovery strategies. The scene is never reset.}
  \label{fig:auriga_overview}
\end{figure}

We present \textbf{Auriga}, a task-agnostic agent harness that narrows this planning--execution gap within a single zero-shot episode (\cref{fig:teaser,fig:auriga_overview}).
In Auriga, the model makes all task decisions, while the harness measures the scene, checks and executes motions, and reports what happened:
\begin{enumerate}
    \item \textbf{Task-agnostic harness and explicit task belief (\cref{sec:method}).}
    Auriga contains no task-specific code, prompts, or heuristics, and exposes no privileged simulator state.
    One persistent model session plans the task, chooses actions, and recovers from errors.
    As an episode grows long, the model records its current plan, open questions, and task premises (\code{pending}, \code{supported}, \code{contradicted}, or \code{unknown}) in a concise \emph{task belief} $B_t$, while full interaction logs stay in an off-context \emph{evidence record} $E_t$ retrieved on demand.
    \item \textbf{Preview and verification through a single admission gateway (\cref{sec:wysiwye}).}
    Every physical motion, whether called directly or from a model-written program, passes through the admission gateway $\mathcal{A}$ in a \emph{preview--execute--verify} loop.
    While simulation is paused, $\mathcal{A}$ plans a multi-stage dual-arm sequence into a trajectory preview showing the path, step cost, and clearance warnings.
    During execution, the previewed trajectory runs while $\mathcal{A}$ checks model-set arrival tolerances $\tau_k = (\delta_p, \delta_\theta)$, gripper openings, and observation checkpoints.
    If a stage misses its tolerance, $\mathcal{A}$ stops the robot and returns a receipt $r_t$ with the completed stages, measured deviations, unexecuted stages, and fresh camera views so the model can recover from the current state (\cref{fig:auriga_overview}).
    \item \textbf{3D geometric measurements and finger-referenced actions (\cref{sec:augment}).}
    On the input side, camera calibration, VGGT-Omega~\citep{wang2026vggt} depth, and SAM~3.1~\citep{carion2026sam} segmentation provide 3D object estimates, ray--plane intersection with height sensitivity, and two-view triangulation, together with in-hand slip measurements and warnings when a motion re-enters a past failure region.
    On the output side, the model sets goals at the finger contact region rather than the wrist, chooses tolerances per stage, and coordinates both arms on a shared timeline.
\end{enumerate}

Auriga separates cross-task tool development from single-episode execution (\cref{sec:rsi_positioning,sec:exp-rsi}).
Over 13 development iterations (\cref{fig:harness_development}), manipulation failures were used to identify missing capabilities in 3D localization, contact checking, and state tracking, raising zero-shot performance across all 42 RoboDojo tasks from 36.2\% SR (45.5 Score) at iteration~\texttt{\#4} to 57.7\% SR (64.0 Score) in the final harness.
During evaluation, the harness is frozen and runs under the single-episode zero-shot protocol (\cref{sec:problem_setting,sec:exp-setup,sec:exp-main}), with no target-task exploration, demonstrations, or resets:
\begin{itemize}
    \item \textbf{RoboDojo~\citep{robodojo2026}.} Across all 42 tasks and five dimensions (6,300 episodes over three seeds), Auriga with GPT-6 Astra reaches an average \textbf{Score of 64.0} and \textbf{SR of 57.7\%}, ranking \textbf{first overall on the leaderboard} across learned policies and agent methods, more than doubling the GPT-6 Astra reference harness (\textbf{29.0 Score}, \textbf{22.5\% SR}), and leading all 12 columns of the agent block (\cref{tab:main}). On the 210-episode diagnostic set, Auriga closes \textbf{68.3\%} of the gap between reference arm control and arm-free object motion (\cref{tab:gap}).
    \item \textbf{Scene randomization.} Across the 12 Generalization tasks (\cref{fig:gen,tab:gen}), learned policies drop by up to 33.9\% in SR on randomized layouts, whereas \textbf{Auriga ranks first on both standard scenes (52.3\% SR) and random scenes (45.6\% SR)} and achieves the highest generalization score ($G = 46.3$).
    \item \textbf{RoboCasa365~\citep{robocasa3652026}.} On the 50 Target50 mobile manipulation tasks in unseen kitchens (250 episodes), Auriga reaches \textbf{73.2\% Overall SR} (\textbf{92.2\%} Atomic-Seen, \textbf{67.5\%} Composite-Seen, \textbf{57.5\%} Composite-Unseen) without a learned policy or cross-episode memory, outperforming VLA-assisted agents RPent~\citep{harnessvla2026} (\textbf{59.2\%}) and PyRUA-Lean~\citep{pyrua2026} (\textbf{56.4\%}) by \textbf{19.4\%} and \textbf{23.1\%} on composite tasks (\cref{tab:robocasa-main}).
\end{itemize}

%% file: problem_setting.tex
\section{Problem Setting and Positioning}

We study zero-shot manipulation, in which a general-purpose model uses a shared robot harness to construct and execute a task procedure from the task specification and interaction feedback.
We first define the single-episode zero-shot evaluation protocol (\cref{sec:problem_setting}), then quantify the planning--execution gap that motivates our harness (\cref{sec:gap}), and finally position cross-task harness evolution relative to within-task trial search (\cref{sec:rsi_positioning}).

\subsection{Zero-Shot Single-Episode Protocol}
\label{sec:problem_setting}

\paragraph{Task and interaction.}
The unit of evaluation is a \emph{case}, defined by a task instruction and a specified scene variant, random seed, and layout.
Each case starts from an initial scene and is evaluated under the benchmark's execution budget and success criteria.
The agent receives the task specification and accesses observations and robot actions through the shared harness.
It uses the available observations and interaction history to choose subsequent tool calls or programs, then receives feedback from their execution.
This interaction continues until success, termination, or budget exhaustion.
The benchmark's evaluator determines task success from the resulting execution.

\paragraph{Zero-shot protocol.}
We define zero-shot evaluation through three conditions:
\begin{enumerate}
    \item \textbf{No target-task exploration or development before evaluation.}
    Evaluation starts with a pretrained model, a fixed shared harness, and task-independent operating guidance.
    No preparatory exploration or trials are conducted on the target task, including its other layouts, to develop or select task-specific prompts, programs, skills, or policies.
    Model parameters remain fixed throughout evaluation.
    \item \textbf{No demonstrations at evaluation time.}
    The agent receives the task specification and permitted observations, without additional solution examples or expert demonstrations.
    For memory tasks, the benchmark-defined event sequence forms part of the case's observation input.
    \item \textbf{One execution per case.}
    Each case is initialized and executed once, without a reset or rerun.
    Its outcome is recorded whether the episode succeeds or fails.
\end{enumerate}

\paragraph{Within-episode adaptation.}
An episode comprises successive model decisions, observations, and physical actions.
During this interaction, the agent may inspect the scene, write and revise programs, preview motions, revisit its history, and recover from a failed grasp or placement.
Recovery continues from the physical state left by earlier actions under the remaining execution budget, rather than resetting the scene to start over.
Plans, generated code, and interaction memory evolve within the current episode and are not transferred to subsequent evaluation cases.
\Cref{tab:evaluation_conditions} compares six aspects of target-task preparation and information available to the test agent across representative manipulation settings (scope and per-setting qualifications are detailed in \cref{app:protocol}).

\begin{table}[htbp]
    \centering
    \footnotesize
    \setlength{\tabcolsep}{2.5pt}
    \renewcommand{\arraystretch}{1.16}
    \caption{Target-task preparation and evaluation inputs across the specified settings.}
    \label{tab:evaluation_conditions}
    \begin{tabular*}{\linewidth}{@{\extracolsep{\fill}}lcccccc@{}}
        \toprule
        & \multicolumn{2}{c}{\textbf{Before evaluation}} & \multicolumn{4}{c}{\textbf{Available to the test agent}} \\
        \cmidrule(lr){2-3}\cmidrule(l){4-7}
        \textbf{Method / setting} & \shortstack{Task\\exploration} & \shortstack{Cross-trial\\refinement} & \shortstack{Task code\\/ skills} & \shortstack{Trial-derived\\memory} & \shortstack{Demo-derived\\artifacts} & \shortstack{Demo\\references} \\
        \midrule
        GPT-6 Astra (zero-shot)~\citep{unexpected2026} & \textcolor[HTML]{287A50}{\ding{55}} & \textcolor[HTML]{287A50}{\ding{55}} & \textcolor[HTML]{287A50}{\ding{55}} & \textcolor[HTML]{287A50}{\ding{55}} & \textcolor[HTML]{287A50}{\ding{55}} & \textcolor[HTML]{287A50}{\ding{55}} \\
        GPT-6 Astra (image-demo ICL)~\citep{unexpected2026} & \textcolor[HTML]{287A50}{\ding{55}} & \textcolor[HTML]{287A50}{\ding{55}} & \textcolor[HTML]{287A50}{\ding{55}} & \textcolor[HTML]{287A50}{\ding{55}} & \textcolor[HTML]{287A50}{\ding{55}} & \textcolor[HTML]{A83232}{\ding{51}} \\
        GPT-Policy (video ICL)~\citep{gptpolicy2026} & \textcolor[HTML]{287A50}{\ding{55}} & \textcolor[HTML]{287A50}{\ding{55}} & \textcolor[HTML]{287A50}{\ding{55}} & \textcolor[HTML]{287A50}{\ding{55}} & \textcolor[HTML]{287A50}{\ding{55}} & \textcolor[HTML]{A83232}{\ding{51}} \\
        GPT-GEAR (full-context ICL)~\citep{gptgear2026} & \textcolor[HTML]{287A50}{\ding{55}} & \textcolor[HTML]{287A50}{\ding{55}} & \textcolor[HTML]{287A50}{\ding{55}} & \textcolor[HTML]{287A50}{\ding{55}} & \textcolor[HTML]{A83232}{\ding{51}} & \textcolor[HTML]{A83232}{\ding{51}} \\
        SenseNova (RoboDojo ICL)~\citep{sensenova2026embodied} & \textcolor[HTML]{287A50}{\ding{55}} & \textcolor[HTML]{287A50}{\ding{55}} & \textcolor[HTML]{A83232}{\ding{51}} & \textcolor[HTML]{287A50}{\ding{55}} & \textcolor[HTML]{A83232}{\ding{51}} & \textcolor[HTML]{A83232}{\ding{51}} \\
        PyRUA-Lean (RoboDojo)~\citep{pyrua2026} & \textcolor[HTML]{888888}{\raisebox{0.45ex}{\rule{0.45em}{0.65pt}}} & \textcolor[HTML]{888888}{\raisebox{0.45ex}{\rule{0.45em}{0.65pt}}} & \textcolor[HTML]{287A50}{\ding{55}} & \textcolor[HTML]{287A50}{\ding{55}} & \textcolor[HTML]{287A50}{\ding{55}} & \textcolor[HTML]{287A50}{\ding{55}} \\
        SEAR (hardware AaP)~\citep{sear2026} & \textcolor[HTML]{A83232}{\ding{51}} & \textcolor[HTML]{A83232}{\ding{51}} & \textcolor[HTML]{A83232}{\ding{51}} & \textcolor[HTML]{A83232}{\ding{51}} & \textcolor[HTML]{287A50}{\ding{55}} & \textcolor[HTML]{287A50}{\ding{55}} \\
        HexaShell (RoboDojo)~\citep{hexashell2026} & \textcolor[HTML]{A83232}{\ding{51}} & \textcolor[HTML]{A83232}{\ding{51}} & \textcolor[HTML]{A83232}{\ding{51}} & \textcolor[HTML]{A83232}{\ding{51}} & \textcolor[HTML]{287A50}{\ding{55}} & \textcolor[HTML]{287A50}{\ding{55}} \\
        EmbodiedEvo (frozen harness)~\citep{embodiedevo2026} & \textcolor[HTML]{A83232}{\ding{51}} & \textcolor[HTML]{A83232}{\ding{51}} & \textcolor[HTML]{A83232}{\ding{51}} & \textcolor[HTML]{A83232}{\ding{51}} & \textcolor[HTML]{287A50}{\ding{55}} & \textcolor[HTML]{287A50}{\ding{55}} \\
        WorldScape R$^2$SI (release)~\citep{r2si2026} & \textcolor[HTML]{A83232}{\ding{51}} & \textcolor[HTML]{A83232}{\ding{51}} & \textcolor[HTML]{A83232}{\ding{51}} & \textcolor[HTML]{287A50}{\ding{55}} & \textcolor[HTML]{287A50}{\ding{55}} & \textcolor[HTML]{287A50}{\ding{55}} \\
        \midrule
        \textbf{Auriga (ours)} & \textcolor[HTML]{287A50}{\ding{55}} & \textcolor[HTML]{287A50}{\ding{55}} & \textcolor[HTML]{287A50}{\ding{55}} & \textcolor[HTML]{287A50}{\ding{55}} & \textcolor[HTML]{287A50}{\ding{55}} & \textcolor[HTML]{287A50}{\ding{55}} \\
        \bottomrule
    \end{tabular*}
    \par\smallskip
    \begin{minipage}{\linewidth}
        \footnotesize\raggedright
        \textit{Notes.} \textcolor[HTML]{A83232}{\ding{51}}: included; \textcolor[HTML]{287A50}{\ding{55}}: absent from the specified setting; \textcolor[HTML]{888888}{\raisebox{0.45ex}{\rule{0.45em}{0.65pt}}}: not reported. Scope and qualifications are given in \cref{app:protocol}.
    \end{minipage}
\end{table}

\subsection{The Planning--Execution Gap}
\label{sec:gap}

\paragraph{Deciding without the arm.}
To separate what a vision-language model decides from how the robot arm executes it, we run the same harness with one additional privileged action, which moves an object along waypoints given by the model without using the arm.
The model still chooses the object and every target pose from its own perception.
The object moves under simulated physics, stops at a collision, and settles under gravity at the end, after which the action reports where the object came to rest.
Success under this action therefore estimates what the model's own decisions achieve when execution by the arm is taken out of the loop.
We compare it with two executed systems on the same 210 episodes, five for each of the 42 RoboDojo tasks (\cref{tab:gap}).
The first is RoboProbe~\citep{unexpected2026}, the reference harness released with the evaluation of GPT-6 Astra on RoboDojo~\citep{robodojo2026}, and the second is \method.
Averaged over the five dimensions, the success rate is 76.2\% with the privileged action, 22.5\% for RoboProbe, and 59.2\% for \method.
The model therefore decides considerably better than it executes, and \method recovers 68.3\% of this planning--execution gap in SR (70.6\% in Score).

\providecommand{\ours}{\rowcolor{serule!55}}
\begin{table}[!htbp]
\centering
\footnotesize
\caption{Planning--execution gap on 210 episodes, five for each of the 42 tasks: SR (\%) and Score (0--100) per dimension and their average over the five dimensions, as in \cref{tab:main}. With direct object motion, the model moves objects through a privileged action instead of the arm. All methods use GPT-6 Astra, RoboProbe at medium and the others at high reasoning effort. $^\S$Memory over 29 episodes, since one episode has no counterpart in the reference results.}
\label{tab:gap}
\setlength{\tabcolsep}{3.0pt}
\renewcommand{\arraystretch}{1.15}
\begin{tabular}{l*{12}{c}}
\toprule
 & \multicolumn{2}{c}{Average} & \multicolumn{2}{c}{Generalization} & \multicolumn{2}{c}{Precision} & \multicolumn{2}{c}{Long-Horizon} & \multicolumn{2}{c}{Memory} & \multicolumn{2}{c}{Open} \\
\cmidrule(lr){2-3}\cmidrule(lr){4-5}\cmidrule(lr){6-7}\cmidrule(lr){8-9}\cmidrule(lr){10-11}\cmidrule(lr){12-13}
Method & SR & Score & SR & Score & SR & Score & SR & Score & SR & Score & SR & Score \\
\midrule
Direct object motion (privileged) & 76.2 & 79.6 & 81.7 & 83.3 & 65.0 & 72.0 & 70.0 & 75.4 & 96.7 & 96.7 & 67.5 & 70.6 \\
\midrule
RoboProbe~\citep{unexpected2026} & 22.5 & 28.5 & 31.7 & 34.1 & 2.5 & 8.5 & 7.5 & 22.1 & 40.8$^\S$ & 45.8$^\S$ & 30.0 & 32.1 \\
\ours \method & 59.2 & 64.6 & 43.3 & 46.2 & 50.0 & 55.6 & 45.0 & 59.5 & 90.0 & 90.5 & 67.5 & 71.2 \\
\bottomrule
\end{tabular}
\end{table}

\paragraph{Why common paradigms leave the gap open.}
Two common ways of placing a general model on a robot address execution only indirectly.
In \emph{step-wise action output}, as in the reference harness and in VLA policies~\citep{pi052025,xvla2025,unexpected2026}, the model emits one motion step or chunk per call from the latest frames.
It never sees the route of a motion before committing to it, and a failed motion is noticed only when the model reads the next frame, by which time the arm has already moved.
In \emph{code as policy}~\citep{liang2023code,capx2026}, the model writes a program that commands the control interface directly.
Execution judgment then moves into a program that cannot see the scene.
A failure is detected only if the program was written to check for it, and a program may exit normally even after the motions it requested have failed.
\method keeps the judgment with the model, which describes a complete action sequence in each call rather than a single step.
Before execution, the model can see the whole trajectory, its cost, and its projection on the camera image (\cref{sec:preview}), and the harness checks the measured outcome after every movement (\cref{sec:verify}).

%% file: sec_rsi.tex
\subsection{General-Purpose Harness Evolution}
\label{sec:rsi_positioning}
\label{sec:exp-rsi}

Improvement in embodied agents can target either a procedure for a particular task or the shared capabilities used across many tasks~\citep{duan2026lastai,zhou2026physicalrsi}.

\paragraph{Within-task trial search versus cross-task tool evolution.}
Recent work on recursive self-improvement (RSI) groups autonomous improvement into six levels from L0 to L5~\citep{duan2026lastai}:
\begin{itemize}
    \item \textbf{L0--L2 (within-task search and refinement):} Under a single-task objective, the system repeats trials to adjust a trajectory, script, or task note, so the resulting gains stay mainly on the practiced task.
    \item \textbf{L3--L4 (cross-task experience and tool evolution):} The system turns recurring failures across different tasks into reusable tools, interfaces, and verification steps that apply to unseen tasks.
    \item \textbf{L5 (meta-improvement):} The system revises its own perception, control, and verification mechanisms so that later tasks are solved more reliably.
\end{itemize}
Target-task refinement methods such as HexaShell, EmbodiedEvo, and SEAR~\citep{hexashell2026,embodiedevo2026,sear2026} work in the L0--L2 regime: they reset the target scene across trials to tune coordinate offsets, scripts, or failure notes for that task, which improves performance on the practiced layout but can overfit when layouts change.
Auriga instead focuses on L3--L5 evolution during development.
Failures such as a jammed pin insertion or a slipped cloth grasp are used not to add a coordinate offset for that scene, but to diagnose missing capabilities in 3D localization, contact checking, or in-hand state tracking and to upgrade the shared tools accordingly.

\paragraph{Role of the harness.}
An agent harness connects the model to the robot: it translates the model's intent into motion and stops the robot when contact or slippage occurs, without making task decisions on the model's behalf.
Existing harnesses often either run generated commands open-loop without checking the physical outcome, or add task-specific rules and thresholds in the backend to pass particular tasks.
Because manipulation involves both tight rigid clearances and deformable objects, an offset or grasp rule tuned for one task often interferes with another.
Auriga keeps the harness task-agnostic through three rules:
(1)~\emph{no task-specific logic or privileged state}, so the same code and prompts run across all tasks and randomized layouts;
(2)~\emph{general measurement and control tools}, including calibrated multi-view triangulation, paused-world trajectory preview, stage tolerances, and in-hand slip checks; and
(3)~\emph{in-place recovery}, stopping when a motion leaves its tolerance and returning the measured deviation so the model can continue from the current state.

\begin{figure}[!htb]
  \centering
  \includegraphics[width=0.98\linewidth]{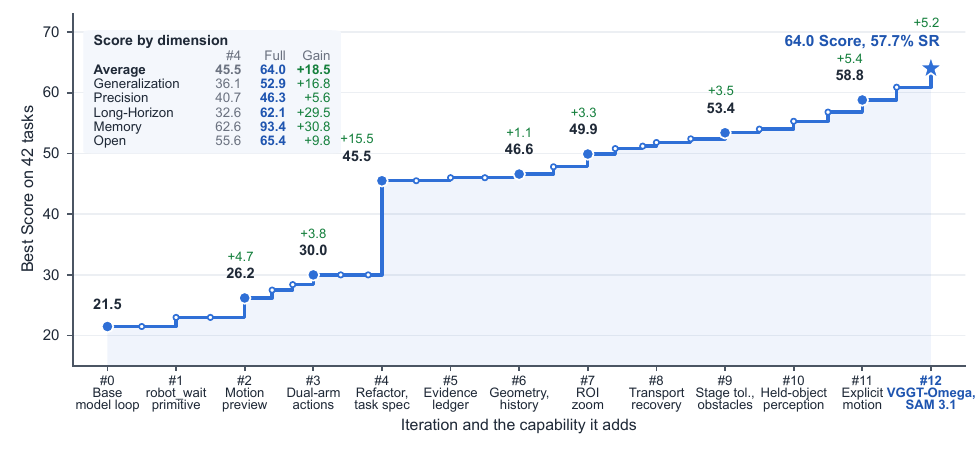}
  \caption{\textbf{General-purpose harness evolution of \method across 13 capability iterations (\texttt{\#0} to \texttt{\#12}).} The horizontal axis lists the task-agnostic tools and harness mechanisms introduced at each iteration, and the vertical axis tracks the resulting performance (overall Score on RoboDojo). Larger nodes, labeled with their Score and gain, mark major capability milestones: motion preview (\texttt{\#2}), unified dual-arm actions (\texttt{\#3}), harness refactoring with verbatim task specifications (\texttt{\#4}), calibrated 3D geometry and history retrieval (\texttt{\#6}), ROI zoom (\texttt{\#7}), stage tolerances and obstacle checks (\texttt{\#9}), and explicit motion control (\texttt{\#11}), culminating in the frozen \method harness with VGGT-Omega and SAM~3.1 perception (\texttt{\#12}, star marker: 64.0 Score, 57.7\% SR across 6,300 episodes). Top-left inset: 42-task Score by capability dimension from iteration~\texttt{\#4} to the full frozen harness.}
  \label{fig:harness_development}
\end{figure}

\paragraph{Two complementary timescales and capability trajectory.}
System development and within-episode adaptation operate at complementary timescales.
Across 13 development iterations (\texttt{\#0} to \texttt{\#12}, \cref{fig:harness_development}), recurring execution bottlenecks across diverse tasks drove task-agnostic upgrades to the shared harness, lifting the RoboDojo Score from 21.5 (4.0\% SR on the 50-case diagnostic set) at the initial base-model loop (\texttt{\#0}) to 45.5 (36.2\% SR across all 42 tasks) at \texttt{\#4} and 64.0 (57.7\% SR across 6,300 episodes) at \texttt{\#12}, with consistent gains across all five capability dimensions ($+16.8$ Score in Generalization, $+5.6$ in Precision, $+29.5$ in Long-Horizon, $+30.8$ in Memory, and $+9.8$ in Open; inset of \cref{fig:harness_development}).
These 13 milestones organize into the three architectural pillars detailed in the next three sections:
\begin{enumerate}
    \item \textbf{Structured task belief and evidence records (\cref{sec:method}):} refactoring the harness around verbatim task specifications (\texttt{\#4}, 45.5 Score), maintaining an explicit evidence ledger (\texttt{\#5}), and adding structured history retrieval (\texttt{\#6}, 46.6 Score);
    \item \textbf{Admission gateway with preview and stage verification (\cref{sec:wysiwye}):} previewing candidate trajectories while simulation is paused (\texttt{\#2}, 26.2 Score), supporting in-place transport recovery (\texttt{\#8}), and enforcing stage arrival tolerances and obstacle clearance checks (\texttt{\#9}, 53.4 Score);
    \item \textbf{Calibrated 3D perception and expressive action interfaces (\cref{sec:augment}):} unifying dual-arm timelines (\texttt{\#3}, 30.0 Score), adding multi-view 3D triangulation (\texttt{\#6}), ROI zoom (\texttt{\#7}, 49.9 Score), held-object perception (\texttt{\#10}), explicit motion control (\texttt{\#11}, 58.8 Score), and VGGT-Omega / SAM~3.1 geometric perception (\texttt{\#12}, 64.0 Score).
\end{enumerate}
During evaluation, the harness version (\texttt{\#12}) is completely frozen across all cases.
Within each evaluated episode, the model uses these shared tools to inspect the scene, diagnose discrepancies, and revise its plan or generated programs from execution feedback, continuing from the physical state left by previous actions under the remaining budget without an environment reset or cross-episode memory transfer.

%% file: methodology.tex
\section{System Overview}
\label{sec:method}

\begin{figure}[!htb]
  \centering
  \includegraphics[width=\linewidth]{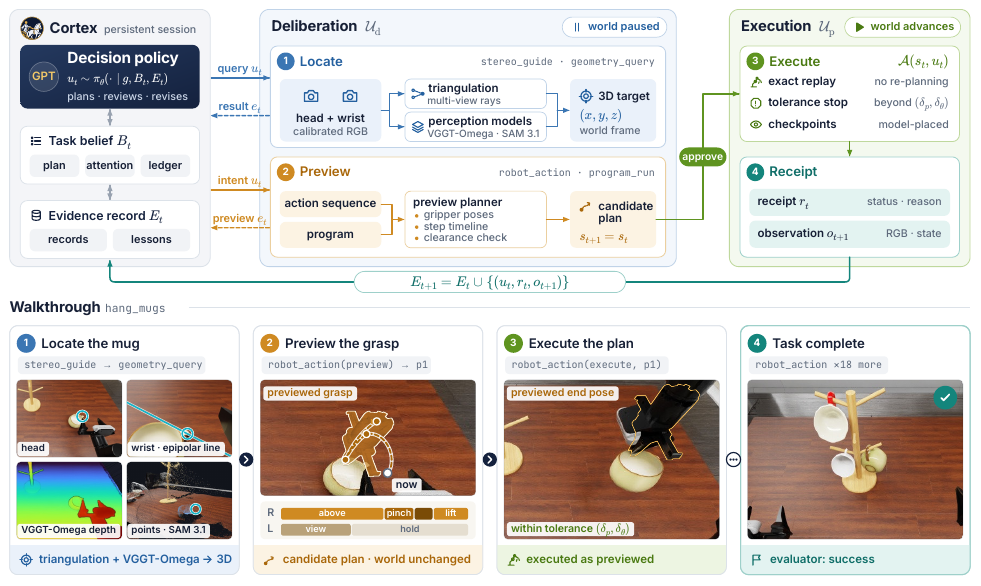}
  \caption{\textbf{Overview of Auriga.}
  \emph{Top:} The model maintains a task belief and an evidence record, uses observations to localize targets, and previews actions while the world is paused. Every physical motion passes through the admission gateway $\mathcal{A}$, which returns a receipt and a new observation.
  \emph{Bottom:} Closed-loop execution on \code{hang_mugs}: (1)~localization, (2)~grasp preview, (3)~execution, and (4)~placement on the rack.}
  \label{fig:overview}
\end{figure}

Large vision-language models can reason about task structure, but reliable manipulation also requires metric scene information, contact-sensitive execution, and feedback after each action.
\method addresses this gap with a simple closed-loop workflow: the model measures and previews while the simulation is paused, submits a physical action through a single admission gateway, and then uses the returned receipt and observation to decide what to do next (\cref{fig:overview}).
The model retains authority over planning and recovery; the harness supplies measurements, enforces the execution boundary, and records what occurred.
This section first formalizes the decision process (\cref{sec:formulation}), then describes the three system layers (\cref{sec:architecture}), the explicit task belief maintained by the model (\cref{sec:belief}), and the three feedback loops that connect planning to execution (\cref{sec:loops}).
\Cref{sec:wysiwye} details execution preview and verification through the admission gateway, and \cref{sec:augment} specifies the perception and action interfaces.

\subsection{Manipulation as Closed-Loop Decision Making}
\label{sec:formulation}

For each benchmark episode, the model must construct a procedure from the task description and the evidence gathered during that episode; no task-specific solution is prepared in advance.
Consequently, each decision uses the information available so far, and each outcome must be returned in a form that can inform the next decision.

\textbf{Formulation.}
A case specifies a task description $g$ and an initial world state $s_0$ (\cref{sec:problem_setting}).
The world state $s_t$ is hidden from the model, which instead receives observations $o_t$ comprising calibrated multi-view RGB images and the measured robot state.
At decision step $t$, a pretrained model with fixed parameters $\theta$ selects a tool call
\begin{equation}
  u_t \sim \pi_\theta(\cdot \mid g, B_t, E_t),
  \label{eq:policy}
\end{equation}
where $B_t$ is its \emph{task belief} (\cref{sec:belief}) and $E_t$ is the episode's \emph{evidence record}, the interactions accumulated so far.
The available calls have two roles.
\emph{Deliberative} calls $u_t \in \mathcal{U}_{\mathrm{d}}$ inspect the scene, measure quantities, compute, retrieve records, revise the belief, or preview a motion.
They return a result $e_t$ while leaving the world unchanged:
\begin{equation}
  s_{t+1} = s_t, \qquad E_{t+1} = E_t \cup \{(u_t, e_t)\}.
  \label{eq:deliberative}
\end{equation}
For example, before hanging a mug, the model can localize the mug and rack, inspect the handle orientation, and preview a grasp without advancing simulated time.
\emph{Physical} calls $u_t \in \mathcal{U}_{\mathrm{p}}$ request robot motion or the passage of simulated time.
They reach the world only through the admission gateway, written as the operator $\mathcal{A}$, which executes an admissible request and returns a receipt $r_t$ and a fresh observation:
\begin{equation}
  (s_{t+1}, r_t, o_{t+1}) = \mathcal{A}(s_t, u_t), \qquad E_{t+1} = E_t \cup \{(u_t, r_t, o_{t+1})\}.
  \label{eq:physical}
\end{equation}
If $\mathcal{A}$ rejects a request, the world remains unchanged and the receipt reports the reason.
The model parameters $\theta$, tool set, and task-independent operating guidance remain fixed throughout an episode; adaptation comes from the evolving $B_t$ and $E_t$.
Because the simulation is paused during deliberation, reasoning and deliberative calls consume no simulated time.

\textbf{Quantitative error feedback.}
A physical request specifies both an intended result and the required precision.
Each stage $k$ has a target $\hat{x}_k$ and a model-chosen tolerance $\tau_k$ (\cref{sec:aug-output}).
During execution, $\mathcal{A}$ measures the reached state $x_k$ and halts when the deviation $\epsilon_k = x_k - \hat{x}_k$ leaves $\tau_k$.
The resulting receipt reports the completed stages, measured deviations, active tolerance, and unexecuted stages (\cref{sec:verify}).
Thus, if a grasp approach in \code{hang_mugs} stops short or the gripper opening differs from the expected value, the model receives the measured discrepancy rather than a generic failure flag.
It can then decide whether to inspect the scene, revise the approach, or execute the remaining stages; $\mathcal{A}$ does not choose the recovery.

\textbf{Information boundary.}
The harness exposes only the published task text, calibrated RGB images, the robot's measured state, and the public robot model and camera calibration.
From these inputs, it may compute learned depth, per-object 3D estimates, and point clouds, reporting each as an estimate with its reliability (\cref{sec:aug-input}).
The model and its tools cannot access simulator depth, hidden object poses, contact state, or evaluator internals.
The benchmark evaluator, rather than the model's own assessment, determines task success.

\subsection{Separating Decision, Capability, and Admission}
\label{sec:architecture}

\method separates task reasoning from measurement and execution so that a general-purpose model can build a procedure while every physical action remains reviewable.

\textbf{Division of responsibility.}
The harness is organized into three layers (\cref{fig:overview}): a model session that makes every task decision, a capability layer of general tools, and an admission gateway that is the only path to the robot.
Each layer has a distinct responsibility:
\begin{itemize}
  \item \emph{Cortex} is one persistent model session for the entire episode. It receives task-independent operating guidance, the official task text, and the robot's capabilities and coordinate conventions. It chooses task decompositions, measurements, action targets, and recovery steps without human intervention or a task-specific prompt, program, or skill.
  \item The \emph{capability layer} provides general tools to measure, act, compute, and recall (\cref{tab:capabilities}). For \code{hang_mugs}, these tools can estimate mug and rack locations, compare views, preview a grasp, or retrieve an earlier receipt; they do not encode a solution for hanging mugs.
  \item The \emph{admission gateway} implements $\mathcal{A}$. It is the only path that moves the robot. It receives a reviewed physical request, determines whether the request is admissible, monitors execution, and records the outcome.
\end{itemize}

\begin{table}[!htb]
  \centering
  \small
  \setlength{\tabcolsep}{4pt}
  \renewcommand{\arraystretch}{1.12}
  \caption{Capabilities available to the model. Only physical calls advance the world, and they do so only through the admission gateway $\mathcal{A}$.}
  \label{tab:capabilities}
  \begin{tabular*}{\linewidth}{@{\extracolsep{\fill}}p{1.6cm}p{5.4cm}p{4.4cm}cc@{}}
    \toprule
    \textbf{Capability} & \textbf{Role} & \textbf{Representative tools} & \textbf{World} & \textbf{Details} \\
    \midrule
    Measure & Return images, robot state, object estimates, and geometric or state quantities with their reliability & \code{robot_observe}, \code{geometry_query}, \code{stereo_guide} & -- & \cref{sec:aug-input} \\
    \addlinespace[3pt]
    Act & Preview and execute multi-stage action sequences for one or both arms, or let simulated time pass & \code{robot_action}, \code{robot_wait} & $\mathcal{A}$ & \cref{sec:preview,sec:verify,sec:aug-output} \\
    \addlinespace[3pt]
    Compute & Run programs for loops, search, and numerical work; their physical requests are reviewed by the model & \code{program_run} & $\mathcal{A}$ & \cref{sec:preview} \\
    \addlinespace[3pt]
    Recall & Revise the task belief, retrieve earlier interactions, and record or query operation lessons & \code{task_plan}, \code{history_*}, \code{operation_memory} & -- & \cref{sec:belief,sec:aug-input} \\
    \bottomrule
  \end{tabular*}
\end{table}

\textbf{One-step dataflow.}
\label{sec:measurement}
At each decision step, the model first gathers the evidence needed for the next choice through the capability layer (\cref{sec:aug-input}).
\label{sec:single_path}
It then submits either a complete multi-stage action sequence or a program that requests one (\cref{sec:aug-output}).
Both routes converge on $\mathcal{A}$, so a program-generated motion is previewed and reviewed in the same way as a direct request (\cref{sec:preview}).
After a physical call, the harness stores the reviewed request, receipt, observations before and after execution, and measured displacement in one interaction record (\cref{sec:aug-input}).
These records make completed motion, normal program termination, and benchmark success separately observable instead of conflating them.

\textbf{Transparent execution.}
\label{sec:records}
The model sets task decomposition, targets, and the need for additional evidence; the harness exposes the thresholds that admit, pause, or stop an action and reports their effects.
Every physical request, whether direct or program-generated, follows the same admission path.
The harness neither substitutes a task-specific rule nor silently changes a parameter supplied by the model.
The records constitute $E_t$, remain outside the model context, and can be retrieved on demand (\cref{sec:aug-input}).
The model may attach short operation lessons to them, but it rechecks each lesson against the current scene before using it.

\subsection{Deliberation over an Explicit Task Belief}
\label{sec:belief}

Long-horizon manipulation accumulates observations, receipts, and intermediate decisions.
Without a compact representation of the current task understanding, a model can lose track of an earlier objective or confuse an issued command with a verified outcome.
\method therefore maintains a revisable task belief $B_t$ alongside the detailed evidence record $E_t$.

\textbf{Structure.}
The belief has three concise components.
The \emph{plan} records the model's current understanding of the goal, its intended approach, and the visible evidence that would indicate completion.
The \emph{attention} list records the uncertainties that matter now, the evidence that could resolve each one, and the part of the procedure that depends on it.
The \emph{ledger} records a small number of task-critical premises.
Each premise includes an expected event, a status (\emph{pending}, \emph{supported}, \emph{contradicted}, or \emph{unknown}), the permitted evidence for that status, and the next observation or action that could change it (the labeled premises in \cref{fig:teaser}).
The harness admits no physical action until the model has submitted a belief.

For example, in \code{hang_mugs}, the plan can state ``grasp the mug by its handle, lift it, and place it on the rack hook.''
An attention item can require a view that resolves the handle orientation before the approach.
A ledger premise can state that the mug is retained by the gripper: it is initially \emph{pending}, and becomes \emph{supported} only after permitted evidence, such as the expected gripper opening together with a confirming lift observation, is available.

\textbf{Belief versus evidence.}
$B_t$ is a compact, revisable working state rather than an action script or a source of evidence.
Concrete poses, arm timing, and gripper commands remain part of individual actions, so the model can change them as new measurements arrive.
Likewise, a belief entry must be checked against fresh observations before it supports a decision: a closed gripper alone does not verify that a mug is retained.
When the conversation is compacted to fit the context window, the harness reinserts the latest $B_t$ as the model's provisional judgment, while the details of earlier interactions remain retrievable from $E_t$.

\subsection{Closing the Loop at Three Time Scales}
\label{sec:loops}

Errors can arise within a movement, between successive actions, or only after a task-level premise has been tested.
\method addresses these cases with three nested loops that have distinct responsibilities: the admission gateway stops unsafe or off-target execution, while the model diagnoses the result and chooses the recovery.
\Cref{fig:loop} shows how the three loops interleave within one episode.

\textbf{Stage loop.}
Within a single action, $\mathcal{A}$ contains errors: applying the stage check of \cref{sec:formulation} during execution, it halts at the first stage whose deviation $\epsilon_k$ leaves $\tau_k$, and also pauses at a failed grasp check or at a checkpoint placed by the model (\cref{sec:verify}).
For a \code{hang_mugs} grasp, it can stop the sequence immediately when the measured gripper opening indicates that the mug handle was not secured.
The stage loop keeps an error from propagating to later stages; it does not repair the motion.

\textbf{Action loop.}
Between actions, the model diagnoses outcomes: after each physical call, it compares the receipt and fresh observation with its intended result and forms a cause hypothesis before acting again.
It may continue the remaining previewed stages, replan from the stopped stage without repeating completed stages, change the contact, route, target, arm, or tolerance, or collect an observation that distinguishes competing explanations.
For example, after an unsuccessful mug grasp, the model can inspect whether the handle orientation, approach route, or target localization caused the discrepancy before choosing a new approach.
Whenever possible, normal task progress tests the hypothesis; a separate probe is used only when its result would alter the next decision.

\textbf{Belief loop.}
Across the episode, the model revises its premises: action outcomes update the ledger and, when necessary, the plan in $B_t$.
If repeated evidence contradicts a premise, the model changes its task-level explanation and selects a different mechanism, such as another contact, route, or target, rather than simply restating the same hypothesis.
In \code{hang_mugs}, evidence that a top-down approach repeatedly fails can update the belief so that the model seeks a handle-oriented grasp or a different approach route.
This loop keeps the model's working understanding aligned with the evidence accumulated during the episode.
\Cref{sec:wysiwye} next describes the admission path in which the stage loop runs.

\begin{figure}[!htb]
  \centering
  \begin{tcolorbox}[colback=white, colframe=black!35, boxrule=0.4pt, arc=1pt,
                    left=6pt, right=6pt, top=4pt, bottom=4pt, width=\linewidth]
    \small
    \textbf{Input:} task description $g$, initial observation $o_0$, task-independent operating guidance, fixed model $\pi_\theta$.\\
    \textbf{Initialize:} $E_0 \leftarrow \{o_0\}$; the model observes, measures, or previews and submits a belief $B_0$ before any physical call.\\[2pt]
    \textbf{repeat} \hfill \emph{belief loop}
    \begin{enumerate}
      \setlength{\itemsep}{1pt}
      \item \emph{Decide:} the model selects $u_t \sim \pi_\theta(\cdot \mid g, B_t, E_t)$ (\cref{eq:policy}).
      \item \emph{Deliberate:} if $u_t \in \mathcal{U}_{\mathrm{d}}$, return an observation, measurement, computation, retrieved record, or preview $e_t$; the world is unchanged (\cref{eq:deliberative}).
      \item \emph{Act:} if $u_t \in \mathcal{U}_{\mathrm{p}}$, issued directly or by a program, apply $\mathcal{A}$ (\cref{eq:physical}). \hfill \emph{stage loop}\\
      For each stage $k$: execute, measure $x_k$, and halt if $\epsilon_k$ leaves $\tau_k$, if the grasp check fails, or at a model checkpoint.
      Return $r_t$ and $o_{t+1}$, and append the record to $E_{t+1}$.
      \item \emph{Assess:} the model compares $r_t$ and $o_{t+1}$ with its intent and states a cause hypothesis for any deviation. \hfill \emph{action loop}\\
      It continues, replans from the stopped stage, changes the contact, route, target, arm, or tolerance, or gathers a distinguishing observation.
      \item \emph{Revise:} the model updates the status of affected premises and, if its understanding has changed, the plan in $B_{t+1}$.
    \end{enumerate}
    \textbf{until} the evaluator reports termination or the execution budget is exhausted.
  \end{tcolorbox}
  \caption{\textbf{The three nested loops of \method within one episode.}
    The stage loop runs inside $\mathcal{A}$ and stops execution when a declared condition is violated; the model closes the action and belief loops.
    Deliberative calls extend the evidence record without advancing the world.}
  \label{fig:loop}
\end{figure}

%% file: sec5_execution.tex
\section{Admission Gateway: Preview and Verification}
\label{sec:wysiwye}

To close the planning--execution gap quantified in \cref{sec:gap,tab:gap}, \method binds physical execution to a trajectory preview that the model can inspect before committing (\cref{sec:preview}) and verifies the measured state as the trajectory runs (\cref{sec:verify}).
\method follows a single rule, enforced by the admission operator $\mathcal{A}$: the motion that reaches the robot is exactly the motion compiled into its preview, and whenever execution departs from it beyond the declared tolerance, the harness stops and reports the measured state rather than repairing it on the model's behalf.

\subsection{Preview Before Acting}
\label{sec:preview}

Preview gives the model a first guarantee, namely that every motion, whether requested by the model or by a program it has written, is compiled into a preview before it runs, and that nothing else is executed.
The model reviews this preview by default and can skip the review to save a call.
Below, we describe how the model sees a motion and how execution is held to it.

\begin{figure}[!htb]
  \centering
  \includegraphics[width=\linewidth]{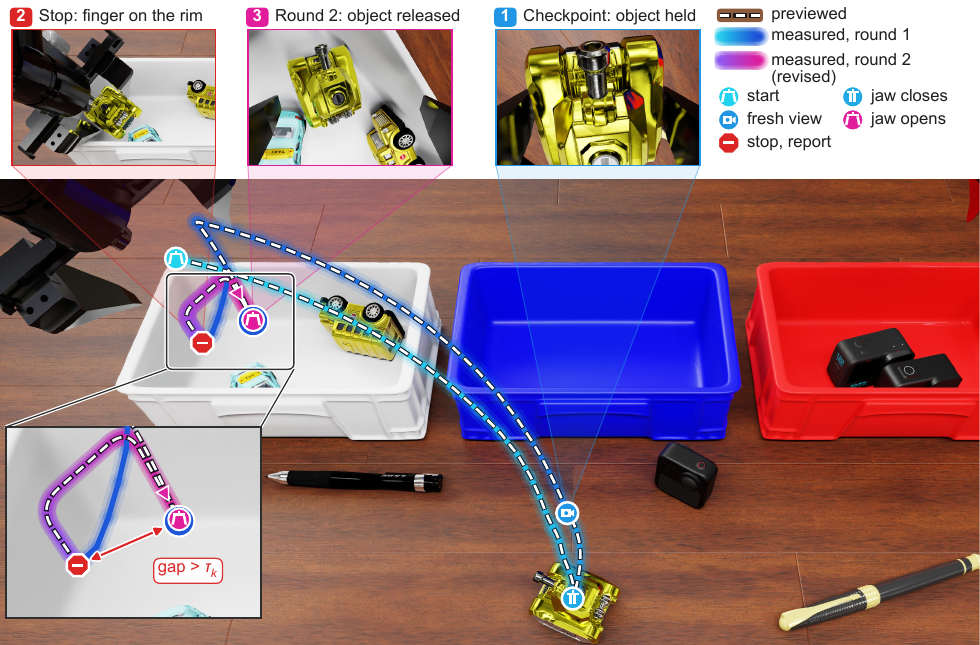}
  \caption{Preview and verification in two rounds, drawn on the head-camera image. In the first round, the previewed gripper path moves to the object, closes on it, and lifts it to a checkpoint placed by the model, where the robot pauses and the model receives a fresh wrist view (inset 1). The measured path follows the preview until a finger lands on the basket rim while the object is lowered. Once the gap exceeds the tolerance, the harness stops the arm and reports with a fresh view (inset 2), so the gripper does not open. In the second round, the model previews a revised motion from the stop that lifts clear of the rim and lowers into the basket, where the gripper opens (inset 3). The rim contact and the second round are illustrative.}
  \label{fig:preview-verify}
\end{figure}

\textbf{Seeing the whole motion.}
Rather than issuing one movement per call, the model describes a complete action sequence for one arm or for both arms at once, such as approaching an object, closing the gripper, lifting, transporting, and releasing it.
The harness plans the entire sequence into a robot trajectory before anything moves and returns a preview of what would happen.
Any adjustment the harness makes to how a movement is realized happens at this stage, so the preview already shows the motion that will run.
The preview draws the planned gripper paths (\cref{fig:preview-verify}) and arm configurations on the camera image and reports how long each part of the sequence would take on each arm.
It also estimates the clearance between the robot, the objects it holds, and the obstacles the model has declared, such as whether the two arms come close to each other.
Such estimates are given as advice, and the decision whether to proceed stays with the model.
If any part of the sequence cannot be planned, nothing is executed, and the preview states which part failed and why.

\textbf{Committing to it.}
Once the model accepts a preview, the harness replays the previewed trajectory without replanning it.
Because the simulation is paused while the model inspects the preview, the robot is still in the state from which the preview was planned, and looking ahead costs no execution time.
The robot therefore performs the previewed motion, so every later report refers to a motion the model can inspect.
In free space the robot follows the previewed trajectory closely, which means that a deviation during execution almost always indicates contact, the case that \cref{sec:verify} is built around.

\subsection{Verify While Acting}
\label{sec:verify}

Preview fixes what will be executed, but not how the world will respond.
Contact, slip, and objects displaced by earlier actions can still make the outcome differ from the plan.
Verification gives the second guarantee, namely that every divergence of the robot from its previewed motion beyond the tolerance is caught at the end of the movement in which it occurs and handed back to the model together with the facts needed to act on it.

\textbf{Catching divergence.}
After each movement, the harness compares where the gripper actually is with where the preview placed it.
The tolerance $\tau_k$ of a movement (\cref{sec:formulation}) consists of a position part $\delta_p$ and an orientation part $\delta_\theta$.
The model can set both for each movement, by default $\delta_p{=}10$\,mm and $\delta_\theta{=}0.15$\,rad.
If the difference exceeds the tolerance set for that movement, the robot stops and holds its pose instead of continuing with the rest of the sequence.
The receipt $r_t$ that follows states which movement failed, by how much it missed, and which parts did not run.
It also adds a fresh camera view from the stopping point and images of the measured and the previewed route.
Movements that already completed are kept rather than undone.
In \cref{fig:preview-verify}, a finger lands on a basket rim while the object is lowered, and the measured path leaves the previewed one beyond its tolerance.
The harness stops the arm and reports only this miss, and the model identifies the rim as the cause from the fresh view.
It then previews a revised motion that clears the rim, which completes the placement in a second round.
This check targets the main source of divergence, contact with an object or fixture, which the planner does not model, so the harness detects its consequence instead of predicting it.

\textbf{Looking where the model chooses.}
Some divergences leave the robot on its planned path and show only in the scene, such as an object that slips or shifts during a grasp.
The model can therefore insert checkpoints at moments it considers critical, either unconditionally or only when the measured gripper opening after a grasp falls outside the range it expects, for example when the fingers close on nothing.
At a checkpoint the robot stops and the model receives a fresh view, after which it either continues the remaining sequence as previewed or replaces it.
In \cref{fig:preview-verify}, the model places a checkpoint after the lift, and the sequence continues as previewed once the fresh wrist view shows the object still held.
Checkpoints are placed by the model rather than at fixed points, since pausing after every gripper action by default fragmented long tasks.
Acting on what a stop or a checkpoint reveals, however, often requires facts that the model cannot read reliably from RGB, such as whether an object is held and where it sits relative to the gripper, which the following section provides.


%% file: sec6_augment.tex
\section{Capability Layer: Perception and Action Interfaces}
\label{sec:augment}

\begin{figure}[!htb]
  \centering
  \includegraphics[width=\linewidth]{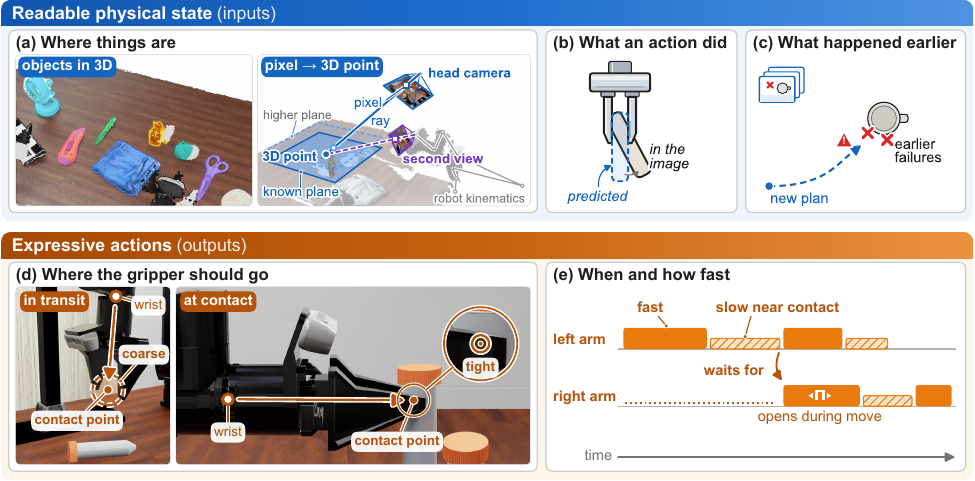}
  \caption{Perception and action interfaces in \method. (a) Objects separated and placed in 3D from the head-camera image, and the position of a selected point from its pixel ray and a plane of known height or a second view. (b) The object estimates keep a held object at its pose in the hand at the grasp, and a measurement against this reference shows that it has turned in the hand. (c) A new plan that returns to where earlier motions failed carries a warning, and earlier frames remain retrievable. (d) A goal refers to the contact point on the fingers rather than the wrist, with a coarse tolerance in transit and a tight one at contact. (e) Both arms share one timeline, with declared waits, a speed per movement that slows near contact, and a gripper that opens during a movement.}
  \label{fig:augment}
\end{figure}

Faithful execution helps only if the motion the model chooses is the right one, which depends on what the model can perceive and on how it can express what it wants.
A general-purpose vision-language model reads a scene and plans a task well, but it is weak at the quantitative side of manipulation.
This section describes how the harness compensates on both sides, by turning physical state into facts the model can read (\cref{sec:aug-input}) and by letting the model state its intent in terms of the task (\cref{sec:aug-output}), as summarized in \cref{fig:augment}.
Two principles guide both sides.
The harness \emph{informs but does not decide}, reporting measurements and estimates together with how reliable they are and leaving the choice of what to do with them to the model.
It also \emph{attaches facts at the point of decision}, so that information that should change the next command arrives with the preview or the report instead of waiting to be requested.

\subsection{Readable Physical State}
\label{sec:aug-input}

On the input side, every quantity the model needs for its next decision reaches it as a measured, computed, or estimated fact together with how reliable it is, rather than as something the model has to infer from pixels.
We describe how the model locates things, learns what an action did, and recalls what happened earlier.

\textbf{3D localization.}
The model understands the layout of a scene well from RGB, but an image does not show metric 3D positions directly, so any estimate made from appearance alone remains uncertain.
A deployed robot, however, comes with physical quantities that are known exactly, namely the intrinsic and extrinsic calibration of every camera, the height of the work surface, and the kinematic model of the robot itself.
With the calibration, every pixel $(u,v)$ of a camera with intrinsics $K$, orientation $R$, and center $\mathbf{c}$ lies on a known ray with direction $\mathbf{d}=R\,K^{-1}(u,v,1)^\top$, so locating it in 3D reduces to finding its distance along this ray.
The harness provides two tools that supply this distance at different levels of precision (\cref{fig:augment}a).
The first estimates every visible object in 3D.
VGGT-Omega~\citep{wang2026vggt} predicts a depth for every pixel of the head-camera image, which places the pixel on its ray, and the depths are scaled so that the table lies at its known height.
SAM~3.1~\citep{carion2026sam} separates the objects in these points, and each object is reported with its 3D position, orientation, and size and with how reliable the estimate is, in every full observation and every action report.
Where such an estimate is not precise enough, as for placing the fingers on a narrow edge, the second tool computes the position of a point that the model selects in an image.
The model fixes its distance either with a plane $\mathbf{n}^\top\mathbf{p}=h$ of known height,
\begin{equation}
  \mathbf{p} = \mathbf{c} + \frac{h-\mathbf{n}^\top\mathbf{c}}{\mathbf{n}^\top\mathbf{d}}\,\mathbf{d},
  \qquad
  \frac{\partial\mathbf{p}}{\partial h} = \frac{\mathbf{d}}{\mathbf{n}^\top\mathbf{d}},
  \label{eq:backproject}
\end{equation}
or by marking the same point in a second view, such as the wrist camera, from which the harness triangulates it.
The two tools meet in the plane height.
For a point on the table, the height is known exactly.
For a point on an object, the model either takes the height from the learned depth of that object, in which case the harness reports how far the point would move if this height were off, or marks the point in a second view, which needs no height.
Any 3D point, including the fingers placed by the kinematic model, can be projected back into the image to check it.

\textbf{Action outcomes and in-hand state.}
After an action, the model needs its physical outcome.
For most of the scene, the next image shows it, and each report also lists the objects whose estimated position changed during the action.
A grasp is harder, since the fingers often hide the object, so the image may not show the state of the object and how it interacts with the gripper.
The harness therefore complements the image with measurements of both.
For the gripper, every report includes its measured opening.
For the object, the model can measure on request where a visible part of it, such as its lower end, lies relative to the gripper, the axis of the object in the 3D points around the fingers, and how far a part marked in a wrist image before an action has moved after it.
The object estimates also keep the pose that the object had in the hand at the grasp, as a reference for these measurements.
Together, these tell the model whether the object is held, how it sits in the hand, and whether it has turned or slipped since the grasp.

\textbf{Episode history and warnings.}
In long episodes the context is compacted and early images are dropped, while recovery often depends on what earlier attempts actually achieved.
The model can therefore retrieve from the evidence record $E_t$ (\cref{sec:formulation}) any earlier camera frame, including a magnified region of it, a list of past actions with the displacement that each one measured, and the planned and measured trajectories of any action.
Full history is retrieved on demand rather than pushed into every turn, since forced reviews of history produced idle calls.
The exception is history that bears directly on the next command.
A report of a stopped or failed action carries a short summary of recent outcomes, and a preview that returns to a place where a motion has failed carries a warning, so that the model does not unknowingly repeat a failed attempt.

\subsection{Expressive Actions}
\label{sec:aug-output}

On the output side, the model states what it wants in terms of the task, such as where the fingers should touch and how the two arms should cooperate, and the harness turns these statements into motion.
We describe how the model specifies the goal of each movement and how it sets the timing and speed of a sequence.

\textbf{Finger-referenced goals.}
A robot interface typically takes the pose of the wrist as its goal, whereas what matters for a task is where the fingers touch the object and how precisely they get there.
In \method, each movement states its goal in these terms.
The model gives where the gripper should end up and how it should be turned in a frame fixed to the workspace, together with the part of the fingers that should make contact, such as the fingertip or the pad center, so that the goal refers to the point that touches the object rather than to the wrist.
It also states how precisely the gripper must arrive, so that a tight insertion and a coarse transit are checked differently.

\textbf{Shared timeline and speed.}
Goals alone do not fix how a sequence unfolds in time.
When both arms take part, or when a movement passes from free space to contact, the order and the speed of its parts matter as much as its goals.
In \method, the stages of both arms belong to the same sequence and advance on a shared timeline.
Stages of the two arms run at the same time unless the model declares that one must wait for another.
The model also sets the speed of each movement, typically slower near contact, and can let the gripper open or close during a movement, for example opening it on the way to an object.
Speed and gripper timing are fixed in the preview like the path itself, so they are part of the reviewed motion.


%% file: experiments.tex

\providecommand{\ours}{\rowcolor{serule!55}}

\FloatBarrier

\begingroup
\setlength{\emergencystretch}{2em}
\makeatletter\setlength{\@fptop}{0pt}\makeatother
\input{sec7_setup}
\input{sec8_results}
\input{sec_exp_qualitative}
\endgroup

%% file: sec7_setup.tex
\section{Experiments}
\label{sec:exp-main}

We evaluate \method under the zero-shot protocol of \cref{sec:problem_setting} on two simulation benchmarks: dual-arm tabletop manipulation on \textbf{RoboDojo}~\citep{robodojo2026} (\cref{sec:exp-robodojo}) and kitchen mobile manipulation on \textbf{RoboCasa365}~\citep{robocasa3652026} (\cref{sec:exp-robocasa}).
After describing the experimental setup (\cref{sec:exp-setup}), we report main results on RoboDojo (\cref{sec:main-overall}) and RoboCasa365 (\cref{sec:main-robocasa}), mechanism ablations (\cref{sec:exp-ablation}), execution efficiency and tool usage (\cref{sec:main-cost}), and qualitative behavior and failure modes (\cref{sec:cases}).

\subsection{Experimental Setup}
\label{sec:exp-setup}

\subsubsection{RoboDojo}
\label{sec:exp-robodojo}
\label{sec:exp-dims}
\label{sec:exp-sets}
\label{sec:exp-metrics}
\label{sec:exp-config}

RoboDojo~\citep{robodojo2026} simulates a dual-arm ARX~X5 robot with one head camera and two wrist cameras in Isaac~Sim.

\textbf{Capability dimensions.}
RoboDojo groups its 42 tasks into five capability dimensions: Generalization (12 tasks, each with standard and random scenes), Precision (8), Long-Horizon (8), Memory (6), and Open (8); \cref{tab:dims} in \cref{app:tasks} lists what each dimension covers and its tasks.
As on the official leaderboard, results are first averaged across the tasks of each dimension and then across the five dimensions, with equal weight at both levels.
For Generalization, the value of each task and of the dimension is the equal-weight mean of its standard-scene average and its random-scene average.

\textbf{Evaluation set.}
Following \cref{sec:problem_setting}, a case specifies a task, a scene variant (standard or random), an evaluation seed, and a layout, and an episode is one execution of a case by the agent.
The evaluation set follows the protocol of the official leaderboard, with three evaluation seeds and, in each seed, 50 cases for each of the 42 tasks, that is, 2,100 cases per seed and 6,300 in all.
In each seed, every Generalization task contributes 25 cases with standard scenes and 25 with random scenes, and every other task contributes 50 with standard scenes.
Each case counts with its first official score, including scores of zero, and any later episode of the same case is ignored.
When an infrastructure fault ends an episode before the evaluator has scored it, the case is re-run under the same configuration.

\textbf{Metrics.}
\emph{SR} is the share of cases that the benchmark evaluator judges fully successful, over a fixed denominator: a case without an official score stays in the denominator and does not count as a success.
\emph{Score} is the official graded process score of the evaluator, from 0 to 100, averaged over the cases with an official score.
Both metrics are first computed for each task in each seed, and a task's value is the mean of its three per-seed values; the dimension values and the five-dimension average then follow from the task values as in \cref{sec:exp-dims}.
\emph{Steps} are the native steps that the simulator executes in an episode, also reported as a percentage of the official native-step limit of the task, which ranges from 200 to 1,900 steps; a single action request may use at most 250 steps.
\emph{Calls} are the model calls (the model responses in an episode) and the direct tool calls issued by the model; tool calls issued by model-written programs are counted separately.
\emph{Cost} comprises the input tokens of the model session (total, share served from the prompt cache in \%, and uncached input tokens), the output tokens, and the wall-clock time of the episode; Steps, Calls, and Cost are reported as means per episode.

\textbf{Agent configuration.}
The RoboDojo results (\cref{sec:main-overall,sec:main-cost,sec:exp-tools}) use \method as described in \cref{sec:method,sec:wysiwye,sec:augment}, with the model \code{gpt-6-astra} at reasoning effort high.
An episode may use at most 170 model responses and the official native-step limit of its task, and the simulation is paused while the model deliberates (\cref{sec:formulation}).
The model observes RGB images from the three cameras, proprioception, and the public camera calibration, together with the object estimates that the harness computes from the head-camera image (\cref{sec:aug-input}); simulator depth and hidden object state are withheld.
Its task text is the native instruction of the task, the official task page with its description and scoring rules, and the task-independent operating guidance of the harness (\cref{sec:formulation,sec:architecture}).
The ablations in \cref{sec:exp-ablation} remove or restrict individual mechanisms of this configuration while keeping the rest fixed.

\subsubsection{RoboCasa365}
\label{sec:exp-robocasa}

RoboCasa365~\citep{robocasa3652026} evaluates household mobile manipulation with a single Franka Panda arm mounted on an Omron mobile base across kitchen environments, spanning 365 tasks (65 atomic and 300 composite).

\textbf{Tasks and task groups.}
Evaluation focuses on the benchmark's 50 Target50 tasks, organized into three groups: 18 \emph{Atomic-Seen} tasks involving single manipulation skills (e.g., opening a drawer or operating a microwave); 16 \emph{Composite-Seen} tasks that chain two to fifteen subtasks whose templates appear in the benchmark's pretraining demonstrations; and 16 \emph{Composite-Unseen} tasks of comparable horizon whose templates are absent from the pretraining demonstrations.

\textbf{Evaluation protocol and metrics.}
Whereas the standard policy leaderboard~\citep{robocasaboard2026} evaluates demonstration-trained policies in the 2,500 pretraining kitchens, agent evaluations are conducted in ten unseen target kitchens with novel layouts, fixtures, and objects.
We follow the protocol of the published agent baselines~\citep{harnessvla2026,pyrua2026}: every task is evaluated at seeds 1--5, giving 250 episodes (90 Atomic-Seen, 80 Composite-Seen, and 80 Composite-Unseen).
We report the success rate (SR, \%) of each group, of all 160 composite episodes, and \emph{Overall} as the mean over the 50 tasks, which equals the SR over all 250 episodes.
The benchmark environment provides camera images, robot proprioception, world-coordinate maps derived from rendered depth, the source code of the task's success check, and live values of the predicates tested by that check.

\textbf{Baselines.}
Both published agent baselines plan with an LLM agent and hand contact-rich phases, such as grasping an object or turning a knob, to the RLDX-1 vision-language-action (VLA) policy~\citep{rldx2026}.
RPent~\citep{harnessvla2026} also retrieves task memory, built by exploring each task on seed~0 before evaluation.
PyRUA-Lean~\citep{pyrua2026} uses no cross-episode memory.
We also list RLDX-1 on its own and, as in the comparison of RPent, three learned policies evaluated in the pretraining kitchens.

\textbf{Agent configuration.}
\method runs \code{gpt-6-astra} at \code{high} reasoning effort with at most 100 model requests per episode.
Unlike both agent baselines, it uses no VLA policy and no cross-episode memory.
Alongside the benchmark-provided observations and task predicates, \method segments camera images via SAM~3.1~\citep{carion2026sam} from model-chosen text prompts and projects each mask onto the world coordinate map, without reading ground-truth object poses.
All physical motions are executed through multi-stage action programs under the preview-and-verification gateway of \cref{sec:wysiwye}, coordinating stages for the arm, gripper, torso, and mobile base.

%% file: sec8_results.tex
\subsection{Main Results on RoboDojo}
\label{sec:main-overall}

\Cref{tab:main} places \method on the official leaderboard~\citep{robodojoboard2026}, reporting Score and SR on average and in each dimension.
The table has two blocks: the eight leading entries that use a learned policy, either a world action model (WAM) or a vision-language-action (VLA) policy, alone or with an agent; and four agent methods, which act through tools or code, together with \method.
Averaged across the five dimensions, \method reaches a Score of 64.00 and an SR of 57.74\%.
Across both blocks of \cref{tab:main}, \method achieves the highest overall averages, ahead of GO-3 Nano Preview, which combines an agent with a world action model, at 61.90 and 56.42\%.
Within the agent block, \method leads all twelve columns, including both averages (ahead of PyRUA-Lean at 55.26 and 50.35\%).
According to their reports~\citep{pyrua2026,hexashell2026,embodiedevo2026,unexpected2026}, PyRUA-Lean, HexaShell-Preview, EmbodiedEvo, and RoboProbe all use \code{gpt-6-astra}, the same base model as \method (marked $^\dagger$ in \cref{tab:main}), with reasoning effort high for PyRUA-Lean, xhigh for EmbodiedEvo, and medium for RoboProbe; \method uses high.
Against RoboProbe~\citep{unexpected2026}, the reference harness released with the evaluation of GPT-6 Astra on RoboDojo~\citep{robodojo2026}, \method raises the average Score from 28.97 to 64.00 and the average SR from 22.48\% to 57.74\% on the same model (\cref{tab:main}).

\begin{table}[!htbp]
\centering
\footnotesize
\caption{Results on the RoboDojo simulation benchmark, with external entries from the official leaderboard~\citep{robodojoboard2026} (\cref{sec:exp-robodojo}), reporting Score (0--100) and SR (\%) in two blocks ordered by ascending average Score: the eight leading entries that use a world action model (WAM) or a vision-language-action (VLA) policy, alone or with an agent, and four agent methods on the leaderboard (Agent, acting through tools or code) together with \method. In each column, the highest value is in bold and the second highest is underlined. $^\dagger$Uses \code{gpt-6-astra}.}
\label{tab:main}
\setlength{\tabcolsep}{1.5pt}
\renewcommand{\arraystretch}{1.08}
\begin{tabular}{l>{\hspace{4pt}}l>{\hspace{5pt}}rr>{\hspace{5pt}}rr>{\hspace{5pt}}rr>{\hspace{5pt}}rr>{\hspace{5pt}}rr>{\hspace{5pt}}rr}
\toprule
 & & \multicolumn{2}{>{\hspace{5pt}}c}{\makebox[0pt]{Average}} & \multicolumn{2}{>{\hspace{5pt}}c}{\makebox[0pt]{Generalization}} & \multicolumn{2}{>{\hspace{5pt}}c}{\makebox[0pt]{Precision}} & \multicolumn{2}{>{\hspace{5pt}}c}{\makebox[0pt]{Long-Horizon}} & \multicolumn{2}{>{\hspace{5pt}}c}{\makebox[0pt]{Memory}} & \multicolumn{2}{>{\hspace{5pt}}c}{\makebox[0pt]{Open}} \\
\cmidrule(lr){3-4}\cmidrule(lr){5-6}\cmidrule(lr){7-8}\cmidrule(lr){9-10}\cmidrule(lr){11-12}\cmidrule(lr){13-14}
Method & Type & Score & SR & Score & SR & Score & SR & Score & SR & Score & SR & Score & SR \\
\midrule
\multicolumn{14}{l}{\emph{Learned policies: WAM or VLA, alone or with an agent}} \\
Rex-M1 Preview & VLA & 33.79 & 27.44 & 34.05 & 26.05 & 32.82 & 23.83 & 46.89 & 33.00 & 48.53 & 48.22 & 6.67 & 6.08 \\
Simate-beta & VLA & 33.95 & 27.96 & 35.09 & 27.95 & 34.35 & 26.92 & 57.84 & 43.42 & 33.33 & 33.00 & 9.12 & 8.50 \\
Awomo-0.5 & WAM & 35.34 & 29.64 & 30.31 & 23.84 & 39.65 & 33.50 & 40.36 & 26.75 & 42.27 & 41.22 & 24.09 & 22.92 \\
PhysicalRSI & Agent + VLA & 36.27 & 31.38 & 21.30 & 15.62 & 38.05 & 32.50 & 46.37 & 37.33 & 46.74 & 46.56 & 28.88 & 24.92 \\
VPP2 & WAM & 39.26 & 32.26 & 36.38 & 29.50 & 43.03 & 35.17 & 56.28 & 37.00 & 56.01 & 55.56 & 4.61 & 4.08 \\
WALL-OSS-1.0-lite & VLA & 42.59 & 34.97 & 46.66 & 37.94 & 38.16 & 29.08 & 49.94 & 32.25 & 57.42 & 56.00 & 20.76 & 19.58 \\
WorldScape-R$^2$SI & Agent + WAM & 46.58 & 39.35 & 42.95 & 35.83 & 23.51 & 15.50 & 51.48 & 34.25 & 65.78 & 63.67 & 49.16 & 47.50 \\
GO-3 Nano Preview & Agent + WAM & \underline{61.90} & \underline{56.42} & 36.45 & 33.66 & \textbf{78.65} & \textbf{76.00} & \textbf{68.21} & \textbf{53.50} & 64.50 & 63.67 & \underline{61.71} & 55.25 \\
\midrule
\multicolumn{14}{l}{\emph{Agent methods, acting through tools or code}} \\
RoboProbe$^\dagger$ & Agent & 28.97 & 22.48 & 33.36 & 30.50 & 12.65 & 4.00 & 21.45 & 8.25 & 43.04 & 38.67 & 34.36 & 31.00 \\
EmbodiedEvo$^\dagger$ & Agent & 45.08 & 38.95 & 35.22 & 30.83 & 25.15 & 16.25 & 33.83 & 21.75 & 79.40 & 76.67 & 51.81 & 49.25 \\
HexaShell-Preview$^\dagger$ & Agent & 53.18 & 46.85 & 39.22 & 34.50 & 41.25 & 32.00 & 42.91 & 31.50 & \underline{84.53} & \underline{81.00} & 57.97 & 55.25 \\
PyRUA-Lean$^\dagger$ & Agent & 55.26 & 50.35 & \underline{48.27} & \underline{43.17} & 44.77 & 38.25 & 54.30 & 43.75 & 68.40 & 68.33 & 60.56 & \underline{58.25} \\
\ours \method{}$^\dagger$ & Agent & \textbf{64.00} & \textbf{57.74} & \textbf{52.90} & \textbf{48.89} & \underline{46.30} & \underline{38.50} & \underline{62.10} & \underline{45.24} & \textbf{93.40} & \textbf{92.99} & \textbf{65.40} & \textbf{63.08} \\
\bottomrule
\end{tabular}
\end{table}

\textbf{By dimension.}
Across both blocks of \cref{tab:main}, \method has the highest Score and SR in Generalization, Memory, and Open, ahead of the next entry by 5.72, 11.99, and 4.83 percentage points in SR.
Precision and Long-Horizon remain the most challenging dimensions for \method, at 38.50\% and 45.24\% SR, against 76.00\% and 53.50\% for GO-3 Nano Preview, which leads both, though \method ranks second overall in both Score and SR across Precision (46.30 and 38.50\%) and Long-Horizon (62.10 and 45.24\%).

\begin{figure}[!htb]
\centering
\includegraphics[width=0.95\linewidth]{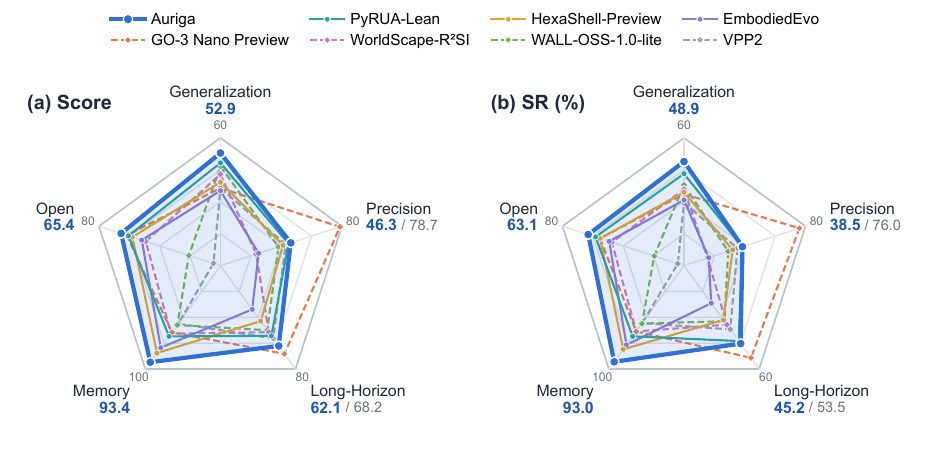}
\caption{Five-dimension profile of \method and seven other methods of \cref{tab:main}, in Score (a) and SR in \% (b). Each axis runs from zero at the center to the value at its outer vertex, with four evenly spaced rings. Below each dimension name is the value of \method, followed after a slash by the highest value of the other methods where it is higher. \method is the thick blue line with the shaded area; the other agent methods are drawn with solid lines and the WAM / VLA methods with dashed lines.}
\label{fig:agents-radar}
\end{figure}

\textbf{Profile against other methods.}
\Cref{fig:agents-radar} compares the five-dimension profile of \method with seven other methods of \cref{tab:main}: the four highest-scoring entries of the WAM / VLA block (GO-3 Nano Preview, WorldScape-R$^2$SI, WALL-OSS-1.0-lite, and VPP2) and the three highest-scoring other entries of the agent block (PyRUA-Lean, HexaShell-Preview, and EmbodiedEvo); \cref{fig:agents} in \cref{app:more} plots the same values as bars, together with RoboProbe.
GO-3 Nano Preview has the opposite profile to \method: its Score exceeds that of \method by 32.35 in Precision and 6.11 in Long-Horizon, while falling below \method in the other three dimensions, most in Memory (64.50 vs.\ 93.40) and Generalization (36.45 vs.\ 52.90).
Against PyRUA-Lean, \method achieves a higher Score in all five dimensions, with gains of 1.53 in Precision, 4.63 in Generalization, 4.84 in Open, 7.80 in Long-Horizon, and 25.00 in Memory.

\textbf{Per task.}
\Cref{tab:pertask,fig:pertask-h2h} in \cref{app:pertask} compare \method task by task with PyRUA-Lean, HexaShell-Preview, EmbodiedEvo, and RoboProbe.
Against each of the four, \method is higher in Score on more of the 42 tasks than it is lower, from 24 vs.\ 17 tasks against PyRUA-Lean to 41 vs.\ 1 against RoboProbe.
In SR, it is likewise higher on more tasks against all four methods: PyRUA-Lean (22 vs.\ 16), HexaShell-Preview (28 vs.\ 9), EmbodiedEvo (32 vs.\ 6), and RoboProbe (38 vs.\ 1).

\begin{figure}[!htb]
\centering
\includegraphics[width=0.95\linewidth]{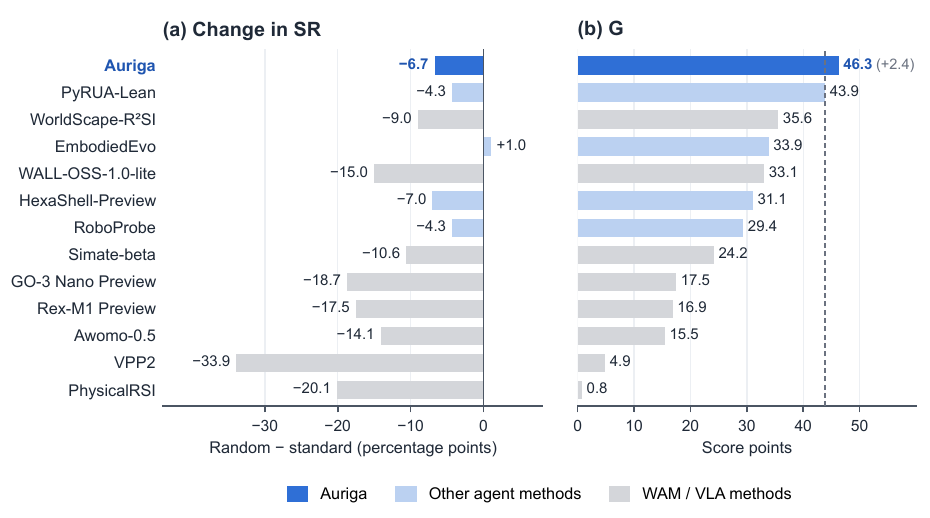}
\caption{Generalization of the entries of \cref{tab:main} on the 12 Generalization tasks under scene randomization, ordered by $G$. (a) Change in SR (percentage points, $\Delta\text{SR} = \text{SR}_{\mathrm{rand}} - \text{SR}_{\mathrm{std}}$) from standard to random scenes. (b) $G = \tfrac{1}{2}(S_{\mathrm{std}} + S_{\mathrm{rand}}) - |S_{\mathrm{rand}} - S_{\mathrm{std}}|$, the mean of the standard- and random-scene Scores minus their absolute difference.}
\label{fig:gen}
\end{figure}

\textbf{Generalization under randomization.}
\Cref{fig:gen} compares the entries of \cref{tab:main} on the 12 Generalization tasks across the standard and random scenes, reporting the SR change $\Delta\text{SR} = \text{SR}_{\mathrm{rand}} - \text{SR}_{\mathrm{std}}$ in panel~(a) and the composite generalization metric $G = \tfrac{1}{2}(S_{\mathrm{std}} + S_{\mathrm{rand}}) - |S_{\mathrm{rand}} - S_{\mathrm{std}}|$ in panel~(b), which rewards a high mean Score and penalizes the gap between the two scene types (\cref{tab:gen} in \cref{app:more} lists the SR and Score on both scene types).
\method achieves the highest $G$ in \cref{fig:gen} at 46.27, ahead of PyRUA-Lean at 43.87 and WorldScape-R$^2$SI at 35.61.
Across both scene types, \method achieves the highest SR and Score of all 13 entries: 52.28\% SR and 56.22 Score on the standard scenes, and 45.61\% SR and 49.58 Score on the random scenes, with a change of $-6.7$ percentage points in SR ($-13\%$ relative) and $-6.6$ points in Score (\cref{tab:gen}).
By contrast, the learned WAM and VLA entries suffer steep drops under scene randomization, losing up to $33.9$ percentage points in SR ($31.5$ points in Score) for VPP2 and $18.7$ percentage points ($-43\%$ relative) for GO-3 Nano Preview (\cref{fig:gen,tab:gen}).

\subsection{Main Results on RoboCasa365}
\label{sec:main-robocasa}

\begin{figure}[!htb]
\centering
\includegraphics[width=0.92\linewidth]{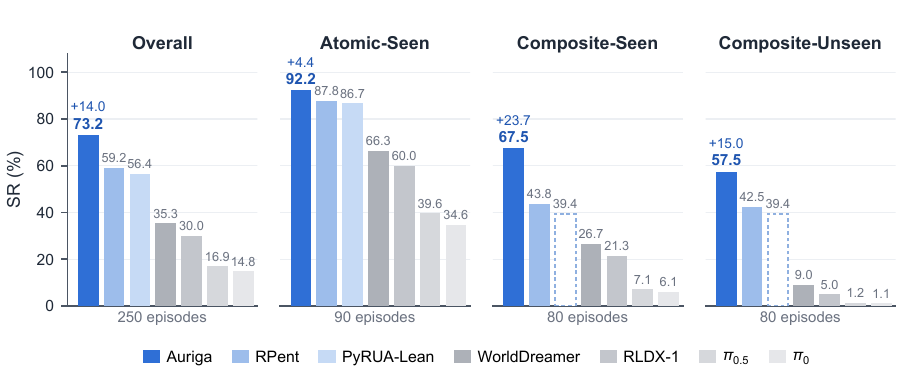}
\caption{Success rates (\%) of the methods of \cref{tab:robocasa-main} on RoboCasa365 Target50, in descending order of Overall SR. Grey bars are learned policies, and light blue bars are agent methods that hand contact-rich phases to the RLDX-1 policy. PyRUA-Lean reports composite tasks only as a whole, so its two composite bars are outlined and both show that rate. Above the bar of \method is its difference from the highest other method of the group.}
\label{fig:robocasa-main}
\end{figure}

RoboCasa365 shifts the evaluation to a mobile manipulator operating doors, drawers, knobs, and appliances across unseen kitchens over chains of up to fifteen subtasks (\cref{sec:exp-robocasa}).
\Cref{tab:robocasa-main,fig:robocasa-main} compare \method with the published agent baselines and with RLDX-1 on the 250 Target50 episodes (\cref{sec:exp-robocasa}).
All three agent methods use \code{gpt-6-astra} at high reasoning effort, and the baseline results are those reported by RPent~\citep{harnessvla2026} and PyRUA-Lean~\citep{pyrua2026}.
As in the comparison of RPent, \cref{tab:robocasa-main} also lists WorldDreamer, $\pi_{0.5}$, and $\pi_0$, which are evaluated in the pretraining kitchens; for these policies, Composite All is the mean of the two composite groups.
These policies, together with the leaderboard entry Xiaomi-Robotics-1, are also listed in \cref{tab:robocasa-board} in \cref{app:robocasa}.

\begin{table}[!htbp]
\centering
\footnotesize
\caption{Success rates (\%) on the 50 Target50 tasks of RoboCasa365~\citep{robocasa3652026}, at seeds 1--5 per task (250 episodes), ordered by ascending Overall SR within each evaluation setting. Overall is the mean over the 50 tasks, and Composite All covers the 160 composite episodes. Agent + VLA methods hand contact-rich phases to the RLDX-1 policy. WorldDreamer, $\pi_{0.5}$, and $\pi_0$ are evaluated in the pretraining kitchens, all other methods in the target kitchens. --: not reported. Bold: highest; underlined: second highest.}
\label{tab:robocasa-main}
\setlength{\tabcolsep}{5pt}
\begin{tabular}{llrrrrr}
\toprule
 & & & & \multicolumn{3}{c}{Composite} \\
\cmidrule(lr){5-7}
Method & Type & Overall & Atomic-Seen & Seen & Unseen & All \\
\midrule
\multicolumn{7}{l}{\emph{Pretraining kitchens: learned WAM or VLA policies}} \\
$\pi_{0}$ & VLA & 14.8 & 34.6 & 6.1 & 1.1 & 3.6 \\
$\pi_{0.5}$ & VLA & 16.9 & 39.6 & 7.1 & 1.2 & 4.2 \\
WorldDreamer & WAM & 35.3 & 66.3 & 26.7 & 9.0 & 17.9 \\
\midrule
\multicolumn{7}{l}{\emph{Target kitchens: VLA policy and agent methods}} \\
RLDX-1 & VLA & 30.0 & 60.0 & 21.3 & 5.0 & 13.1 \\
PyRUA-Lean & Agent + VLA & 56.4 & 86.7 & -- & -- & 39.4 \\
RPent & Agent + VLA & \underline{59.2} & \underline{87.8} & \underline{43.8} & \underline{42.5} & \underline{43.1} \\
\ours \method{} & Agent & \textbf{73.2} & \textbf{92.2} & \textbf{67.5} & \textbf{57.5} & \textbf{62.5} \\
\bottomrule
\end{tabular}
\end{table}

\textbf{Comparison with agent baselines.}
\method succeeds in 73.2\% of the 250 episodes, against 59.2\% for RPent and 56.4\% for PyRUA-Lean with the same model and reasoning effort, and 30.0\% for RLDX-1 alone.
It has the highest success rate in every group of \cref{tab:robocasa-main}.
The margin is largest on the composite tasks, which chain several subtasks: \method completes 62.5\% of these episodes, 19.4 percentage points more than RPent and 23.1 more than PyRUA-Lean.
Both baselines rely on the RLDX-1 policy for contact-rich phases, and RPent also on task memory gathered before evaluation; \method uses neither.

\textbf{Where \method gains and where it fails.}
The composite gap reflects how contact-rich phases are executed.
The baselines hand them to RLDX-1, which on its own completes 21.3\% of the Composite-Seen and 5.0\% of the Composite-Unseen tasks, while \method builds every phase, including grasps, placements, and fixture operations, from action programs whose stages are checked as they execute (\cref{sec:wysiwye}).
Of the 67 failed episodes of \method, 23 ended when the 100 model requests were used and 44 after the model stopped acting for several turns (\cref{fig:robocasa-pertask} in \cref{app:robocasa}).
Four composite tasks fail in all five episodes, PackIdenticalLunches and StoreLeftoversInBowl among the seen and MakeIceLemonade and SeparateFreezerRack among the unseen, while every Atomic-Seen task succeeds in at least three of its five episodes.

\subsection{Ablation Studies}
\label{sec:exp-ablation}

Whereas \cref{fig:harness_development} (\cref{sec:rsi_positioning}) tracked the cumulative gains across development iterations, \crefrange{tab:abl-measure}{tab:abl-programs} isolate the contribution of each mechanism of \cref{sec:method,sec:wysiwye,sec:augment} on the frozen configuration of \cref{sec:exp-config}.
Each table groups related mechanisms and evaluates its variants on ten tasks (\cref{tab:abl-tasks} in \cref{app:ablation}): \crefrange{tab:abl-measure}{tab:abl-action} on five standard layouts of each task (50 cases), and \cref{tab:abl-programs} on 30 cases.
Each variant removes or restricts the mechanisms marked in its row while keeping the rest of the configuration fixed, and the \emph{Full} row reports the main configuration on the same cases.
Alongside the number of successes and the mean Score, each table reports a mechanism-specific diagnostic per episode for the ablated variants (and for all rows in \cref{tab:abl-programs}), taken from the execution receipts $r_t$, the evidence record $E_t$, or the model calls (\cref{sec:formulation,sec:records}).

\textbf{Spatial measurement.}
\Cref{tab:abl-measure} evaluates the mechanisms that provide metric 3D positions (\cref{sec:aug-input}): the camera calibration and 3D object estimates, together with \code{geometry_query} for ray--plane intersection (\cref{eq:backproject}), two-view triangulation, and 3D-to-image projection.
In \emph{No geometry tools}, the calibration, the 3D object estimates, and \code{geometry_query} are all removed, so the model locates targets from the RGB images and the robot state alone (\cref{sec:aug-input}).
In \emph{No triangulation}, the calibration, the 3D object estimates, two-view triangulation, and 3D-to-image projection are removed while ray--plane intersection (\cref{eq:backproject}) remains available at one plane height per pixel.
Compared with \emph{Full} (46/50 successes, mean Score 94.9), \emph{No geometry tools} reaches 30/50 successes and a Score of 67.1, with 65.3\% of first grasps held.
With ray--plane intersection available, \emph{No triangulation} reaches 40/50 successes and a Score of 84.0, with 85.7\% of first grasps held; \emph{Full} adds two-view triangulation, 3D-to-image projection, the calibration, and the 3D object estimates (\cref{sec:aug-input}).

\begin{table}[!htbp]
\centering
\footnotesize
\caption{Spatial measurement (\cref{sec:aug-input,eq:backproject}): successes over the 50 cases, mean Score, and the share of first grasps that hold (\%). The highest successes and Score are in bold. \checkmark: available; $\times$: removed.}
\label{tab:abl-measure}
\renewcommand{\arraystretch}{1.15}
\begin{tabular}{lcccrrr}
\toprule
Variant & \shortstack{Plane\\intersection} & Triangulation & \shortstack{Calibration and\\object estimates} & Successes & Score & \shortstack{First grasps\\held (\%)} \\
\midrule
No geometry tools & $\times$ & $\times$ & $\times$ & 30/50 & 67.1 & 65.3 \\
No triangulation & \checkmark & $\times$ & $\times$ & 40/50 & 84.0 & 85.7 \\
\ours Full & \checkmark & \checkmark & \checkmark & \textbf{46/50} & \textbf{94.9} & -- \\
\bottomrule
\end{tabular}
\end{table}

\textbf{Collision handling.}
\Cref{tab:abl-collision} evaluates two mechanisms that keep execution clear of unintended contact (\cref{sec:preview,sec:verify}).
In \emph{No clearance review}, the preview no longer estimates the clearance between the arms, held objects, and declared obstacles before motion (\cref{sec:preview}).
In \emph{No divergence stop}, the stage loop in $\mathcal{A}$ no longer halts the sequence when a movement exceeds its arrival tolerance $\tau_k = (\delta_p, \delta_\theta)$, so later stages continue from the displaced pose (\cref{sec:loops,sec:verify}).
Compared with \emph{Full} (36/50 successes, Score 79.8), \emph{No clearance review} reaches 31/50 and 69.1, with 1.20 tolerance stops per episode; without the clearance review, the stage check during execution is the remaining guard against contact (\cref{sec:verify}).
Compared with \emph{Full}, \emph{No divergence stop} reaches 34/50 and 75.1, with 0.00 stops and 0.92 out-of-tolerance movements per episode as sequences run on, whereas stage verification in \emph{Full} halts at the first exceeded tolerance and returns a receipt $r_t$ and fresh view (\cref{sec:loops,sec:verify}).

\begin{table}[!htbp]
\centering
\footnotesize
\caption{Collision handling (\cref{sec:preview,sec:verify}): successes over the 50 cases, mean Score, and the stops per episode at a movement that exceeded its arrival tolerance ($\delta_p$, $\delta_\theta$). The highest successes and Score are in bold. \checkmark: available; $\times$: removed.}
\label{tab:abl-collision}
\renewcommand{\arraystretch}{1.15}
\begin{tabular}{lccrrr}
\toprule
Variant & \shortstack{Clearance\\review} & \shortstack{Divergence\\stop} & Successes & Score & \shortstack{Stops\\per episode} \\
\midrule
No clearance review & $\times$ & \checkmark & 31/50 & 69.1 & 1.20 \\
No divergence stop & \checkmark & $\times$ & 34/50 & 75.1 & 0.00 \\
\ours Full & \checkmark & \checkmark & \textbf{36/50} & \textbf{79.8} & -- \\
\bottomrule
\end{tabular}
\end{table}

\textbf{Preview and checkpoints.}
\Cref{tab:abl-checkpoint} evaluates two mechanisms of the admission operator $\mathcal{A}$ (\cref{sec:single_path,sec:wysiwye}): previewing a trajectory before execution (\cref{sec:preview}) and pausing only at model-placed checkpoints (\cref{sec:verify}).
In \emph{No preview}, an action request is planned and executed in one call, so the model sees the trajectory only after it has run (\cref{sec:preview}).
In \emph{Pause after every gripper action}, $\mathcal{A}$ inserts a checkpoint after every gripper stage that a later movement depends on, rather than only where the model placed one (\cref{sec:verify}).
Compared with \emph{Full} (42/50 successes, Score 88.5), \emph{No preview} reaches 36/50 and 74.7, with 0.96 checkpoint pauses and 4.31 unreviewed motions per episode, 0.98 of which carry a clearance risk that preview reports before motion (\cref{sec:preview}).
Compared with \emph{Full}, \emph{Pause after every gripper action} reaches 37/50 and 78.7 with 13.82 pauses per episode, whereas model-placed checkpoints in \emph{Full} pause only at stages selected for inspection (\cref{sec:verify}).

\begin{table}[!htbp]
\centering
\footnotesize
\caption{Preview and checkpoints in $\mathcal{A}$ (\cref{sec:single_path,sec:preview,sec:verify}): successes over the 50 cases, mean Score, and the checkpoint pauses per episode. The highest successes and Score are in bold. \checkmark: available; $\times$: removed.}
\label{tab:abl-checkpoint}
\renewcommand{\arraystretch}{1.15}
\begin{tabular}{lccrrr}
\toprule
Variant & Preview & Checkpoints & Successes & Score & \shortstack{Pauses\\per episode} \\
\midrule
No preview & $\times$ & placed by the model & 36/50 & 74.7 & 0.96 \\
Pause after every gripper action & \checkmark & after every gripper action & 37/50 & 78.7 & 13.82 \\
\ours Full & \checkmark & placed by the model & \textbf{42/50} & \textbf{88.5} & -- \\
\bottomrule
\end{tabular}
\end{table}

\textbf{Expressive actions.}
\Cref{tab:abl-action} evaluates the two output-side mechanisms of \cref{sec:aug-output}: finger-referenced movement goals and the shared dual-arm timeline.
In \emph{Wrist-flange goals}, each movement specifies the wrist-flange position and quaternion rather than a finger contact point and workspace rotations (\cref{sec:aug-output}), and the overlay that projects the finger positions into the image (\cref{sec:aug-input}) is removed.
In \emph{One arm at a time}, an action request may contain stages for only one arm, so the two arms no longer advance on a shared timeline (\cref{sec:aug-output}).
Compared with \emph{Full} (39/50 successes, Score 83.9), \emph{Wrist-flange goals} reaches 37/50 and 77.1 with 3.42 unplannable stages per episode, whereas finger-referenced goals in \emph{Full} let the model specify the finger contact region and workspace orientation directly (\cref{sec:aug-output}).
Compared with \emph{Full}, \emph{One arm at a time} reaches 35/50 and 74.8 with 1.14 unplannable stages per episode, whereas the shared timeline in \emph{Full} coordinates both arms with declared waits in one sequence (\cref{sec:aug-output}).

\begin{table}[!htbp]
\centering
\footnotesize
\caption{Expressive actions (\cref{sec:aug-output}): successes over the 50 cases, mean Score, and the planning failures per episode (requested stages that could not be planned). The highest successes and Score are in bold. \checkmark: available; $\times$: removed.}
\label{tab:abl-action}
\renewcommand{\arraystretch}{1.15}
\begin{tabular}{lccrrr}
\toprule
Variant & \shortstack{Finger-referenced\\goals} & \shortstack{Shared\\timeline} & Successes & Score & \shortstack{Planning failures\\per episode} \\
\midrule
One arm at a time & \checkmark & $\times$ & 35/50 & 74.8 & 1.14 \\
Wrist-flange goals & $\times$ & \checkmark & 37/50 & 77.1 & 3.42 \\
\ours Full & \checkmark & \checkmark & \textbf{39/50} & \textbf{83.9} & -- \\
\bottomrule
\end{tabular}
\end{table}

\textbf{Programs.}
\Cref{tab:abl-programs} evaluates how model-written programs interact with $\mathcal{A}$ (\cref{sec:single_path,sec:preview}): in \emph{Full}, a program computes and calls deliberative tools ($\mathcal{U}_{\mathrm{d}}$) freely, while any physical request ($\mathcal{U}_{\mathrm{p}}$) enters $\mathcal{A}$, waits for the model to review its preview (\cref{sec:preview}), and returns a receipt $r_t$ (\cref{sec:formulation,sec:records}).
In \emph{No programs}, the Compute capability (\code{program_run}, \cref{tab:capabilities}) is removed, so the model issues every call directly.
In \emph{Programs cannot act}, programs still run for computation and deliberative queries, but $\mathcal{A}$ rejects their physical requests, so the model issues every physical action directly (\cref{sec:single_path}).
In \emph{Code as policy}, physical requests from a program run without preview review and return no receipts $r_t$ to the program or the model, leaving only the program's output and exit status, as in the paradigm contrasted with \method in \cref{sec:gap,sec:records}.
Compared with \emph{Full} (30/30 successes, Score 100.0, 46.2 model calls per episode), \emph{No programs} reaches 19/30 and 75.3 with 66.0 calls per episode (one episode exhausting the 170-response limit), and \emph{Programs cannot act} reaches 19/30 and 73.0 with 67.1 calls per episode; in \emph{Full}, programs issue 28.8 tool calls per episode while the model executes 5.9 physical actions directly (\cref{sec:single_path}).
Compared with \emph{Full}, \emph{Code as policy} reaches 20/30 and 77.2 with 71.8 calls per episode.
Its programs issue 120 physical actions over the 30 episodes, of which 2 halt when a movement exceeds its arrival tolerance and 1 stops at a checkpoint; without receipts $r_t$, each of these programs still exits with code 0, one of them after issuing 8 further actions (\cref{sec:verify,sec:records}).

\begin{table}[!htbp]
\centering
\footnotesize
\caption{Programs (\cref{sec:single_path,sec:preview,sec:gap}): successes over the 30 cases, mean Score, and the model calls per episode. The highest successes and Score are in bold. \checkmark: available; $\times$: removed.}
\label{tab:abl-programs}
\renewcommand{\arraystretch}{1.15}
\begin{tabular}{lcccrrr}
\toprule
Variant & Programs & \shortstack{Physical actions\\from programs} & \shortstack{Review and\\receipts} & Successes & Score & \shortstack{Model calls\\per episode} \\
\midrule
Programs cannot act & \checkmark & $\times$ & \checkmark & 19/30 & 73.0 & 67.1 \\
No programs & $\times$ & $\times$ & \checkmark & 19/30 & 75.3 & 66.0 \\
Code as policy & \checkmark & \checkmark & $\times$ & 20/30 & 77.2 & 71.8 \\
\ours Full & \checkmark & \checkmark & \checkmark & \textbf{30/30} & \textbf{100.0} & 46.2 \\
\bottomrule
\end{tabular}
\end{table}

\subsection{Execution Efficiency and Tool Usage}
\label{sec:main-cost}
\label{sec:exp-tools}

\begin{figure}[!htb]
\centering
\begin{minipage}[t]{0.485\linewidth}
\vspace{0pt}
\centering
\parbox[t][0.8254\linewidth][c]{\linewidth}{\centering\footnotesize
\setlength{\tabcolsep}{3.2pt}
\renewcommand{\arraystretch}{1.12}
\begin{tabular}{@{}lrrrr@{}}
\toprule
 & \method & \shortstack[r]{PyRUA-\\Lean} & \shortstack[r]{HexaShell-\\Preview} & \shortstack[r]{Embodied-\\Evo} \\
\midrule
Reasoning effort & high & high & -- & xhigh \\
Score & \textbf{64.00} & 55.26 & 53.18 & 45.08 \\
\midrule
Model calls & 47.5 & 25.7 & -- & -- \\
Input (M) & 5.63 & 0.75 & -- & 3.94 \\
Cached (\%) & 85.4 & -- & -- & 97.6 \\
Uncached (M) & 0.82 & -- & -- & 0.09$^\ddagger$ \\
Output (k) & 18.3 & 9.0 & -- & 19$^\ddagger$ \\
Time (min) & 18.8 & 12.7 & 12.3$^\S$ & -- \\
\bottomrule
\end{tabular}}
\captionof{table}{Cost per RoboDojo episode of \method (over the 6,299 episodes in \cref{tab:cost-dim}) and of the three methods in \cref{tab:main} that report it, with each method's average Score from \cref{tab:main}; the quantities are defined in \cref{sec:exp-metrics}. All four use \code{gpt-6-astra}; a dash marks a quantity the method does not report. $^\ddagger$Derived from reported totals or shares. $^\S$Median.}
\label{tab:cost-agents}
\end{minipage}\hfill
\begin{minipage}[t]{0.485\linewidth}
\vspace{0pt}
\centering
\includegraphics[width=\linewidth]{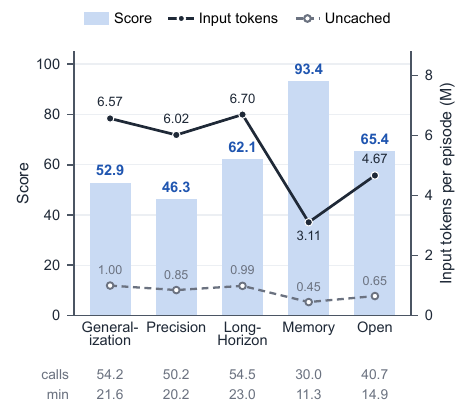}
\caption{Five-dimension Score (\cref{tab:main}) and per-episode cost of \method by dimension (over the 6,299 episodes in \cref{tab:cost-dim}): five-dimension Score (bars, left axis), mean input tokens per episode in millions (solid line, right axis), and uncached input tokens in millions (dashed line, right axis). Below the axis: model calls and wall-clock minutes per episode.}
\label{fig:cost-dim}
\end{minipage}
\end{figure}

\textbf{Episode cost across methods and dimensions.}
Of the other methods in \cref{tab:main}, PyRUA-Lean~\citep{pyrua2026}, HexaShell-Preview~\citep{hexashell2026}, and EmbodiedEvo~\citep{embodiedevo2026} report the test-episode cost of their RoboDojo runs.
\Cref{tab:cost-agents} lists the cost per test episode of these three methods and of \method (measured over the 6,299 scored episodes of \cref{tab:cost-dim}), together with each method's overall average Score in \cref{tab:main}.
Of the methods that report input tokens, \method achieves the highest average Score (64.00) and uses 5.63 million input tokens per episode, of which 85.4\% are served from the prompt cache, leaving 0.82 million uncached input tokens per episode (against 3.94 million total and 0.09 million uncached for EmbodiedEvo at 45.08 Score, and 0.75 million total for PyRUA-Lean at 55.26 Score).
An \method episode averages 47.5 model calls and 18.8 minutes, compared with 25.7 calls and 12.7 minutes for PyRUA-Lean and a median of 12.3 minutes for HexaShell-Preview.
Note that cross-trial evolution methods (\cref{tab:evaluation_conditions}) also incur pre-evaluation trial costs that are not included in their per-test-episode numbers, whereas \method runs entirely within the evaluated episode.

\Cref{fig:cost-dim} compares the five-dimension Score of \method (\cref{tab:main}) against the mean input tokens and uncached input tokens per episode across dimensions (\cref{tab:cost-dim} in \cref{app:more}).
Memory, the dimension with the highest Score (93.40), is also the cheapest, with 30.0 model calls, 3.11 million input tokens (0.45 million uncached), and 11.3 minutes per episode, and Open, with the second-highest Score (65.40), is the second cheapest, with 40.7 model calls, 4.67 million input tokens (0.65 million uncached), and 14.9 minutes.
Long-Horizon and Generalization episodes are the most expensive, with 54.5 and 54.2 model calls, 6.70 and 6.57 million input tokens, and 23.0 and 21.6 minutes, and across all 6,299 episodes \method uses 79.7\% of the native-step limit on average (\cref{tab:cost-dim}).

\begin{figure}[!htb]
\centering
\begin{minipage}[t]{0.485\linewidth}
\vspace{0pt}
\centering
\includegraphics[width=\linewidth]{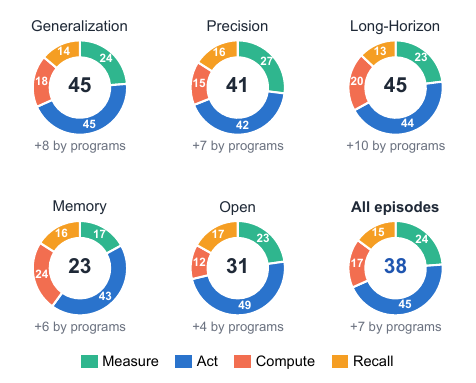}
\caption{Tool calls per episode by dimension (6,299 episodes): mean number of direct tool calls (center), share of each capability in \% (ring segments), and mean number of additional calls made by programs (below each ring).}
\label{fig:tool-kinds}
\end{minipage}\hfill
\begin{minipage}[t]{0.485\linewidth}
\vspace{0pt}
\centering
\includegraphics[width=\linewidth]{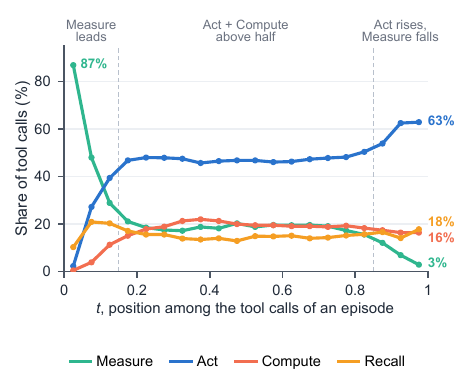}
\caption{Share of each capability (\%) along an episode: the equal-weight mean across episodes (6,298 episodes). The tool calls of each episode are spread in order over $t \in [0, 1]$ and split into 20 equal segments.}
\label{fig:tool-progress}
\end{minipage}
\end{figure}

\textbf{Tool usage across capabilities and episode progress.}
\Cref{fig:tool-kinds,fig:tool-progress} group the tool calls of the model by the four capabilities of \cref{tab:capabilities}: Measure, Act, Compute, and Recall.
An episode makes 38 direct tool calls on average, from 45 in Long-Horizon and Generalization episodes to 23 in Memory episodes (\cref{fig:tool-kinds}).
Act takes the largest share (45\%), followed by Measure (24\%), Compute (17\%), and Recall (15\%).
Within each capability, spatial measurement (\code{geometry_query}) accounts for 16\% of all direct tool calls and scene observation (\code{robot_observe}, alongside \code{stereo_guide}) for 8\% under Measure; under Act, trajectory preview ($\mathcal{U}_{\mathrm{d}}$, \cref{eq:deliberative}) accounts for 19\% and physical execution ($\mathcal{U}_{\mathrm{p}}$, \cref{eq:physical}) for 24\% in \code{robot_action} (with \code{robot_wait} at 2\%), so 44\% of all \code{robot_action} calls preview a candidate sequence without advancing the world; under Compute, running programs (\code{program_run}) accounts for 17\% (with programs issuing 7 additional tool calls per episode on average, rising to 10 in Long-Horizon); and under Recall, updating the task belief (\code{task_plan}, \cref{sec:belief}) accounts for 13\% while querying evidence records (\code{history_*} and \code{operation_memory}, \cref{sec:records}) accounts for 2\%.
Across normalized episode progress $t \in [0, 1]$ (\cref{fig:tool-progress}), Measure takes 87\% of calls before $t = 0.05$ as the model inspects the scene and localizes targets before acting; from $t = 0.15$ onward, Act (46\%--54\%) and Compute (17\%--22\%) account for over 60\% of calls while Recall stays between 13\% and 17\%.

\textbf{Execution granularity against the reference harness.}
\label{sec:analysis_baseline}
To compare model interaction and physical execution granularity on matched zero-shot scenes, \cref{tab:analysis_baseline} evaluates \method against the published RoboProbe release~\citep{unexpected2026} across the 15 standard-variant scenes of \code{swap_T} (full step-by-step execution traces and mechanism statistics are detailed in \cref{app:trace}).
\method completes 14 of the 15 scenes (93.3\% SR), compared with 3 of 15 (20.0\% SR) for RoboProbe ($+73.3$ percentage points), while using fewer model calls per episode (25.4 vs.\ 36.6 mean, 24 vs.\ 38 median).
Across the 15 episodes, RoboProbe issues 501 physical execution batches (33.4 per episode, 10.9 native steps per batch), whereas \method issues 47 batches (3.1 per episode, 107.3 native steps per batch)---a 90.6\% reduction in physical execution invocations for a comparable number of native steps (336.2 vs.\ 365.1), because approach, grasp, transport, and release are previewed and verified as multi-stage operations rather than single end-effector steps.

\begin{table}[!htbp]
  \centering
  \small
  \setlength{\tabcolsep}{3pt}
  \renewcommand{\arraystretch}{1.15}
  \caption{Task outcomes and execution profiles on 15 matched standard-variant T-block exchange (\code{swap_T}) scenes (detailed trace analysis in \cref{app:trace}).}
  \label{tab:analysis_baseline}
  \begin{tabular*}{\linewidth}{@{\extracolsep{\fill}}ccccccccc@{}}
    \toprule
    & \multicolumn{3}{c}{Task completion}
    & \multicolumn{2}{c}{Model calls / ep.}
    & \multicolumn{3}{c}{Physical execution} \\
    \cmidrule(lr){2-4}\cmidrule(lr){5-6}\cmidrule(l){7-9}
    System & Successes & SR (\%) & Exclusive & Mean & Median
    & \shortstack[c]{Batches / ep.}
    & \shortstack[c]{Steps /ep.}
    & \shortstack[c]{Steps / batch} \\
    \midrule
    RoboProbe~\citep{unexpected2026} & 3/15 & 20.0 & 0 & 36.6 & 38 & 33.4 & 365.1 & 10.9 \\
    \ours \method & 14/15 & \textbf{93.3} & \textbf{11} & 25.4 & 24
    & 3.1 & 336.2 & 107.3 \\
    \bottomrule
  \end{tabular*}
  \par\smallskip
  \begin{minipage}{\linewidth}
    \footnotesize\raggedright
    SR: success rate; Exclusive: scenes solved only by that system. Batches count action invocations that advance the environment, including early stops; steps are native actions, and steps/batch is their aggregate ratio.
  \end{minipage}
\end{table}

%% file: sec_exp_qualitative.tex
\subsection{Qualitative Analysis and Failure Modes}
\label{sec:cases}

We complement the aggregate metrics and ablations with qualitative episode traces across both RoboDojo and RoboCasa365, organized around four operational challenges: recovery after grasp failure and transport slippage (\cref{sec:case_regrasp}), scope-matched replanning and budget-aware preview (\cref{sec:case_replanning}), 3D geometry-guided manipulation with VGGT-Omega and SAM~3.1 (\cref{sec:case_vggt_sam}), and failure modes when repeated corrections under multi-body coupling exhaust the episode budget (\cref{sec:case_failure}).
Full step-by-step trace tables on \code{swap_T} and additional case studies are provided in \cref{app:trace,app:cases}.

\begin{figure}[!htbp]
  \centering
  \begin{subfigure}{0.90\linewidth}
    \centering
    \includegraphics[width=\linewidth]{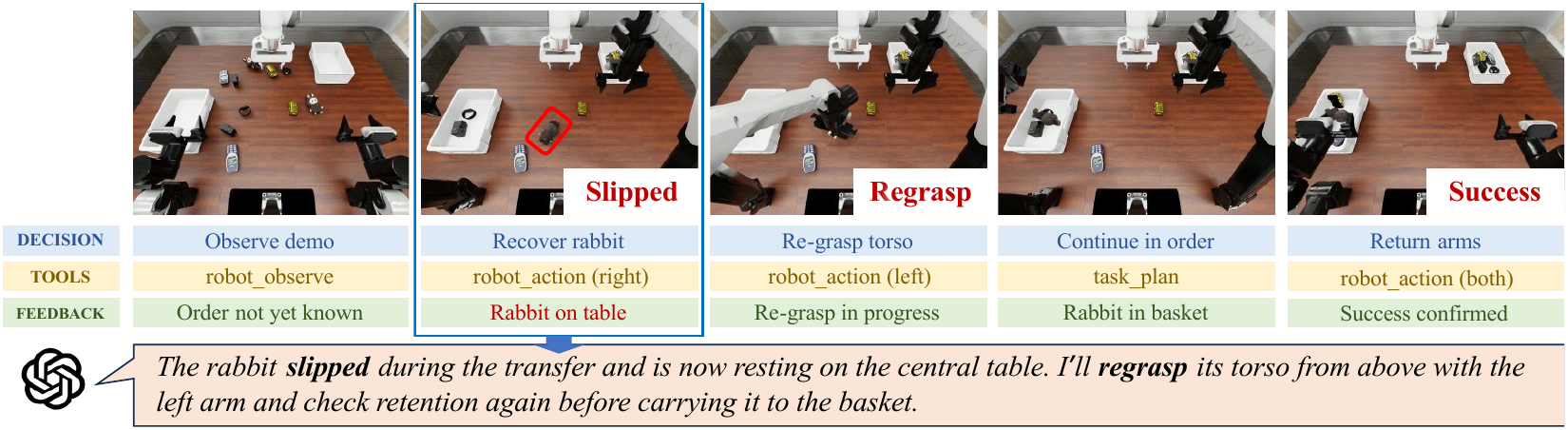}
    \caption{RoboDojo sequence reproduction (\code{imitate_sorting_sequence}): preserving sequence order in $B_t$ and switching from a rim-blocked head grasp to a cross-torso grasp after a handover slip.}
    \label{fig:regrasp_sequence}
  \end{subfigure}
  \par\vspace{2pt}
  \begin{subfigure}{0.90\linewidth}
    \centering
    \includegraphics[width=\linewidth]{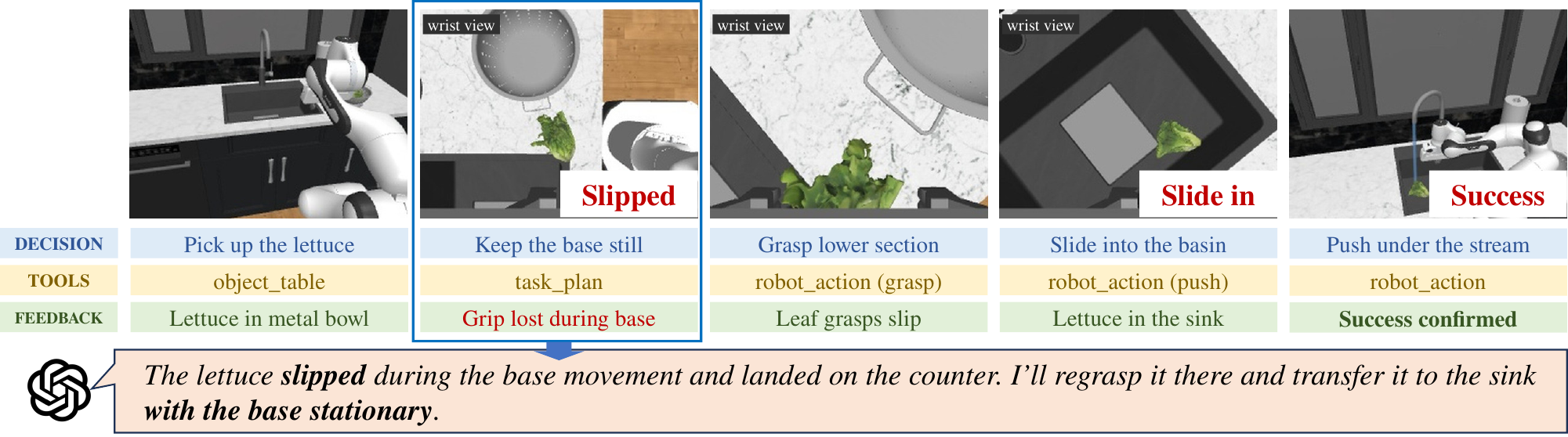}
    \caption{RoboCasa365 lettuce washing (\code{WashLettuce}): keeping the mobile base still and replacing an aerial lift with a counter-supported slide into the sink after transport slippage.}
    \label{fig:regrasp_lettuce}
  \end{subfigure}
  \caption{\textbf{Recovery from grasp failure and transport slippage across RoboDojo and RoboCasa365.} Rather than repeating the failed command or discarding prior progress, the model updates the contradicted retention condition in $B_t$ and revises the contact region, active arm, or transport mode.}
  \label{fig:regrasp_main}
\end{figure}

\Needspace{2\baselineskip}
\textbf{Recovery from grasp failure, transport slippage, and arm handover.}
\label{sec:case_regrasp}
Grasp acquisition, retention during transport, and stable support after release are tracked as distinct conditions in the task belief $B_t$, allowing the model to revise how an object is grasped or transported without losing already verified subgoals (\cref{fig:regrasp_main}).

In RoboDojo \code{imitate_sorting_sequence} (\cref{fig:regrasp_sequence}), the model must reproduce a five-object placement order (wristwatch, black device, rabbit, phone, gold vehicle).
After the first two objects are verified inside the basket, the rabbit lies outside the left arm's reach, so the right arm transfers it toward the center.
When the rabbit slips during transfer and lands outside the basket's far rim, the task belief keeps the third slot marked \code{pending} rather than advancing to the phone; after a head grasp catches the rim and triggers a \code{tracking_error} stop, the model inspects the wrist view, rotates the jaws across the rabbit's torso in a clearer region, verifies placement inside the basket, and only then proceeds to the remaining two objects.
Similarly, in T-block pose exchange (\code{swap_T}, \cref{tab:analysis_recovery} in \cref{app:trace}), when the blue block slips onto the table during a right-arm transfer whose robot motion nonetheless reports \code{completed} with zero final gripper aperture, the model marks blue retention as \code{contradicted} while preserving the already placed pink block as \code{supported}, switches recovery to the nearby left arm, and completes the exchange to the saved initial pose.

In RoboCasa365 \code{WashLettuce} (\cref{fig:regrasp_lettuce}), transport slippage prompts the model to change the transport mode altogether.
After the initial grasp, moving the mobile base causes the head of lettuce to slip onto the counter, as flagged by the execution receipt and wrist-camera object table.
When holding the mobile base stationary still leaves three upper-leaf re-grasps blocked or slipping, the model targets a narrower lower section of the lettuce beside the sink and replaces an aerial transfer with a counter-supported slide directly into the basin, before turning on the faucet and pushing the lettuce under the water stream to complete the task (additional slippage recovery cases on \code{pick_from_conveyor_by_image}, \code{stack_bowls}, and \code{PickPlaceDrawerToCounter} are analyzed in \cref{app:cases}).

\begin{figure}[!htbp]
  \centering
  \begin{subfigure}{0.90\linewidth}
    \centering
    \includegraphics[width=\linewidth]{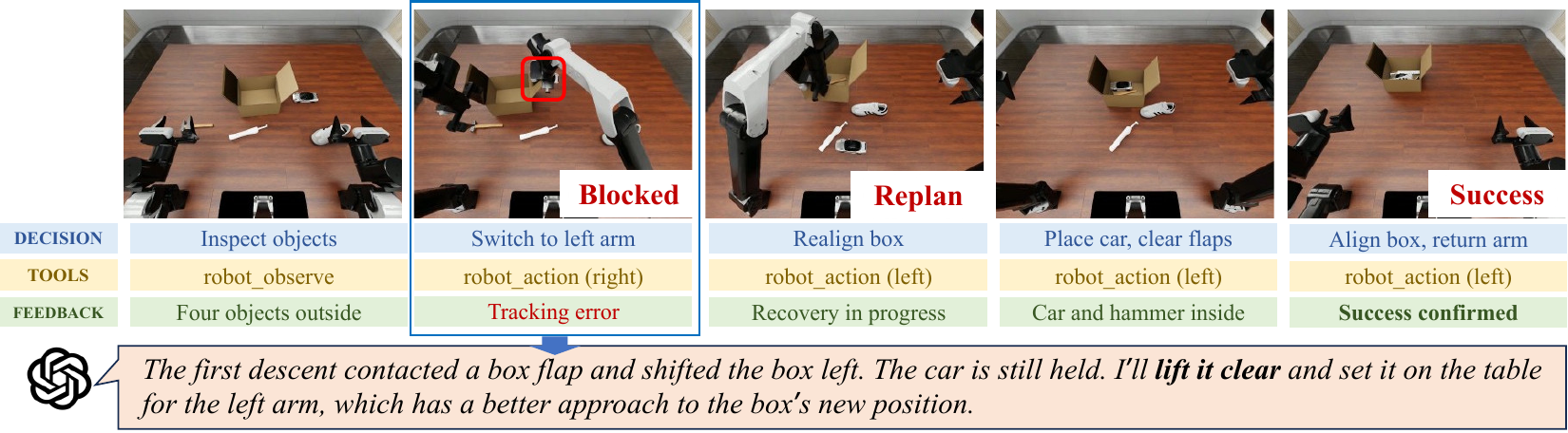}
    \caption{RoboDojo object packing (\code{pack_objects_into_box}): restoring a box displaced by flap contact to the central workspace and revising descent clearances.}
    \label{fig:replanning_packing}
  \end{subfigure}
  \par\vspace{2pt}
  \begin{subfigure}{0.90\linewidth}
    \centering
    \includegraphics[width=\linewidth]{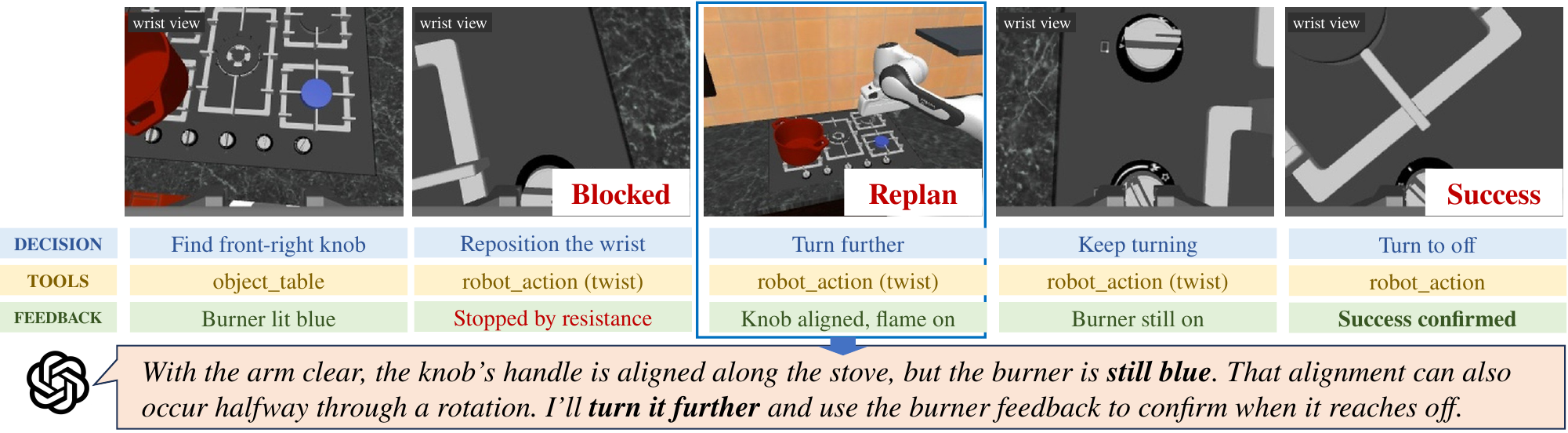}
    \caption{RoboCasa365 burner shutoff (\code{TurnOffStove}): re-posturing the wrist when mechanical resistance halts rotation, then continuing past a deceptive halfway handle alignment while the flame remains blue.}
    \label{fig:replanning_stove}
  \end{subfigure}
  \caption{\textbf{Scope-matched replanning after contact and state discrepancies.} Each revision addresses the specific assumption contradicted by the execution receipt or visual state---restoring a displaced container, re-posturing the wrist for rotational range, or overriding an ambiguous handle angle.}
  \label{fig:replanning_main}
\end{figure}

\Needspace{2\baselineskip}
\textbf{Scope-matched replanning and budget-aware trajectory preview.}
\label{sec:case_replanning}
When an execution receipt and post-action observation invalidate a physical assumption, the model matches the scope of its revision to the contradicted premise (\cref{fig:replanning_main}).

In RoboDojo \code{pack_objects_into_box} (\cref{fig:replanning_packing}), the robot must pack a car, shoe, hammer, and toothbrush into a box with their fronts facing left.
During the first car descent, wrist contact with a box flap stops the arm with an $8.2\,\mathrm{cm}$ vertical residual (\code{tracking_error}) and shifts the box across the table.
Rather than repeatedly targeting coordinates measured before contact, the model treats the displaced container as state that must be restored: it sets down the held car, repositions the box to the central workspace, and revises subsequent trajectories to keep the wrists above the flaps with vertical descents and withdrawals.

\begin{figure}[!bp]
  \centering
  \begin{subfigure}{0.90\linewidth}
    \centering
    \includegraphics[trim=0 0 24.5bp 0,clip,width=\linewidth]{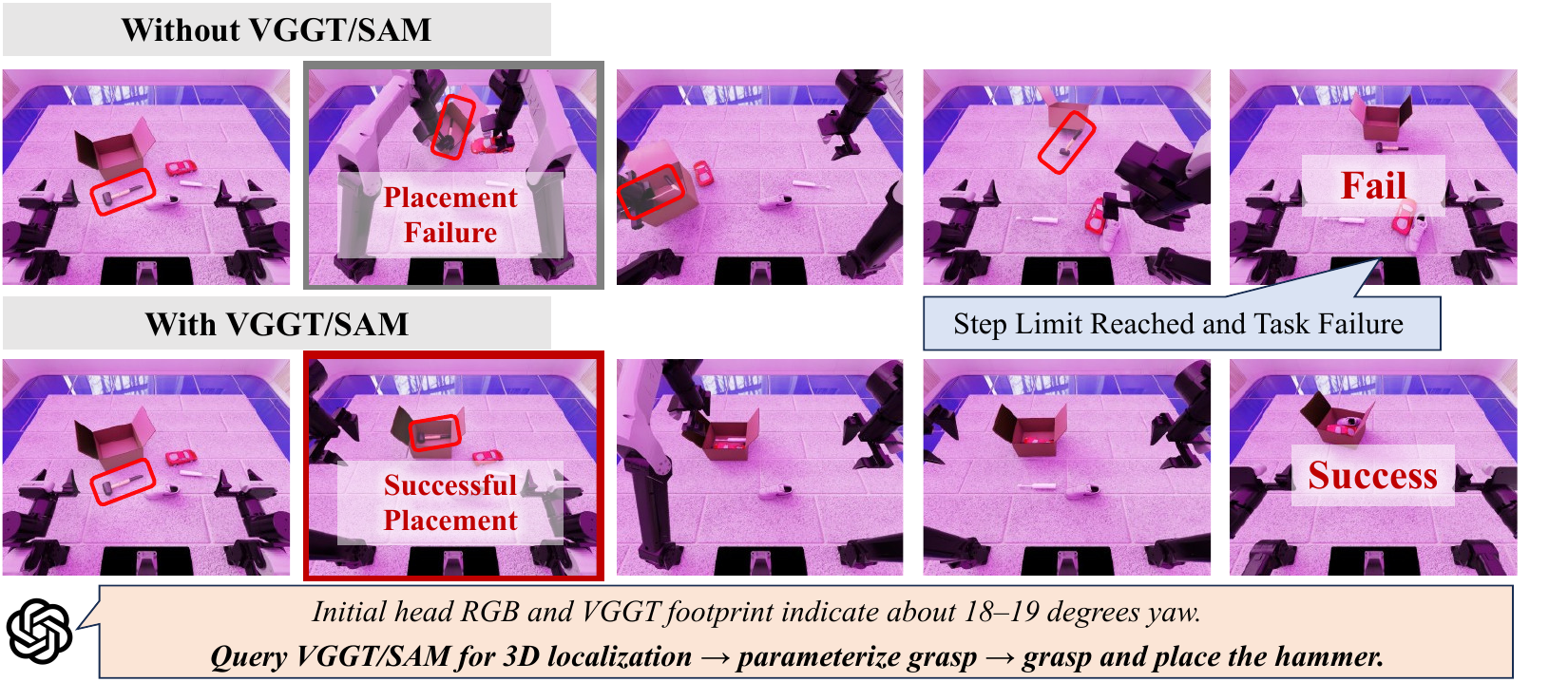}
    \caption{RoboDojo object packing (\code{pack_objects_into_box}, randomized scene) without (top) and with (bottom) VGGT-Omega and SAM~3.1.}
    \label{fig:case_vggt_sam}
  \end{subfigure}
  \par\vspace{2pt}
  \begin{subfigure}{0.90\linewidth}
    \centering
    \includegraphics[width=\linewidth]{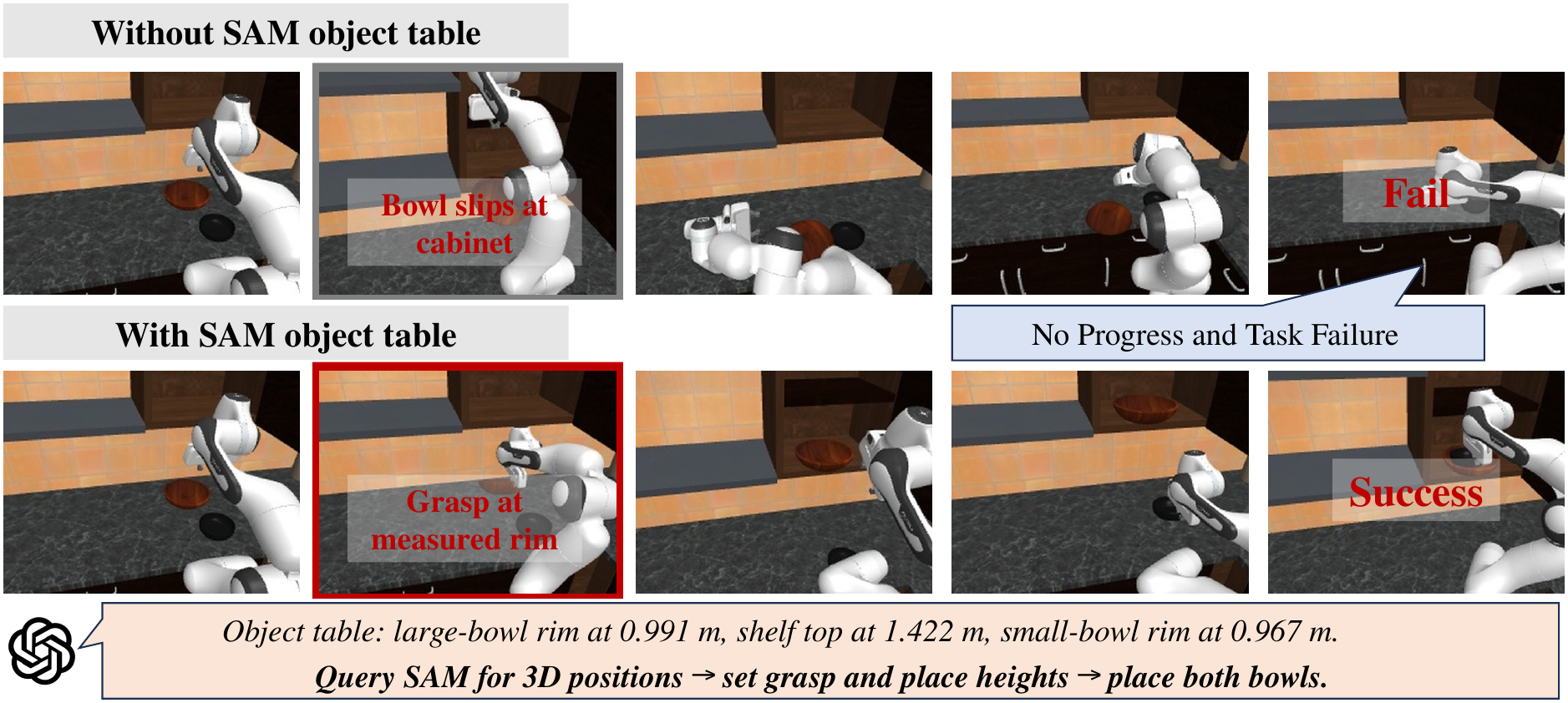}
    \caption{RoboCasa365 cabinet loading (\code{StackBowlsCabinet}) without (top) and with (bottom) the SAM~3.1 3D object table.}
    \label{fig:case_sam_robocasa}
  \end{subfigure}
  \caption{\textbf{3D geometry-guided manipulation on matched scenes.} Estimated 3D bounding boxes, orientations, and surface heights guide container alignment, below-rim grasps, and shelf clearance across both benchmarks.}
  \label{fig:case_geometry_main}
\end{figure}

In RoboCasa365 \code{TurnOffStove} (\cref{fig:replanning_stove}), two distinct assumptions are contradicted in sequence during knob manipulation.
First, after a secure grip turns the front-right knob only a few degrees before the execution receipt reports that rotation was stopped by mechanical resistance, the model keeps the verified knob target and releases only to reposition its wrist posture for a wider rotational range.
Second, after the next turn brings the knob handle aligned horizontally along the stove---a pose that visually resembles the off position---the burner flame remains visibly blue in the camera view and the burner-on predicate stays true; recognizing that a symmetric handle alignment can occur halfway through a rotation, the model commands a further turn guided by the burner state until the flame extinguishes.
Furthermore, when late corrections face a tight step limit---as in a \code{swap_T} episode where a $94^\circ$ stem misalignment is detected at native step 345 with only 55 steps remaining (\cref{tab:analysis_budget} in \cref{app:trace})---paused-world preview rejects an initial 95-step correction and a 57-step revision \emph{before} moving the robot, enabling the model to compress approach, rotation, release, and withdrawal into a 51-step coordinated trajectory that succeeds at step 396 (further replanning episodes on \code{plug_in_charger}, \code{DeliverStraw}, and \code{SetUpCuttingStation} are detailed in \cref{app:cases}).

\Needspace{2\baselineskip}
\textbf{Geometry-guided manipulation with VGGT-Omega and SAM~3.1.}
\label{sec:case_vggt_sam}
\Cref{fig:case_geometry_main} compares recorded episodes with and without 3D geometric perception (\cref{sec:aug-input}) on matched scenes in RoboDojo and RoboCasa365.

\begingroup
\clubpenalty=10000
\emergencystretch=3em
On a randomized RoboDojo \code{pack_objects_into_box} scene (\cref{fig:case_vggt_sam}), the VGGT-Omega/SAM~3.1 object table reports an initial $18^\circ$ box yaw misalignment, prompting the model to grasp the front wall and rotate the box from $109^\circ$ to $90^\circ$ before packing; to grasp the hammer, querying pixel $(267, 274)$ via \code{geometry_query} with a VGGT-derived center-height plane yields $\mathbf{p}_{\mathrm{query}} \simeq (-0.1051, -0.1827, 0.7795)\,\mathrm{m}$, guiding a finger-referenced grasp at $(-0.105, -0.183, 0.778)\,\mathrm{m}$ to pack all four objects, whereas the run without VGGT/SAM exhausts the step limit.
\par\endgroup

On RoboCasa365 \code{StackBowlsCabinet} (\cref{fig:case_sam_robocasa}), the SAM~3.1 object table locates the open cabinet shelf top at $1.422\,\mathrm{m}$, the large-bowl rim at $0.991\,\mathrm{m}$ ($x = 5.431\,\mathrm{m}$), and the small-bowl rim at $0.967\,\mathrm{m}$ ($x = 5.502\,\mathrm{m}$); the model grasps each bowl $2\,\mathrm{cm}$ below its measured rim ($0.968\,\mathrm{m}$ and $0.948\,\mathrm{m}$) and places them at $1.477\,\mathrm{m}$ inside the cabinet to succeed in 29 responses, whereas without the object table the large bowl slips at the cabinet opening, settles inverted on the counter, and eventually falls to the floor.

\Needspace{2\baselineskip}
\textbf{Failure modes under multi-body coupling and expanding recovery scope.}
\label{sec:case_failure}

\begin{figure}[!htbp]
  \centering
  \begin{subfigure}{0.90\linewidth}
    \centering
    \includegraphics[trim=0 0 32.5bp 0,clip,width=\linewidth]{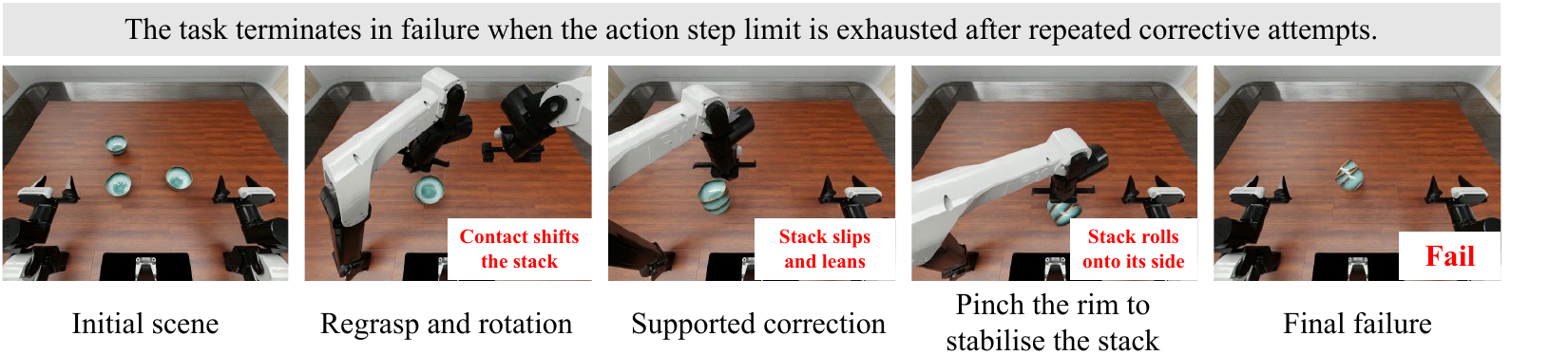}
    \caption{RoboDojo bowl stacking (\code{stack_bowls}): leveling a tilted third bowl lifts and rolls the nested stack, exhausting the 800-step native budget.}
    \label{fig:case_failure}
  \end{subfigure}
  \par\vspace{2pt}
  \begin{subfigure}{0.90\linewidth}
    \centering
    \includegraphics[width=\linewidth]{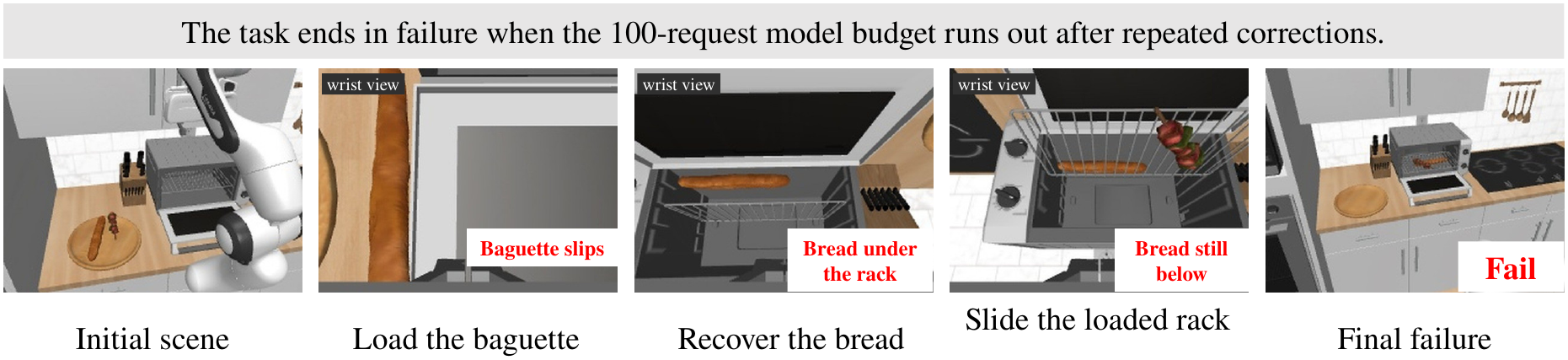}
    \caption{RoboCasa365 toaster loading (\code{HeatKebabSandwich}): the baguette lands below the wire rack, and pulling the thin rack rail consumes the 100-request budget.}
    \label{fig:case_failure_robocasa}
  \end{subfigure}
  \caption{\textbf{Failure modes after repeated corrections under multi-body coupling.} When a recovery action displaces an already assembled stack or traps an object beneath a fixture rack, local motion completions do not restore cumulative task progress before the budget expires.}
  \label{fig:case_failure_main}
\end{figure}
\Cref{fig:case_failure_main} illustrates two representative failure episodes where local corrections perturb coupled structures and exhaust the episode budget (step-by-step budget accounting in \cref{tab:case_failure_budget,tab:case_failure_budget_robocasa} in \cref{app:cases}).
In RoboDojo \code{stack_bowls} (\cref{fig:case_failure}), after the model recovers from an initial slip to form a two-bowl stack (242 steps used, 558 left), placing the third bowl leaves it tilted; attempting to level the top bowl at step 440 lifts the nested three-bowl stack together, which slips and rolls onto its side during rotation, and subsequent supported pushes roll again upon arm withdrawal until the 800-step native budget expires (Score 15/100).

In RoboCasa365 \code{HeatKebabSandwich} (\cref{fig:case_failure_robocasa}), while loading a baguette into a toaster oven, contact with the oven entrance causes the bread to slip and land inside the oven \emph{below} the wire rack at request 41, leaving its rack-contact predicate false.
To expose the bread underneath, the model must pull out the wire rack---expanding the problem to grasping a thin front rail that repeatedly slips between requests 63 and 86; although the kebab is placed on the rack at request 50 and the rack finally slides outward at request 86, only 14 requests remain and the 100-request budget expires before the baguette can be lifted onto the rack.
These failures highlight the boundary of kinematic trajectory execution: while stage verification and belief updates reliably detect discrepancies, recovery actions that disturb coupled multi-body assemblies or introduce secondary fixture obstacles can consume the remaining budget without achieving stable support.

\FloatBarrier

%% file: related_work.tex
\section{Related Work}

\subsection{LLMs and VLMs for Embodied Reasoning}

Robotic applications of language and vision-language models span task planning, spatial grounding, and direct action generation.
Alongside general-purpose models such as GPT-6~\citep{openai2026gpt6} and Claude Opus 5.5~\citep{anthropic2026opus55}, embodied foundation models specialize perception and reasoning for physical interaction~\citep{gemini2025robotics15,rynnbrain2026,hyembodied2026,embodiedr2026,physbrain2026}.
Evaluations of GPT-6 Astra~\citep{unexpected2026,exploring2026} investigate direct numerical control and cooperation with learned policies, revealing a gap between semantic task understanding and precise physical execution.
Other work adapts robot behavior through context: GPT-Policy~\citep{gptpolicy2026} preserves task-relevant demonstration transitions, while RoboICL~\citep{roboicl2026} organizes demonstrations and interaction memory around observations, actions, and execution receipts.

\subsection{Agentic Harnesses for Robotics}

Agentic harnesses organize model interaction with robot tools, programs, and execution feedback.
\emph{Trial-and-error interaction} operates within a task: Agent as Policy~\citep{agp2026} and CaP-Agent0~\citep{capx2026} within CaP-X allow agents to construct programs and revise execution using observations and tool results, while RoboHarness~\citep{roboharness2026} emphasizes a visual-geometric control interface.
The agent may also delegate action generation: PhysicalRSI~\citep{physicalrsi2026} provides task-aware memory and skills backed by code policies or $\pi_{0.5}$~\citep{pi052025}, and FastHarness~\citep{fastharness2026} adapts the frequency of VLM supervision over a frozen vision-language-action (VLA) model.

\emph{Cross-trial evolution} uses rollout feedback to revise programs, tools, or policies for subsequent executions.
HarnessPAI~\citep{harnesspai2026} keeps the generated program fixed within a rollout, including its observation-dependent checks, and revises it across rollouts.
HexaShell~\citep{hexashell2026} separates an acting agent from an optimizer that repairs and tests executable tools; EmbodiedEvo~\citep{embodiedevo2026} separates exploration from evaluation, passing distilled code skills and failure lessons to independent test agents.
Both freeze the acquired experience for evaluation without updating model weights.
Related approaches evolve skills and context code, attribute execution failures, or select informative physical trials~\citep{shaper2026,dynaharness2026,embodiedrsi2026}; rollout scale and candidate selection also affect harness improvement~\citep{scaleselection2026}.
WorldScape-R$^2$SI~\citep{r2si2026} jointly improves reusable skills and a reactive action model, while ENPIRE~\citep{enpire2026} lets agents revise policy-training algorithms and infrastructure using real-robot evaluations.

\emph{Skill libraries} complement cross-trial improvement by storing reusable procedures for later retrieval and composition.
ASPIRE~\citep{aspire2026} stores validated repair knowledge as reusable in-context guidance, including applicability conditions, repair strategies, and optional code sketches; RATs~\citep{rats2026} stores code skills acquired through self-directed play and retrieves them from a frozen library for downstream tasks.
Teach-and-Grow~\citep{teachgrow2026} derives closed-loop Skill Blocks from subgoals shared across demonstrations and retains execution conditions and repairs in a separate experience memory.
Real2Gym~\citep{real2gym2026} retains task procedures, object-relative motions, and recovery strategies from interaction in simulation environments reconstructed from videos, then reuses them on physical robots.
GaP~\citep{gap2026} focuses on composition, assembling existing perception, planning, and control modules into computation graphs refined through simulation.
Auriga organizes interaction within a persistent model session, where generated programs and direct tool calls share a physical execution interface and return observations and outcomes to guide subsequent decisions.

\subsection{Vision-Language-Action and World Action Models}

Learned action models map visual and language inputs to robot actions and can serve as policies invoked by an agent.
Recent VLA work explores heterogeneous-data co-training and cross-embodiment adaptation~\citep{pi052025,xvla2025,pi072026,lingbotvla2026,xiaomi2026,qwenrobot2026}, experience-driven policy improvement~\citep{pi062025}, and joint reasoning and action generation~\citep{g052026}.
InternVLA-A1.5~\citep{internvla152026} further introduces latent foresight supervised by a pretrained video model while retaining semantic training objectives.

World action models (WAMs) incorporate predictive modeling or supervision from future observations into action learning.
Video-based approaches jointly model actions and future observations~\citep{dreamzero2026,cosmospolicy2026,openwam2026}; VPP2~\citep{vpp22026} strengthens manipulation-specific video prediction, and X-WAM~\citep{xwam2026} adds multi-view RGB-D prediction with asynchronous action decoding.
Future supervision can also shape action representations without requiring future-video generation at deployment, as in Being-H0.7, Fast-WAM, and InternW0-$\Delta$~\citep{beingh2026,fastwam2026,internwdelta2026}.
JEPA-WAM~\citep{jepawam2026} couples latent transition prediction to action generation, while Long-WAM~\citep{longwam2026} studies causal pretraining and longer observation histories.
When these policies are used within a harness, the agent coordinates their invocation, evaluates execution outcomes, and revises the task procedure.

%% file: conclusion.tex
\section{Conclusion}
\label{sec:conclusion}

We have presented \textbf{Auriga}, a task-agnostic agent harness for zero-shot manipulation within a single episode, closing 68.3\% of the planning--execution gap of vision-language models on RoboDojo (from 22.5\% under the reference interface toward the 76.2\% arm-free upper bound) without target-task exploration, demonstrations, or environment resets.
In Auriga, the model makes all task decisions over an explicit task belief $B_t$ and acts through direct tool calls or synthesized multi-step Python programs, while the harness supplies calibrated 3D perception, finger-referenced dual-arm control, and a single admission gateway $\mathcal{A}$ that previews trajectories while simulation is paused and verifies stage tolerances and grasp bounds during execution for in-place recovery.
Developed across 13 cross-task capability iterations and frozen for evaluation, Auriga ranks first overall on the 42-task RoboDojo leaderboard with a 64.0 Score and 57.7\% success rate, maintains first place under scene randomization ($G = 46.3$) while learned policies drop by up to 33.9\%, and achieves a 73.2\% success rate on RoboCasa365 mobile manipulation.
Future work will extend this closed-loop execution architecture toward continuous force-sensitive manipulation, lower deliberation latency, and automated cross-task tool synthesis.

\par\addvspace{1.5\baselineskip}
\noindent\begin{minipage}{\linewidth}
\section*{Author List}
\phantomsection
\addcontentsline{toc}{section}{Author List}
\small
\noindent{\sffamily\bfseries\color{seheading}Contributors}\\*[1pt]
Chengcheng Wang, Yuheng Shi, Siyu Xu, Yutong Wang, Jinjing Zhao, Chenghao Xia, Jinqi Liao, Jiankun Peng, Ligong Bi, Yuechuan Ma, Jiaqi Liu, Wenbo Wang, Zijian Wang, Yuemin Wu, Minjing Dong, Xiu Su

\vspace{2pt}\nopagebreak
\noindent{\sffamily\bfseries\color{seheading}Project Leads}\\*[1pt]
Yunke Wang, Chengbin Du, Tao Huang, Jianyuan Guo, Chang Xu
\end{minipage}

%% file: appendix.tex
\clearpage
\appendix

\input{sec8_appendix}
\input{sec9_analysis}
\input{sec10_cases}

%% file: sec8_appendix.tex
\section{Benchmark Tasks and Evaluation Scope}
\label{app:tasks}
\label{app:protocol}

\Cref{tab:dims} lists the 42 tasks of RoboDojo by capability dimension (\cref{sec:exp-dims}).

\paragraph{Evaluation protocol comparison scope (\cref{tab:evaluation_conditions}).}
\Cref{tab:evaluation_conditions} in \cref{sec:problem_setting} distinguishes target-task environment trials from offline demonstration analysis.
Prepared code/skills include procedures later generalized across tasks; trial-derived memory denotes separate notes inherited from earlier trials.
Demo-derived artifacts are distilled text or code, whereas demo references are recorded images or motions, including those retrieved from frozen packages.
Task-independent tools, public task specifications, current-episode history, benchmark-required memory events, and training data encoded in model weights are outside these categories.
A fixed evaluation agent does not establish the absence of earlier task-driven tool development: PyRUA-Lean uses shared primitives without cross-episode memory, but does not report whether target-task trials informed their development, so its two preparation entries are marked as not reported~\citep{pyrua2026}; HexaShell is shown for non-Open tasks, as its Open setting omits task-specific development and notes~\citep{hexashell2026}; and the R$^2$SI row describes the inspected public release, where outer-loop trials exclude Open tasks and the snapshot supplies no separate memory files or demo context, although the framework supports memory updates and its WAMs use trained weights~\citep{r2si2026}.

\begin{table}[!htbp]
\centering
\footnotesize
\caption{The five capability dimensions of RoboDojo and their tasks.}
\label{tab:dims}
\setlength{\tabcolsep}{3.5pt}
\renewcommand{\arraystretch}{0.78}
\begin{tabular}{@{}lc>{\raggedright\arraybackslash}p{3.4cm}>{\raggedright\arraybackslash}p{9.7cm}@{}}
\toprule
Dimension & Tasks & What it covers & Task identifiers \\
\midrule
Generalization & 12 & Manipulation across varied objects and layouts, including randomized scene variants & \code{stack_bowls}, \code{push_T}, \code{pack_objects_into_box}, \code{fold_clothes}, \code{hang_mugs}, \code{sweep_blocks}, \code{pour_liquid_into_cup}, \code{make_toast}, \code{arrange_largest_number}, \code{sort_nesting_dolls_by_size}, \code{store_laptop_and_headphones}, \code{stack_blocks} \\
Precision & 8 & Fine-grained manipulation under tight spatial constraints & \code{fasten_screws}, \code{plug_in_charger}, \code{insert_tubes}, \code{pour_balls_into_vase}, \code{play_Xylophone}, \code{deposit_coin}, \code{insert_key}, \code{build_tower} \\
Long-Horizon & 8 & Multi-step tasks that require completing several subgoals & \code{put_bottles_into_dustbin}, \code{fill_pen_holder}, \code{classify_objects}, \code{play_tic_tac_toe}, \code{fill_egg_holder}, \code{organize_table}, \code{make_kong}, \code{play_stacking_toy} \\
Memory & 6 & Tracking state, recalling a sequence, or matching after a delay & \code{cover_blocks}, \code{match_and_pick_from_conveyor}, \code{swap_blocks}, \code{swap_T}, \code{press_by_number}, \code{imitate_sorting_sequence} \\
Open & 8 & Open-ended tasks and manipulation conditioned on language or images & \code{align_blocks}, \code{general_pickup}, \code{stack_blocks_by_language}, \code{solve_equation}, \code{classify_objects_by_language}, \code{pick_from_conveyor_by_image}, \code{store_tools_in_toolbox}, \code{pour_by_language} \\
\bottomrule
\end{tabular}
\end{table}

\section{Ablation Details}
\label{app:ablation}

\Cref{tab:abl-tasks} lists the ten tasks of each ablation table in \cref{sec:exp-ablation}: five standard layouts per task (50 cases) in \crefrange{tab:abl-measure}{tab:abl-action}, and 30 cases in \cref{tab:abl-programs}.

\begin{table}[!htbp]
\centering
\scriptsize
\caption{The ten tasks of each ablation table by dimension: five standard layouts per task (50 cases) in \crefrange{tab:abl-measure}{tab:abl-action}, and 30 cases in \cref{tab:abl-programs}.}
\label{tab:abl-tasks}
\setlength{\tabcolsep}{4pt}
\renewcommand{\arraystretch}{0.78}
\begin{tabular}{@{}l>{\raggedright\arraybackslash}p{14.8cm}@{}}
\toprule
Table & Tasks \\
\midrule
\cref{tab:abl-measure} & Generalization: \code{stack_blocks}, \code{stack_bowls}, \code{push_T}; Long-Horizon: \code{play_tic_tac_toe}; Memory: \code{cover_blocks}, \code{press_by_number}, \code{imitate_sorting_sequence}; Open: \code{align_blocks}, \code{stack_blocks_by_language}, \code{classify_objects_by_language} \\
\cref{tab:abl-collision} & Generalization: \code{stack_bowls}, \code{sort_nesting_dolls_by_size}; Precision: \code{fasten_screws}, \code{insert_tubes}; Long-Horizon: \code{classify_objects}, \code{play_stacking_toy}; Memory: \code{cover_blocks}, \code{swap_T}; Open: \code{classify_objects_by_language}, \code{align_blocks} \\
\cref{tab:abl-checkpoint} & Generalization: \code{fold_clothes}, \code{arrange_largest_number}, \code{sort_nesting_dolls_by_size}; Precision: \code{fasten_screws}; Long-Horizon: \code{classify_objects}; Memory: \code{swap_T}, \code{cover_blocks}, \code{imitate_sorting_sequence}; Open: \code{pick_from_conveyor_by_image}, \code{classify_objects_by_language} \\
\cref{tab:abl-action} & Generalization: \code{fold_clothes}, \code{arrange_largest_number}; Precision: \code{fasten_screws}, \code{play_Xylophone}; Long-Horizon: \code{classify_objects}, \code{play_tic_tac_toe}; Memory: \code{press_by_number}, \code{cover_blocks}; Open: \code{classify_objects_by_language}, \code{align_blocks} \\
\cref{tab:abl-programs} & Generalization: \code{sort_nesting_dolls_by_size}, \code{pack_objects_into_box}; Precision: \code{fasten_screws}; Long-Horizon: \code{play_stacking_toy}, \code{play_tic_tac_toe}, \code{classify_objects}; Memory: \code{press_by_number}, \code{imitate_sorting_sequence}; Open: \code{pick_from_conveyor_by_image}, \code{classify_objects_by_language} \\
\bottomrule
\end{tabular}
\end{table}

\FloatBarrier

\section{RoboCasa365 Results}
\label{app:robocasa}

\begin{figure}[!htbp]
\centering
\includegraphics[width=\linewidth]{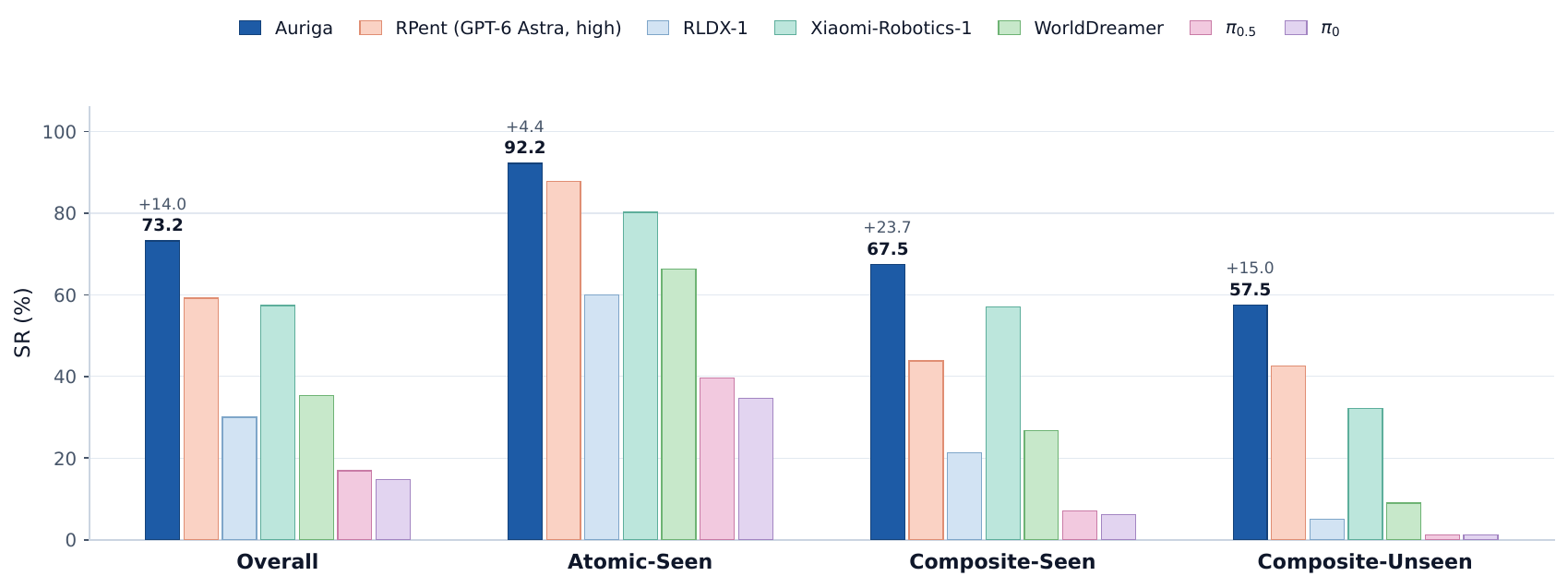}
\caption{Success rates (\%) on RoboCasa365 by group: \method, RPent, and RLDX-1 from \cref{tab:robocasa-main} (target kitchens, 250 episodes) and the leaderboard entries of \cref{tab:robocasa-board} (pretraining kitchens), with \method first. In each group, the bar of \method shows its rate and, above it, its difference from the highest other method of \cref{tab:robocasa-main}; another method's bar is labeled with its rate only where it reaches that of \method (\cref{tab:robocasa-main,tab:robocasa-board} list all values).}
\label{fig:robocasa-dims}
\end{figure}

\begin{figure}[!htbp]
\centering
\includegraphics[width=\linewidth]{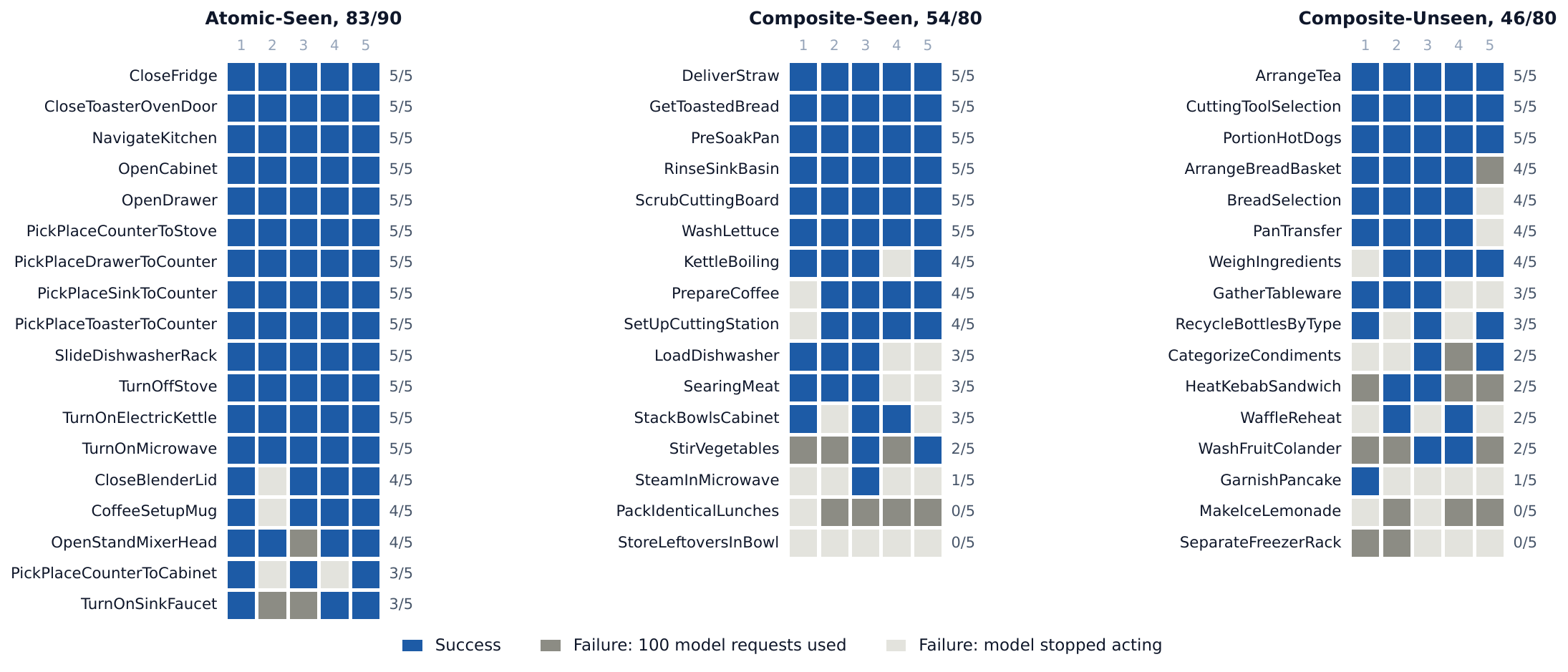}
\caption{Outcome of every episode of \method on the 250 Target50 episodes, by task and seed: success, failure after the 100 model requests of the episode were used, and failure after the model stopped acting for several turns. Within each group, tasks are ordered by their number of successes.}
\label{fig:robocasa-pertask}
\end{figure}

\Cref{tab:robocasa-board} lists four entries of the official RoboCasa365 leaderboard~\citep{robocasaboard2026}, which are evaluated in the pretraining kitchens rather than in the target kitchens of \cref{tab:robocasa-main}; Xiaomi-Robotics-1 leads all four columns among these entries, with 57.4\% Overall.
\Cref{fig:robocasa-dims} plots these four entries alongside \method, RPent, and RLDX-1 from \cref{tab:robocasa-main}.
\Cref{fig:robocasa-pertask} gives the outcome of every task and episode of \method.

\begin{table}[!htbp]
\centering
\footnotesize
\caption{Success rates (\%) of four entries of the official RoboCasa365 leaderboard~\citep{robocasaboard2026} on the 50 Target50 tasks, evaluated in the pretraining kitchens and ordered by ascending Overall SR. Overall is the mean over the 50 tasks.}
\label{tab:robocasa-board}
\setlength{\tabcolsep}{4pt}
\begin{tabular}{@{}llrrrr@{}}
\toprule
Method & Type & Overall & Atomic-Seen & Composite-Seen & Composite-Unseen \\
\midrule
$\pi_{0}$ & VLA & 14.8 & 34.6 & 6.1 & 1.1 \\
$\pi_{0.5}$ & VLA & 16.9 & 39.6 & 7.1 & 1.2 \\
WorldDreamer & WAM & 35.3 & 66.3 & 26.7 & 9.0 \\
Xiaomi-Robotics-1 & VLA & 57.4 & 80.2 & 57.1 & 32.1 \\
\bottomrule
\end{tabular}
\end{table}

\section{Additional Results}
\label{app:more}

\begin{figure}[!htbp]
\centering
\includegraphics[width=\linewidth]{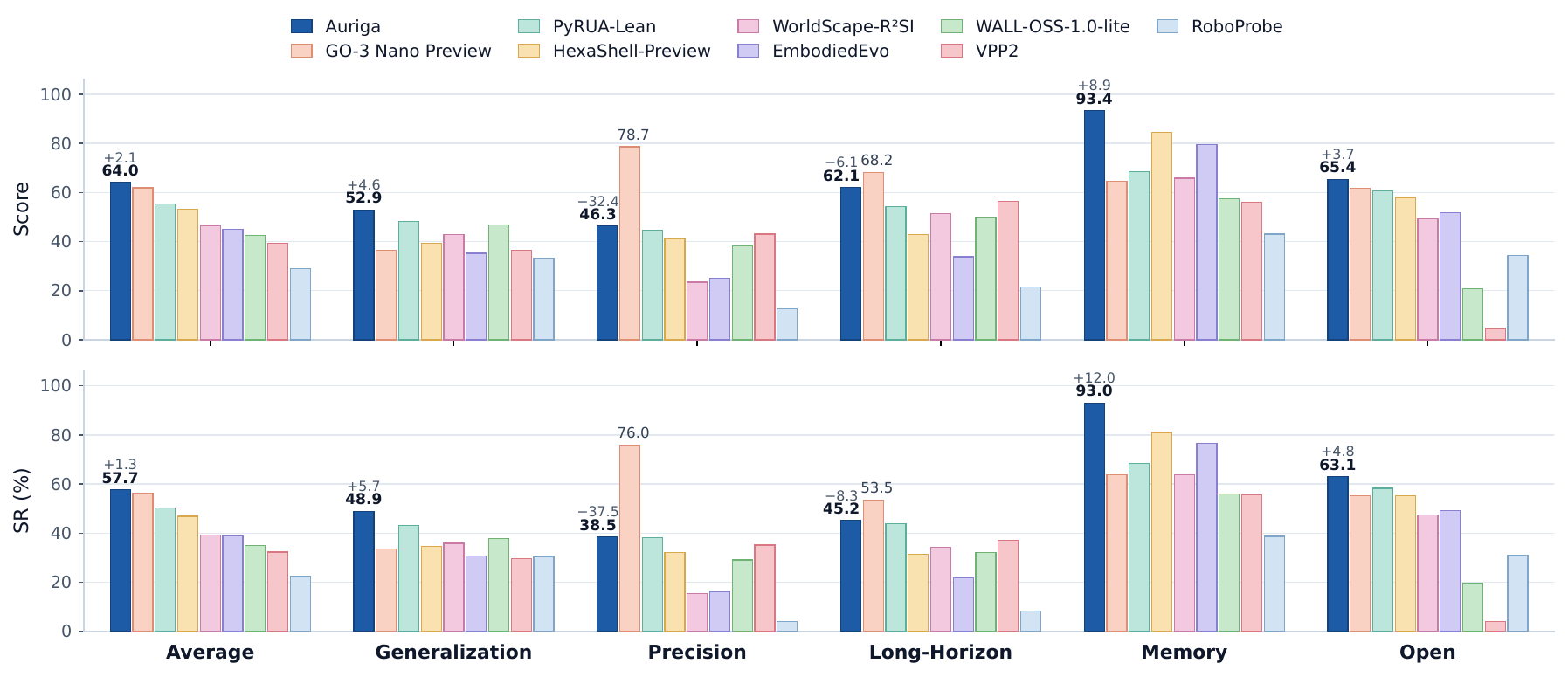}
\caption{Score (top) and SR in \% (bottom) of \method, the seven other methods of \cref{fig:agents-radar}, and RoboProbe, on average and by dimension. \method is the first bar of each group, followed by the other methods in order of average Score. Above the bar of \method are its value and, higher up in grey, its difference from the highest of the other eight methods, whose value is given above its own bar where it is higher than \method.}
\label{fig:agents}
\end{figure}

\Cref{tab:cost-dim} lists the episode cost of \method by dimension corresponding to \cref{fig:cost-dim}.
Across the 6,299 episodes in \cref{tab:cost-dim}, successful episodes (3,405 episodes) complete efficiently once the task subgoals are verified, averaging 30.3 model calls, 3.10 million input tokens, 11.5 minutes, and 65.6\% of the native-step limit without ever exhausting the step budget.
The remaining 2,894 episodes, which continue online recovery attempts until the budget expires, average 67.6 model calls, 8.61 million input tokens, 27.5 minutes, and 96.4\% of the native-step limit, accounting for 65.5\% of all model calls and 70.2\% of all input tokens.
\Cref{tab:gen} lists the SR and Score on the standard and random scenes of the 12 Generalization tasks together with the change between the two scene types (with the SR change and $G$ plotted in \cref{fig:gen}), and \cref{fig:agents} plots the Score and SR of the methods in \cref{fig:agents-radar} and of RoboProbe as bars.

\begin{table}[!htbp]
\centering
\footnotesize
\caption{Cost of an \method episode by dimension across the 6,299 evaluated episodes: episodes, mean model calls, input tokens and their cached share, output tokens, the share of the native-step limit used, and wall-clock time.}
\label{tab:cost-dim}
\setlength{\tabcolsep}{6pt}
\renewcommand{\arraystretch}{0.90}
\begin{tabular}{@{}l*{7}{r}@{}}
\toprule
Dimension & Episodes & Model & Input & Cached & Output & Steps & Time \\
 & & calls & (M) & (\%) & (k) & used (\%) & (min) \\
\midrule
Generalization & 1,800 & 54.2 & 6.57 & 84.8 & 21.9 & 79.6 & 21.6 \\
Precision & 1,200 & 50.2 & 6.02 & 85.9 & 19.5 & 86.9 & 20.2 \\
Long-Horizon & 1,199 & 54.5 & 6.70 & 85.2 & 21.8 & 88.8 & 23.0 \\
Memory & 900 & 30.0 & 3.11 & 85.5 & 9.0 & 68.9 & 11.3 \\
Open & 1,200 & 40.7 & 4.67 & 86.1 & 15.4 & 71.8 & 14.9 \\
\midrule
All & 6,299 & 47.5 & 5.63 & 85.4 & 18.3 & 79.7 & 18.8 \\
\bottomrule
\end{tabular}
\end{table}

\begin{table}[!htbp]
\centering
\footnotesize
\caption{Generalization under scene randomization on the 12 Generalization tasks for the entries of \cref{tab:main}: SR (\%) and Score on the standard and random scenes, with the change (random $-$ standard) and relative SR change. Above \method, entries are ordered by ascending composite generalization score $G$ (\cref{fig:gen}). Bold: best entry per column.}
\label{tab:gen}
\setlength{\tabcolsep}{4.5pt}
\renewcommand{\arraystretch}{0.88}
\begin{tabular}{lrrrrrrr}
\toprule
 & \multicolumn{4}{c}{SR (\%)} & \multicolumn{3}{c}{Score} \\
\cmidrule(lr){2-5}\cmidrule(lr){6-8}
Method & Standard & Random & Change & Rel.\ change & Standard & Random & Change \\
\midrule
PhysicalRSI & 25.67 & 5.56 & $-$20.1 & $-$78\% & 31.57 & 11.04 & $-$20.5 \\
VPP2 & 46.44 & 12.56 & $-$33.9 & $-$73\% & 52.14 & 20.63 & $-$31.5 \\
Awomo-0.5 & 30.89 & 16.78 & $-$14.1 & $-$46\% & 37.70 & 22.92 & $-$14.8 \\
Rex-M1 Preview & 34.78 & 17.33 & $-$17.5 & $-$50\% & 42.64 & 25.46 & $-$17.2 \\
GO-3 Nano Preview & 43.00 & 24.33 & $-$18.7 & $-$43\% & 45.93 & 26.97 & $-$19.0 \\
Simate-beta & 33.22 & 22.67 & $-$10.6 & $-$32\% & 40.54 & 29.63 & $-$10.9 \\
RoboProbe & 32.67 & 28.33 & $-$4.3 & $-$13\% & 35.32 & 31.40 & $-$3.9 \\
HexaShell-Preview & 38.00 & 31.00 & $-$7.0 & $-$18\% & 43.28 & 35.15 & $-$8.1 \\
WALL-OSS-1.0-lite & 45.44 & 30.44 & $-$15.0 & $-$33\% & 53.44 & 39.88 & $-$13.6 \\
EmbodiedEvo & 30.33 & 31.33 & \textbf{{\boldmath$+$}1.0} & \textbf{{\boldmath$+$}3\%} & 34.55 & 35.88 & \textbf{{\boldmath$+$}1.3} \\
WorldScape-R$^2$SI & 40.33 & 31.33 & $-$9.0 & $-$22\% & 46.62 & 39.28 & $-$7.3 \\
PyRUA-Lean & 45.33 & 41.00 & $-$4.3 & $-$10\% & 50.47 & 46.07 & $-$4.4 \\
\midrule
\ours \method & \textbf{52.28} & \textbf{45.61} & $-$6.7 & $-$13\% & \textbf{56.22} & \textbf{49.58} & $-$6.6 \\
\bottomrule
\end{tabular}
\end{table}

\FloatBarrier

\section{Per-Task Results}
\label{app:pertask}

\Cref{fig:pertask-h2h} compares \method with the four other agent methods of \cref{tab:main} (PyRUA-Lean, HexaShell-Preview, EmbodiedEvo, and RoboProbe) task by task in Score and in SR, and \cref{tab:pertask} lists the corresponding per-task Score and SR values.

\noindent\begin{minipage}{\linewidth}
\captionsetup{type=figure}
\centering
\includegraphics[width=0.96\linewidth]{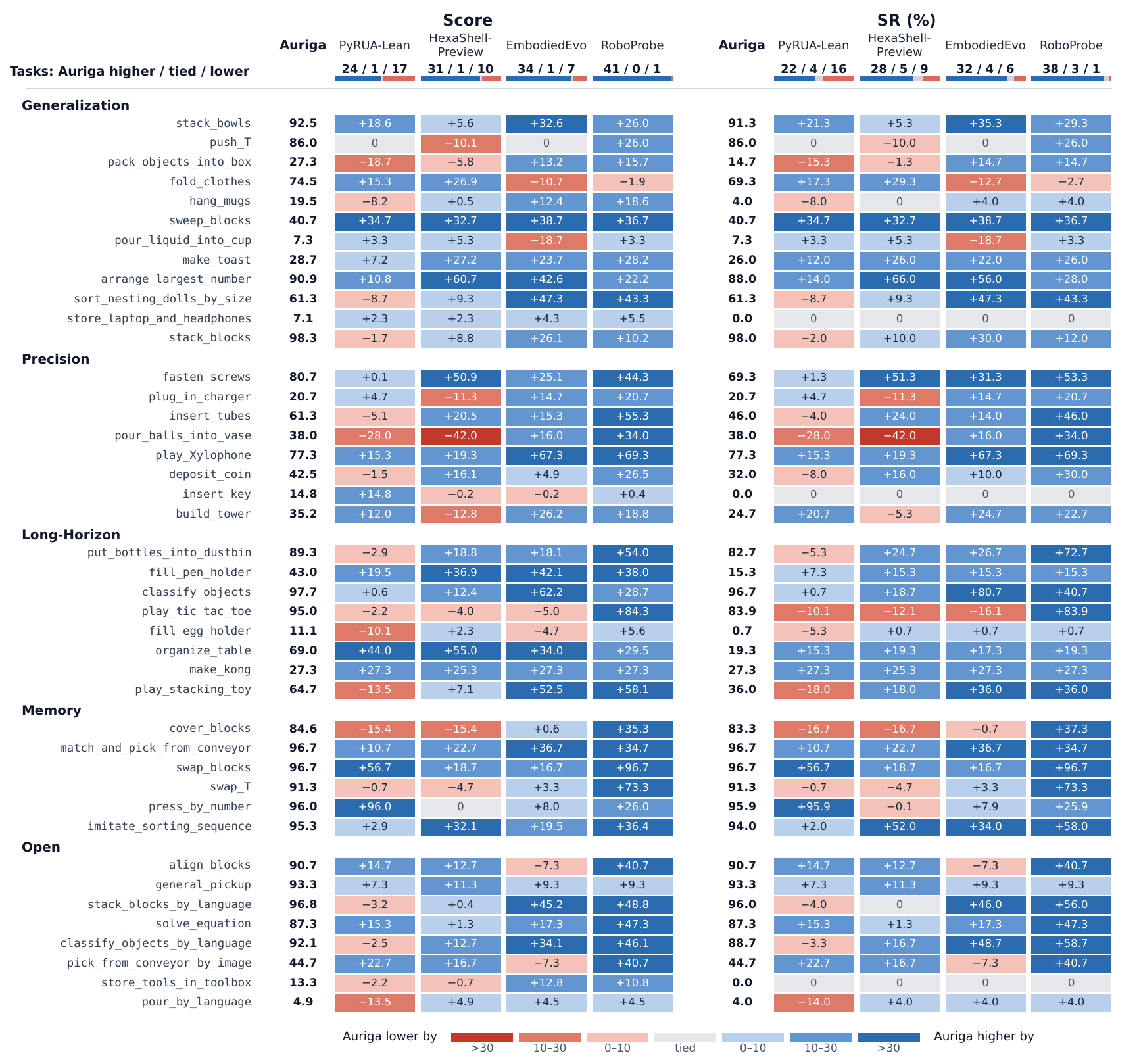}
\caption{\method against PyRUA-Lean, HexaShell-Preview, EmbodiedEvo, and RoboProbe on each task, in Score (left) and SR in \% (right). In each panel, the first column gives the value of \method and each remaining column gives the difference (\method minus that method): blue where \method is higher, red where it is lower, and grey where the two are tied, darker for larger differences. The top row counts the tasks on which \method is higher, tied, and lower, with a bar of their shares.}
\label{fig:pertask-h2h}
\end{minipage}

\begingroup
\footnotesize
\setlength{\tabcolsep}{4.2pt}
\renewcommand{\arraystretch}{0.9}
\setlength{\LTcapwidth}{\linewidth}
\begin{longtable}{@{}lrrrrrr@{}}
\caption{Per-task results of \method (SR in \% and Score) and per-task SR (\%) of the four other agent methods in \cref{tab:main} (\cref{sec:exp-robodojo}). Bold: highest SR of a task (none where all five are at 0.0).}
\label{tab:pertask} \\
\toprule
 & \multicolumn{2}{c}{\method} & \multicolumn{4}{c}{SR (\%) of the four other agent methods} \\
\cmidrule(lr){2-3}\cmidrule(l){4-7}
Task & SR & Score & \shortstack{PyRUA-\\Lean} & \shortstack{HexaShell-\\Preview} & EmbodiedEvo & RoboProbe \\
\midrule
\endfirsthead
\caption[]{Per-task results (continued).} \\
\toprule
 & \multicolumn{2}{c}{\method} & \multicolumn{4}{c}{SR (\%) of the four other agent methods} \\
\cmidrule(lr){2-3}\cmidrule(l){4-7}
Task & SR & Score & \shortstack{PyRUA-\\Lean} & \shortstack{HexaShell-\\Preview} & EmbodiedEvo & RoboProbe \\
\midrule
\endhead
\midrule
\multicolumn{7}{r}{\emph{Continued on the next page}} \\
\endfoot
\bottomrule
\endlastfoot
\multicolumn{7}{@{}l}{\emph{Generalization}} \\
\code{stack_bowls} & \textbf{91.3} & 92.5 & 70.0 & 86.0 & 56.0 & 62.0 \\
\code{push_T} & 86.0 & 86.0 & 86.0 & \textbf{96.0} & 86.0 & 60.0 \\
\code{pack_objects_into_box} & 14.7 & 27.3 & \textbf{30.0} & 16.0 & 0.0 & 0.0 \\
\code{fold_clothes} & 69.3 & 74.5 & 52.0 & 40.0 & \textbf{82.0} & 72.0 \\
\code{hang_mugs} & 4.0 & 19.5 & \textbf{12.0} & 4.0 & 0.0 & 0.0 \\
\code{sweep_blocks} & \textbf{40.7} & 40.7 & 6.0 & 8.0 & 2.0 & 4.0 \\
\code{pour_liquid_into_cup} & 7.3 & 7.3 & 4.0 & 2.0 & \textbf{26.0} & 4.0 \\
\code{make_toast} & \textbf{26.0} & 28.7 & 14.0 & 0.0 & 4.0 & 0.0 \\
\code{arrange_largest_number} & \textbf{88.0} & 90.9 & 74.0 & 22.0 & 32.0 & 60.0 \\
\code{sort_nesting_dolls_by_size} & 61.3 & 61.3 & \textbf{70.0} & 52.0 & 14.0 & 18.0 \\
\code{store_laptop_and_headphones} & 0.0 & 7.1 & 0.0 & 0.0 & 0.0 & 0.0 \\
\code{stack_blocks} & 98.0 & 98.3 & \textbf{100.0} & 88.0 & 68.0 & 86.0 \\
\multicolumn{7}{@{}l}{\emph{Precision}} \\
\code{fasten_screws} & \textbf{69.3} & 80.7 & 68.0 & 18.0 & 38.0 & 16.0 \\
\code{plug_in_charger} & 20.7 & 20.7 & 16.0 & \textbf{32.0} & 6.0 & 0.0 \\
\code{insert_tubes} & 46.0 & 61.3 & \textbf{50.0} & 22.0 & 32.0 & 0.0 \\
\code{pour_balls_into_vase} & 38.0 & 38.0 & 66.0 & \textbf{80.0} & 22.0 & 4.0 \\
\code{play_Xylophone} & \textbf{77.3} & 77.3 & 62.0 & 58.0 & 10.0 & 8.0 \\
\code{deposit_coin} & 32.0 & 42.5 & \textbf{40.0} & 16.0 & 22.0 & 2.0 \\
\code{insert_key} & 0.0 & 14.8 & 0.0 & 0.0 & 0.0 & 0.0 \\
\code{build_tower} & 24.7 & 35.2 & 4.0 & \textbf{30.0} & 0.0 & 2.0 \\
\multicolumn{7}{@{}l}{\emph{Long-Horizon}} \\
\code{put_bottles_into_dustbin} & 82.7 & 89.3 & \textbf{88.0} & 58.0 & 56.0 & 10.0 \\
\code{fill_pen_holder} & \textbf{15.3} & 43.0 & 8.0 & 0.0 & 0.0 & 0.0 \\
\code{classify_objects} & \textbf{96.7} & 97.7 & 96.0 & 78.0 & 16.0 & 56.0 \\
\code{play_tic_tac_toe} & 83.9 & 95.0 & 94.0 & 96.0 & \textbf{100.0} & 0.0 \\
\code{fill_egg_holder} & 0.7 & 11.1 & \textbf{6.0} & 0.0 & 0.0 & 0.0 \\
\code{organize_table} & \textbf{19.3} & 69.0 & 4.0 & 0.0 & 2.0 & 0.0 \\
\code{make_kong} & \textbf{27.3} & 27.3 & 0.0 & 2.0 & 0.0 & 0.0 \\
\code{play_stacking_toy} & 36.0 & 64.7 & \textbf{54.0} & 18.0 & 0.0 & 0.0 \\
\multicolumn{7}{@{}l}{\emph{Memory}} \\
\code{cover_blocks} & 83.3 & 84.6 & \textbf{100.0} & \textbf{100.0} & 84.0 & 46.0 \\
\code{match_and_pick_from_conveyor} & \textbf{96.7} & 96.7 & 86.0 & 74.0 & 60.0 & 62.0 \\
\code{swap_blocks} & \textbf{96.7} & 96.7 & 40.0 & 78.0 & 80.0 & 0.0 \\
\code{swap_T} & 91.3 & 91.3 & 92.0 & \textbf{96.0} & 88.0 & 18.0 \\
\code{press_by_number} & 95.9 & 96.0 & 0.0 & \textbf{96.0} & 88.0 & 70.0 \\
\code{imitate_sorting_sequence} & \textbf{94.0} & 95.3 & 92.0 & 42.0 & 60.0 & 36.0 \\
\multicolumn{7}{@{}l}{\emph{Open}} \\
\code{align_blocks} & 90.7 & 90.7 & 76.0 & 78.0 & \textbf{98.0} & 50.0 \\
\code{general_pickup} & \textbf{93.3} & 93.3 & 86.0 & 82.0 & 84.0 & 84.0 \\
\code{stack_blocks_by_language} & 96.0 & 96.8 & \textbf{100.0} & 96.0 & 50.0 & 40.0 \\
\code{solve_equation} & \textbf{87.3} & 87.3 & 72.0 & 86.0 & 70.0 & 40.0 \\
\code{classify_objects_by_language} & 88.7 & 92.1 & \textbf{92.0} & 72.0 & 40.0 & 30.0 \\
\code{pick_from_conveyor_by_image} & 44.7 & 44.7 & 22.0 & 28.0 & \textbf{52.0} & 4.0 \\
\code{store_tools_in_toolbox} & 0.0 & 13.3 & 0.0 & 0.0 & 0.0 & 0.0 \\
\code{pour_by_language} & 4.0 & 4.9 & \textbf{18.0} & 0.0 & 0.0 & 0.0 \\
\end{longtable}
\endgroup

%% file: sec9_analysis.tex
\section{Extended Execution Trace Analysis}
\label{app:trace}
\label{sec:analysis}

We examine \code{swap_T}, a pose-exchange task in the Memory dimension, to study how a remembered objective guides manipulation after the scene changes (complementing the matched-scene baseline comparison in \cref{sec:main-cost,tab:analysis_baseline}).
The robot must exchange two T-shaped blocks so that each occupies the other's initial position and orientation.
This task combines three demands in a compact setting: preserving a spatial reference after its source object moves, coordinating access to a shared workspace, and correcting the object pose when grasping or release changes it.
The asymmetric T shape makes orientation part of the goal, so exchanging positions alone is insufficient.
We analyze the 15-scene standard-variant diagnostic set for this task, using their task-belief records, action specifications, execution receipts, and evaluator outcomes.

\paragraph{Preserving object goals across changes in execution.}
\label{sec:analysis_recovery}
For analysis, let $\hat{T}_{A,0}$ and $\hat{T}_{B,0}$ denote the two initial planar poses comprising planar position and yaw estimated from the calibrated images.
The exchange assigns the object goals
\begin{equation}
  G_A=\hat{T}_{B,0}, \qquad G_B=\hat{T}_{A,0}.
  \label{eq:swap_goals}
\end{equation}
These goals remain fixed when the model changes a grasp or assigns the remaining transfer to another arm.
The current object location and its relation to the gripper, however, must be updated from subsequent observations.
This distinction matters after slippage: re-localizing the fallen block supplies a new pickup pose, while the saved initial pose still supplies its destination.
The recorded \code{task_plan} separates these roles through the plan, attention, and ledger described in \cref{sec:belief}.
Its updates retain the original placement requirement while revising which physical conditions are supported and which remain unresolved.

\textbf{A completed motion can still require a task-level correction.}
\Cref{tab:analysis_recovery} follows a successful exchange in which the pink block has already reached the blue block's initial pose.
During the remaining right-arm transfer, the blue block slips onto the table.
The crossing stage reports \code{completed}, but its grip feedback records a zero final aperture, and the returned image places the block to the left of its target.
The next belief revision marks blue retention as \code{contradicted} while retaining the pink placement as \code{supported}.
The model then selects the nearby left arm, re-localizes the blue stem, and specifies a short, slow transfer to the original target.
Thus the correction changes the arm, pickup coordinates, and transport motion together, while preserving both the completed placement and the outstanding object goal.

\begin{table}[!htbp]
  \centering
  \footnotesize
  \setlength{\tabcolsep}{4pt}
  \renewcommand{\arraystretch}{1.15}
  \caption{Task state, execution feedback, and correction in a successful T-block exchange.
  Entries condense the recorded task belief and actions; steps are cumulative native actions from receipt budget counters.
  The destination remains the opposite block's initial pose throughout.}
  \label{tab:analysis_recovery}
  \begin{tabular}{@{}p{0.08\linewidth}p{0.40\linewidth}p{0.47\linewidth}@{}}
    \toprule
    Step & Recorded state and feedback & Decision or verified outcome \\
    \midrule
    72 & Both blocks are lifted; the belief records retention as \code{supported} and the exchange as \code{pending}. &
    Preserve both initial outlines; place pink first and inspect its release. \\
    169 & Pink has been released at blue's initial pose; blue remains to be transferred. &
    Keep the completed placement and move blue with the right arm. \\
    278 & The transfer motion completes, but the final aperture is zero and the image shows blue on the table. &
    Mark blue retention as \code{contradicted}; preserve pink as \code{supported}; switch recovery to the left arm. \\
    332 & The left-arm re-grasp and slow lift complete; the right arm returns. &
    Use a short, slow placement at the saved target, followed by release and withdrawal. \\
    380 & The evaluator reports task success within the 400-step budget. &
    Both object goals are satisfied after the local recovery. \\
    \bottomrule
  \end{tabular}
\end{table}

\textbf{Recovery preserves completed work.}
The two ledger entries distinguish an unresolved blue transfer from an already supported pink placement.
Consequently, the model does not repeat the entire exchange or derive a new destination from the displaced scene.
It repairs the remaining relation between one object and its saved goal.
This is the operational value of maintaining task state: evidence can invalidate a local assumption without erasing the broader objective or the progress already achieved.
It also explains why counting only \code{tracking_error} receipts would miss an important class of recovery.
Here, the discrepancy lies between successful robot motion and unsuccessful object transport, and is exposed by combining the receipt with the resulting observation.

\paragraph{Using feedback to revise feasible operations.}
\label{sec:analysis_mechanisms}
\Cref{tab:analysis_mechanisms} summarizes mechanism use across the 15 episodes.
Geometric queries provide coordinates for pickup and placement; belief updates retain the goal and revise progress; previews evaluate candidate operations before execution.
All episodes use each of these capabilities.
Their roles become concrete in the preceding trace: the saved target survives the transfer failure, a fresh localization supplies the new grasp, and the revised action specification changes which arm performs the remaining work.
The table counts usage at the interface where it occurs, including calls made through model-written programs.

\begin{table}[!htbp]
  \centering
  \footnotesize
  \setlength{\tabcolsep}{4pt}
  \renewcommand{\arraystretch}{1.12}
  \caption{Mechanism use in the 15 T-block exchange episodes.
  Belief updates exclude the initial submission.
  Preview evaluations include unsuccessful candidates; generated plans are the candidates returned with a plan identifier.
  Plan and guard counts refer to specifications, including revisions.
  Observation pauses count actual receipt-reported stops.}
  \label{tab:analysis_mechanisms}
  \begin{tabular*}{\linewidth}{@{\extracolsep{\fill}}llrr@{}}
    \toprule
    Mechanism & Interface & Episodes & Count and unit \\
    \midrule
    Task-belief updates & \code{task_plan} & 15/15 & 25 updates \\
    Geometric queries & \code{geometry_query} & 15/15 & 17 calls \\
    Preview evaluations & \code{robot_action} & 15/15 & 106 evaluations \\
    Generated action plans & \code{robot_action} & 15/15 & 74 plans \\
    \quad Multi-stage plans & \code{move}, \code{gripper}, \code{observe} & 15/15 & 73 plans \\
    \quad Plans involving both arms & \code{arm} & 15/15 & 54 plans \\
    Grasp-guard specifications & \code{grip_guard_preset}$^\dagger$ & 15/15 & 70 stages \\
    Observation pauses & \code{observation_checkpoint}$^\ddagger$ & 1/15 & 1 pause \\
    \bottomrule
  \end{tabular*}
  \par\smallskip
  \begin{minipage}{\linewidth}
    \footnotesize\raggedright
    $^\dagger$Includes explicit aperture bounds as well as guard presets.
    $^\ddagger$The receipt reason identifying an observation pause; a pause returns control for inspection and does not itself imply an error.
  \end{minipage}
\end{table}

\textbf{Coordination is expressed at the scale of an operation.}
Multi-stage plans account for 73 of the 74 generated plans.
The sequence can include pickup, lift, transfer, release, and withdrawal, allowing the model to consider the entire operation before committing to it.
The 54 plans involving both arms include staggered motion and one-arm holding while the other places an object.
This is well suited to pose exchange: both destinations are initially occupied, and the arms need access to the same central region.
The recorded plans repeatedly resolve this dependency by holding one block clear, delivering the other, and then completing the exchange.
The relevant coordination is the ordering of access and the preservation of held objects, which can be expressed within one plan even when the arms move at different times.

\textbf{Preview makes correction cost available before motion.}
A second successful exchange shows how the model uses the remaining budget when a late observation contradicts placement.
After 345 native steps, the blue block is correctly placed, but the red block's stem points in the wrong direction.
The next belief revision preserves the required destination and calls for a right-arm re-grasp and approximately 94-degree clockwise correction.
\Cref{tab:analysis_budget} records the subsequent candidate costs.
With 55 steps remaining, the model shortens the route and coordinates gripper motion with the approach and return.
The executed specification also contains an explicit release stage.
Task success is reached at step 396.
The useful change is therefore more specific than issuing another grasp: the correction is reorganized to include the outstanding orientation change, release, and return within the available budget.

\begin{table}[!htbp]
  \centering
  \small
  \setlength{\tabcolsep}{5pt}
  \caption{Recorded preview costs during the late orientation correction.
  The world remains at native step 345 during these previews, with 55 steps available.
  The first two candidates are rejected because their planned lengths
exceed the remaining 55-step native budget.
  Candidate rows give planned lengths; the last row gives steps actually used before the evaluator signals success.}
  \label{tab:analysis_budget}
  \begin{tabular*}{\linewidth}{@{\extracolsep{\fill}}lrr@{}}
    \toprule
    Candidate & Native steps & Status \\
    \midrule
    Initial correction & 95 & Rejected: insufficient budget \\
    Revised route & 57 & Rejected: insufficient budget \\
    Coordinated approach and return & 54 & Planned \\
    Final plan with explicit release & 55 & Planned \\
    \midrule
    Executed correction to task success & 51 & Success at step 396 \\
    \bottomrule
  \end{tabular*}
\end{table}

The 106 preview evaluations and 74 generated plans in \cref{tab:analysis_mechanisms} reflect this distinction between evaluating a candidate and obtaining a plan for execution.
The remaining 32 evaluations report unsuccessful candidates without generating such a plan.
Likewise, the observation pause provides an opportunity to inspect intermediate results; it is counted separately from grasp-guard specifications and execution errors.
The two traces show how deliberation and physical execution interact: one revises the physical route after object loss, while the other revises the cost and timing of a correction before moving.
Together with the matched-scene comparison against RoboProbe in \cref{tab:analysis_baseline} (\cref{sec:main-cost}), these records connect task-level objectives to concrete physical operations.

%% file: sec10_cases.tex
\section{Extended Case Studies}
\label{app:cases}

\begin{figure}[!htbp]
  \centering
  \begin{subfigure}{0.90\linewidth}
    \centering
    \includegraphics[width=\linewidth]{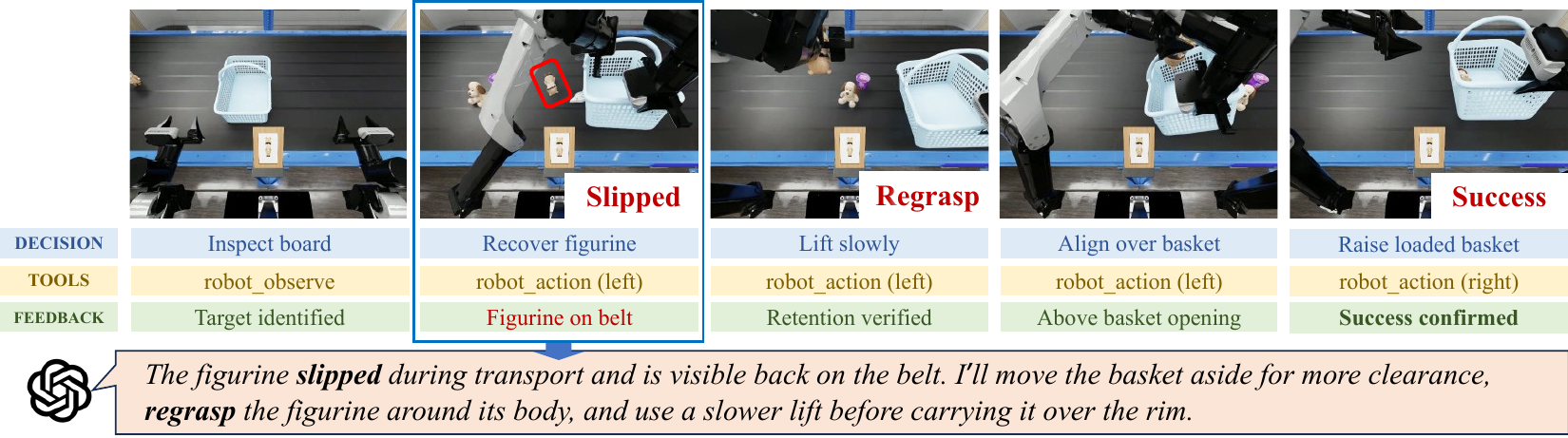}
    \caption{Conveyor picking (\code{pick_from_conveyor_by_image}).}
    \label{fig:regrasp_conveyor}
  \end{subfigure}
  \par\smallskip
  \begin{subfigure}{0.90\linewidth}
    \centering
    \includegraphics[width=\linewidth]{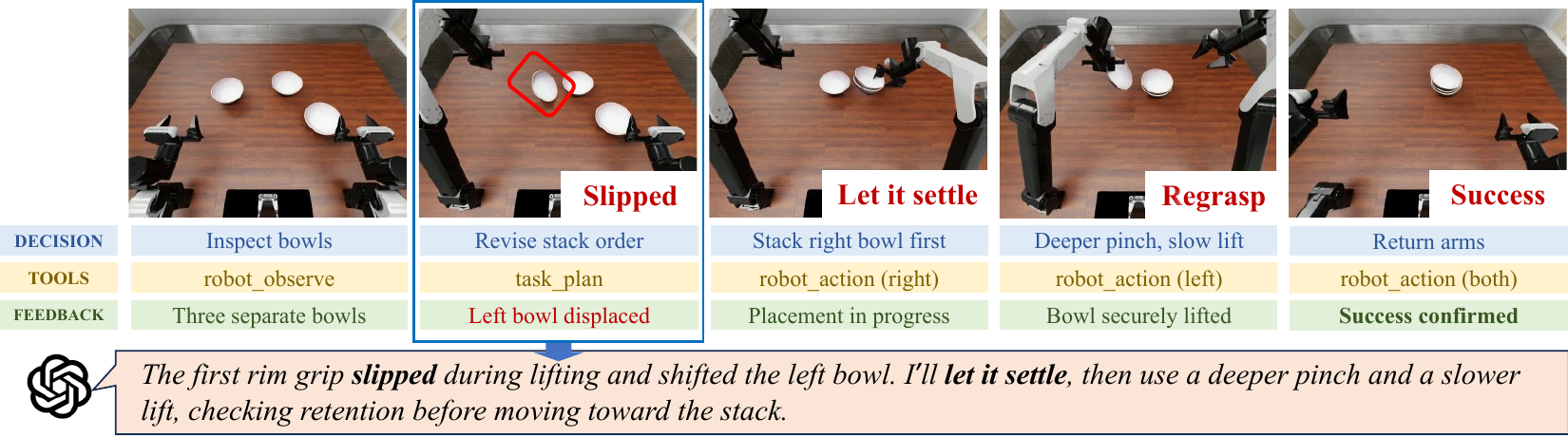}
    \caption{Bowl stacking (\code{stack_bowls}).}
    \label{fig:regrasp_bowls}
  \end{subfigure}
  \caption{Additional RoboDojo episodes recovering from grasp failure and slippage. The model changes the grasped region for the figurine and uses a deeper pinch and slower lift for the bowl.}
  \label{fig:case_regrasp}
\end{figure}

\begin{figure}[!htbp]
  \centering
  \includegraphics[width=0.90\linewidth]{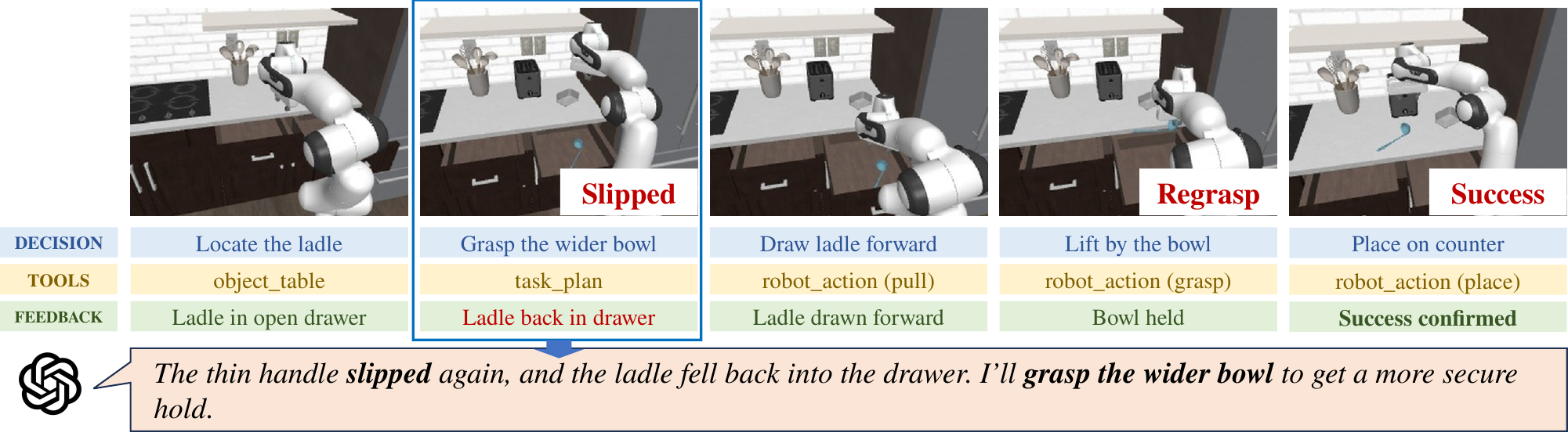}
  \caption{Recovery from grasp failure and slippage in RoboCasa365 drawer-to-counter transfer (\code{PickPlaceDrawerToCounter}). After grasps on the thin ladle handle slip during lifting and base motion, the model pulls the ladle forward and switches contact to the wider bowl.}
  \label{fig:regrasp_ladle}
\end{figure}

\begin{figure}[!htbp]
  \centering
  \begin{subfigure}{0.90\linewidth}
    \centering
    \includegraphics[width=\linewidth]{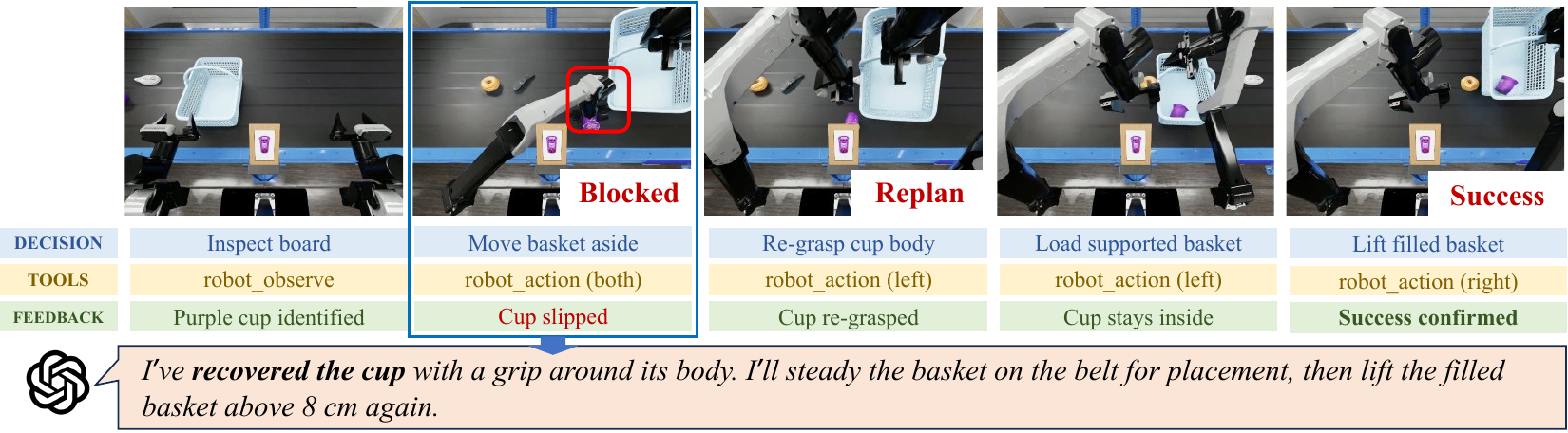}
    \caption{Conveyor picking (\code{pick_from_conveyor_by_image}).}
    \label{fig:replanning_conveyor}
  \end{subfigure}
  \par\smallskip
  \begin{subfigure}{0.90\linewidth}
    \centering
    \includegraphics[width=\linewidth]{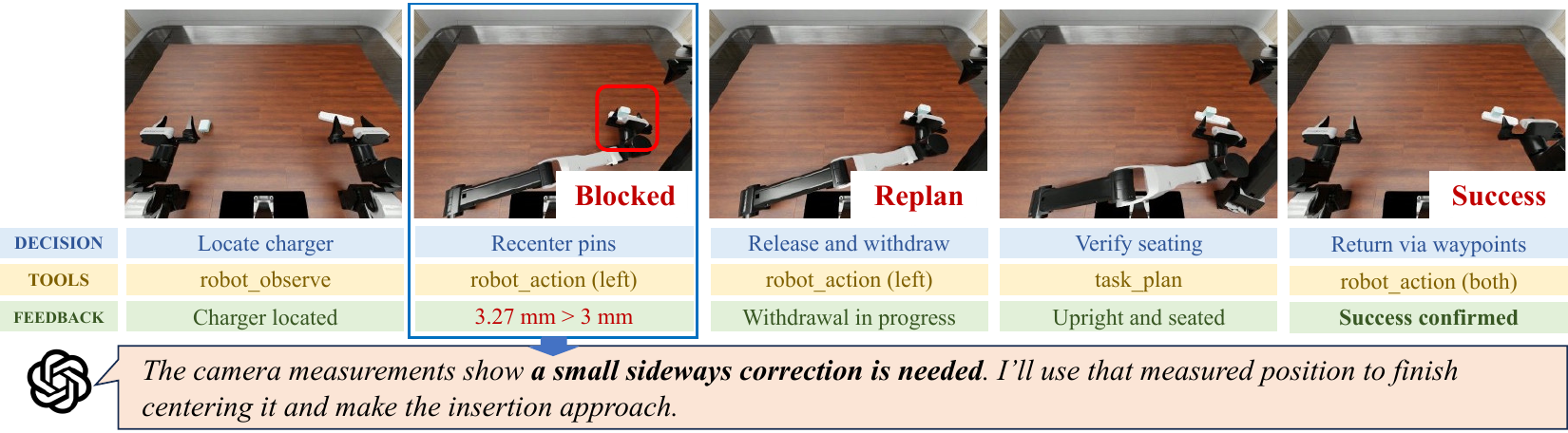}
    \caption{Charger insertion (\code{plug_in_charger}).}
    \label{fig:replanning_charger}
  \end{subfigure}
  \caption{Additional RoboDojo episodes replanning after contact and execution disruptions. The model supports the basket during cup placement and corrects the charger's alignment before insertion.}
  \label{fig:case_replanning}
\end{figure}

\begin{figure}[!htbp]
  \centering
  \begin{subfigure}{0.90\linewidth}
    \centering
    \includegraphics[width=\linewidth]{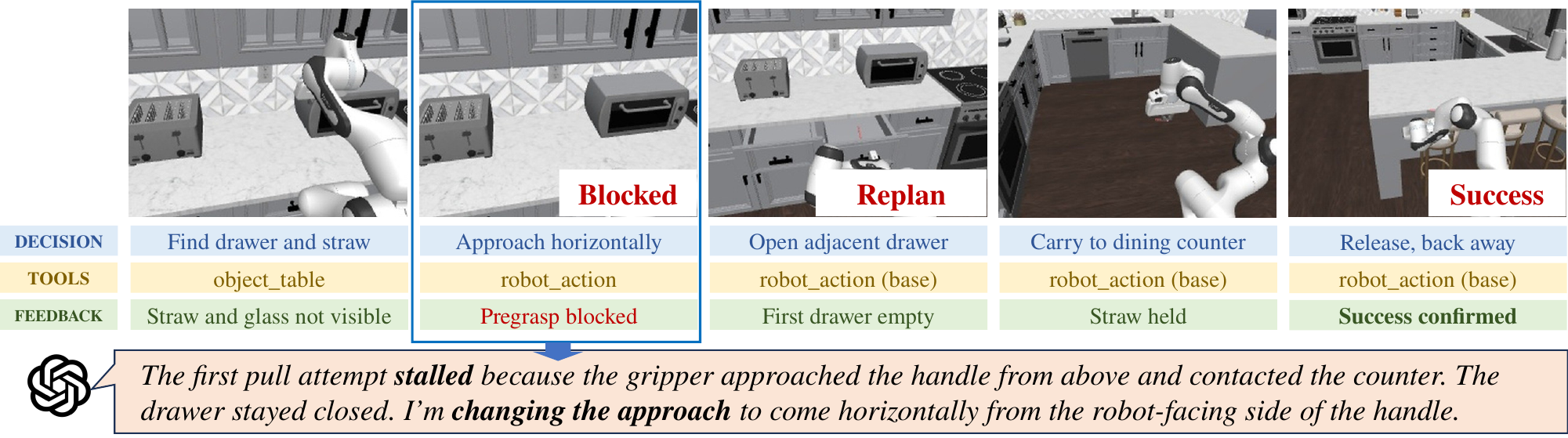}
    \caption{Straw delivery (\code{DeliverStraw}).}
    \label{fig:replanning_straw}
  \end{subfigure}
  \par\smallskip
  \begin{subfigure}{0.90\linewidth}
    \centering
    \includegraphics[width=\linewidth]{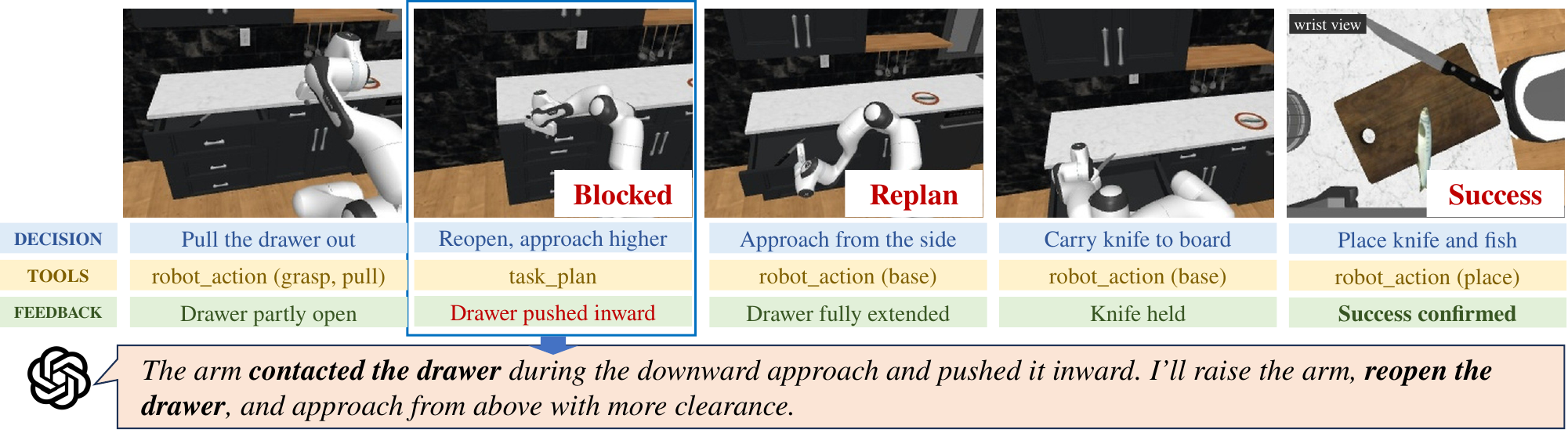}
    \caption{Cutting-station setup (\code{SetUpCuttingStation}).}
    \label{fig:replanning_cutting}
  \end{subfigure}
  \caption{Additional RoboCasa365 episodes replanning after contact and execution disruptions. The model changes the approach to the drawer handle and then switches to the adjacent drawer when the first is empty (\code{DeliverStraw}), and reopens a drawer pushed inward by arm contact before approaching from the side (\code{SetUpCuttingStation}).}
  \label{fig:case_replanning_robocasa}
\end{figure}

This appendix provides the extended RoboDojo and RoboCasa365 case analyses accompanying \cref{sec:cases}: additional recovery episodes after slippage (\cref{fig:case_regrasp,fig:regrasp_ladle}), replanning episodes after contact and execution disruptions (\cref{fig:case_replanning,fig:case_replanning_robocasa}), and detailed step-by-step budget accounting for the failure episodes (\cref{tab:case_failure_budget,tab:case_failure_budget_robocasa}).

\paragraph{Additional slippage recovery episodes.}
Alongside sequence reproduction and lettuce washing (\cref{fig:regrasp_main} in \cref{sec:case_regrasp}), \cref{fig:case_regrasp,fig:regrasp_ladle} show how the model separates acquisition, retention during transport, and post-release support on conveyor picking, bowl stacking, and drawer-to-counter transfer.

\textbf{Conveyor picking: separating acquisition, retention, and destination access.}
The model must match a figurine to the board image, place it inside the basket, and lift the basket by more than 8\,cm.
The moving conveyor first invalidates a basket pickup location, prompting a switch to right-arm interception.
For the figurine, an approach contacts the hat before closure; a later empty closure leaves the body exposed.
A torso grasp then acquires the object, but the figurine slips during transport (\cref{fig:regrasp_conveyor}).
The model consequently changes both the grasp and the transfer conditions: it moves the basket aside, re-grasps across the torso, and verifies retention during a slower lift.

Once the figurine is retained, the remaining obstacle is the destination.
Moving the basket underneath it causes contact with the image board and tilts the opening.
The model keeps the figurine held, moves the basket upward and away from the board, and then completes the release.
A later check against the initial image revises the height judgment because the earlier estimate followed a rim on a tilted basket.
The loaded basket is raised farther before the evaluator signals success.
These successive revisions address distinct predicates (target acquired, target retained, opening accessible, and required basket elevation), so progress on one predicate can be preserved while another is corrected.

\textbf{Bowl stacking: changing the assembly order after a failed lift.}
The initial plan uses the rear middle bowl as the base, adds the left bowl, and then adds the right bowl.
The first rim grip slips and displaces the left bowl.
The revised plan lets that bowl settle while the robot approaches the right bowl, using a deeper pinch and slower lift (\cref{fig:regrasp_bowls}).
After the right bowl is nested into the base and remains supported after release, the model re-localizes the displaced left bowl and adds it last.
Thus recovery changes the assembly order as well as the grasp.

This reordering is possible because the task constrains the final stack rather than the order in which bowls enter it.
It also gives the displaced bowl time to settle before its next localization.
The plan checks co-motion after lifting and upright support after release and withdrawal; the final return supplies another observation of whether the stack persists without gripper support.
The evaluator confirms successful stacking.
The distinction from sequence reproduction is consequential: the bowl order can change to simplify execution, whereas the observed object order must remain fixed.

\textbf{Drawer-to-counter transfer: shifting contact from a thin handle to the wider bowl.}
The robot must take a ladle from an open drawer and place it on the counter, where the single task predicate tests only whether the ladle rests on the counter and remains false until final placement.
An initial grasp on the thin handle slips during the lift, and after a lower re-grasp on a thicker section of the handle, the ladle slips again during mobile-base motion and falls back into the drawer (\cref{fig:regrasp_ladle}).
Both losses are reported by the execution receipts rather than the task predicate, and after the second slip the object table no longer finds the ladle in the wrist-camera view.

The model revises its task plan to grasp the ladle's wider bowl instead, though roughly ten attempts are blocked by the counter overhang or miss inside the drawer while it pulls the drawer out farther.
It then uses the handle to draw the ladle forward within the drawer and grasps the bowl once clear of the overhang.
Transferring the ladle by its bowl and releasing it on the counter turns the placement predicate true, and the evaluator confirms task success.
This revision corresponds to the acquisition condition at the start of \cref{sec:case_regrasp}: when a thin handle fails to sustain lift and base transport, recovery changes the contact geometry to a wider region of the object.

\paragraph{Additional contact and replanning episodes.}
Alongside object packing and burner shutoff (\cref{fig:replanning_main} in \cref{sec:case_replanning}), \cref{fig:case_replanning,fig:case_replanning_robocasa} follow the model from a measured discrepancy to a procedure revision across conveyor picking, charger insertion, straw delivery, and cutting-station setup.

\textbf{Conveyor picking: using support to decouple placement from basket motion.}
The cup initially remains held while the other arm raises the basket.
During approach, the basket swings and the cup shifts in the left grip; a wrist-orientation change then loses the cup.
After recovering it with a body grasp, the model preserves that wrist orientation and changes the basket procedure (\cref{fig:replanning_conveyor}).
The revised plan lowers the basket onto the belt, loads the cup, and then lifts the filled basket above the required height.

Supporting the basket changes the manipulation problem.
Insertion can now be organized around a supported opening, while lifting becomes a separate operation on the loaded container.
When a side adjustment loses the handle grip, the plan retains the cup grasp, clears the right arm for placement, and postpones handle acquisition until the cup has been released.
The model then verifies that the cup stays inside after left-arm withdrawal, re-grasps the handle from above, and lifts the basket to complete the task.
This sequence preserves the final object-in-basket and elevation requirements while removing the need to maintain a suspended basket throughout insertion.

\textbf{Charger insertion: revising only the part of the procedure contradicted by evidence.}
The initial plan allows a table-supported handover if the left arm cannot reach the socket.
A preview establishes a direct route, so the model keeps the charger in the left hand and moves the right arm outward to clear that route.
Near the socket, the \texttt{clear\_pins} clearance lift returns
\texttt{tracking\_error}: the position error is 3.27~mm, predominantly
vertical, exceeding the stage's tight 3~mm tolerance.
The charger remains held, and camera measurements identify a small lateral alignment correction (\cref{fig:replanning_charger}).
The next action uses the returned position to center the pins and complete the insertion, followed by release and a check that the charger stays seated.

The later return-path failure calls for a different revision.
Because seating remains visually supported, the model leaves the inserted charger undisturbed and splits the right-arm return into shorter movements through clear intermediate poses.
The evaluator then reports success.
This separates a local alignment correction from a route-planning problem: one changes the target relative to the socket, while the other changes how the arm reaches its home pose.
Neither requires repeating the already verified insertion.

\textbf{Straw delivery: matching the scope of revision to the invalidated assumption.}
The task requires taking a straw from a drawer to a glass cup on the dining counter; initially, the object table locates the drawer handles while neither the straw nor the cup is visible.
Repeated attempts to open the first drawer return execution receipts for a blocked or stalled approach (\cref{fig:replanning_straw}), as approaching the handle from above brings the gripper against the counter.
The model switches to a horizontal approach from the robot-facing side of the handle, backs the mobile base away for clearance, and writes that direction into the grasp stage before pulling the drawer open with the base.

Inspecting the opened drawer reveals a different discrepancy when the object table finds no straw inside.
The task plan records that this drawer is empty while the adjacent front drawer remains unchecked, and the robot opens that second drawer using the same horizontal handle approach and base pull.
After grasping the straw and driving around the counter end, it rights the glass after a missed first insertion, places the straw inside to satisfy the straw-in-cup predicate, and backs away as the evaluator confirms success.
As in charger insertion, the two disruptions call for different scopes of revision: a blocked grasp changes the approach route, whereas an empty drawer changes the target container without altering the delivery objective.

\textbf{Cutting-station setup: restoring a displaced container before revising the approach.}
The task requires taking a knife from a drawer and a fish from a plate and placing both onto a cutting board.
When the arm reaches downward for the knife, the execution receipt reports a blocked approach, and the camera view shows that arm contact has pushed the partly open drawer inward (\cref{fig:replanning_cutting}).
The model initially plans to raise the arm, reopen the drawer, and approach from above with more clearance.

Across roughly ten subsequent attempts, however, passing the arm over the drawer front repeatedly pushes the drawer inward and leaves grasps blocked or empty.
The model then pulls the drawer fully open, withdraws the gripper clear of the handle before turning the wrist to approach from the side, and lowers its grasp slightly to place the finger pads around the knife handle.
Delivering the knife satisfies the knife-on-board predicate, and after the fish is re-grasped following a slip during base motion, the evaluator signals task success during the final placement onto the board.
As in object packing, the central revision treats the displaced drawer as state that must be restored and paired with a non-colliding approach route, rather than targeting coordinates measured before contact.

\paragraph{Detailed budget accounting for failure episodes.}
\Cref{tab:case_failure_budget,tab:case_failure_budget_robocasa} give the step-by-step native motion and model-request budget accounting for the failed \code{stack_bowls} and \code{HeatKebabSandwich} episodes shown in \cref{fig:case_failure_main} (\cref{sec:case_failure}).

\noindent\begin{minipage}{\linewidth}
  \captionsetup{type=table}
  \centering
  \footnotesize
  \setlength{\tabcolsep}{5pt}
  \renewcommand{\arraystretch}{0.92}
  \caption{Progress and remaining motion budget in the failed RoboDojo stacking episode (\code{stack_bowls}, \cref{fig:case_failure}).
  Step counts come from execution accounting; state descriptions condense the observation-linked model records.}
  \label{tab:case_failure_budget}
  \begin{tabular}{@{}rrp{0.69\linewidth}@{}}
    \toprule
    Steps used & Steps left & State guiding the next decision \\
    \midrule
    242 & 558 & Two bowls remain nested after withdrawal; add the third. \\
    440 & 360 & The third bowl remains tilted after seating attempts; re-grasp and level it. \\
    564 & 236 & The nested stack was lifted and slipped; restore it using table support. \\
    691 & 109 & The stack has rolled onto its side again during the return; attempt a compact supported correction. \\
    756 & 44 & The rim pinch has not retained the stack; change contact to the raised base. \\
    800 & 0 & Native step limit reached; task failure, Score 15/100. \\
    \bottomrule
  \end{tabular}
\end{minipage}

\noindent\begin{minipage}{\linewidth}
  \captionsetup{type=table}
  \centering
  \footnotesize
  \setlength{\tabcolsep}{5pt}
  \renewcommand{\arraystretch}{0.92}
  \caption{Progress and remaining model budget in the failed RoboCasa365 episode (\code{HeatKebabSandwich}, \cref{fig:case_failure_robocasa}).
  Request counts are model responses out of the 100 allowed per episode; state descriptions condense the recorded model messages.}
  \label{tab:case_failure_budget_robocasa}
  \begin{tabular}{@{}rrp{0.69\linewidth}@{}}
    \toprule
    Requests used & Requests left & State guiding the next decision \\
    \midrule
    14 & 86 & The baguette has slipped onto the counter beside the open door; re-grasp it with an explicit approach direction. \\
    41 & 59 & The baguette is inside the oven but below the wire rack, so its rack-contact check is false; move the rack to expose the bread. \\
    51 & 49 & The kebab rests on the rack and its check passes; slide the loaded rack inward and recover the baguette from below. \\
    63 & 37 & The rack has not moved because the grip on its thin front rail was not secure; grasp the rail from a raised angle. \\
    86 & 14 & The rack has slid outward with the kebab, leaving the baguette behind it; lift the bread onto the rack. \\
    97 & 3 & The model reports the task unfinished: kebab on the rack, baguette below it, door open; the harness asks it to continue. \\
    100 & 0 & Model-request limit reached; task failure. \\
    \bottomrule
  \end{tabular}
\end{minipage}

\textbf{Coupled structures and local completion versus cumulative progress.}
In \code{stack_bowls} (\cref{tab:case_failure_budget}), the plan at step 440 intends to lift and level the tilted top bowl, but the grasp lifts the entire nested stack, changing the recovery target from seating one bowl to restoring upright support for all three bowls.
Although subsequent tipping maneuvers, supported pushes, and contact changes receive \code{completed} motion receipts, the structure rolls again upon arm withdrawal as the remaining budget drops from 360 steps to 44.
Similarly, in \code{HeatKebabSandwich} (\cref{tab:case_failure_budget_robocasa}), after the baguette lands below the toaster wire rack at request 41, exposing the bread requires pulling the thin front rail of the rack, which repeatedly slips between requests 63 and 86; although placing the kebab at request 50 and sliding the rack outward at request 86 succeed locally, only 14 requests remain and the 100-request budget expires before the baguette is lifted onto the rack.

%% file: references.bib
@misc{openai2026gpt6,
  author = {{OpenAI}},
  title = {{GPT-6 Astra System Card}},
  year = {2026},
  month = sep,
  howpublished = {System card},
  url = {https://deploymentsafety.openai.com/gpt-6-astra}
}

@misc{anthropic2026opus55,
  author = {{Anthropic}},
  title = {{System Card: Claude Opus 5.5}},
  year = {2026},
  month = sep,
  howpublished = {System card},
  url = {https://anthropic.com/claude-opus-5-5-system-card}
}

@misc{gemini2025robotics15,
  title = {{Gemini Robotics 1.5: Pushing the Frontier of Generalist Robots with Advanced Embodied Reasoning, Thinking, and Motion Transfer}},
  author = {{Gemini Robotics Team} and Abbas Abdolmaleki and Saminda Abeyruwan and Joshua Ainslie and Jean-Baptiste Alayrac and Montserrat Gonzalez Arenas and Ashwin Balakrishna and Nathan Batchelor and Alex Bewley and Jeff Bingham and Michael Bloesch and Konstantinos Bousmalis and Philemon Brakel and Anthony Brohan and Thomas Buschmann and Arunkumar Byravan and Serkan Cabi and Ken Caluwaerts and Federico Casarini and Christine Chan and Oscar Chang and London Chappellet-Volpini and Jose Enrique Chen and Xi Chen and Hao-Tien Lewis Chiang and Krzysztof Choromanski and Adrian Collister and David B. D'Ambrosio and Sudeep Dasari and Todor Davchev and Meet Kirankumar Dave and Coline Devin and Norman Di Palo and Tianli Ding and Carl Doersch and Adil Dostmohamed and Yilun Du and Debidatta Dwibedi and Sathish Thoppay Egambaram and Michael Elabd and Tom Erez and Xiaolin Fang and Claudio Fantacci and Cody Fong and Erik Frey and Chuyuan Fu and Ruiqi Gao and Marissa Giustina and Keerthana Gopalakrishnan and Laura Graesser and Oliver Groth and Agrim Gupta and Roland Hafner and Steven Hansen and Leonard Hasenclever and Sam Haves and Nicolas Heess and Brandon Hernaez and Alex Hofer and Jasmine Hsu and Lu Huang and Sandy H. Huang and Atil Iscen and Mithun George Jacob and Deepali Jain and Sally Jesmonth and Abhishek Jindal and Ryan Julian and Dmitry Kalashnikov and M. Emre Karagozler and Stefani Karp and Matija Kecman and J. Chase Kew and Donnie Kim and Frank Kim and Junkyung Kim and Thomas Kipf and Sean Kirmani and Ksenia Konyushkova and Li Yang Ku and Yuheng Kuang and Thomas Lampe and Antoine Laurens and Tuan Anh Le and Isabel Leal and Alex X. Lee and Tsang-Wei Edward Lee and Guy Lever and Jacky Liang and Li-Heng Lin and Fangchen Liu and Shangbang Long and Caden Lu and Sharath Maddineni and Anirudha Majumdar and Kevis-Kokitsi Maninis and Andrew Marmon and Sergio Martinez and Assaf Hurwitz Michaely and Niko Milonopoulos and Joss Moore and Robert Moreno and Michael Neunert and Francesco Nori and Joy Ortiz and Kenneth Oslund and Carolina Parada and Emilio Parisotto and Amaris Paryag and Acorn Pooley and Thomas Power and Alessio Quaglino and Haroon Qureshi and Rajkumar Vasudeva Raju and Helen Ran and Dushyant Rao and Kanishka Rao and Isaac Reid and David Rendleman and Krista Reymann and Miguel Rivas and Francesco Romano and Yulia Rubanova and Peter Pastor Sampedro and Pannag R Sanketi and Dhruv Shah and Mohit Sharma and Kathryn Shea and Mohit Shridhar and Charles Shu and Vikas Sindhwani and Sumeet Singh and Radu Soricut and Rachel Sterneck and Ian Storz and Razvan Surdulescu and Jie Tan and Jonathan Tompson and Saran Tunyasuvunakool and Jake Varley and Grace Vesom and Giulia Vezzani and Maria Bauza Villalonga and Oriol Vinyals and Ren{\'e} Wagner and Ayzaan Wahid and Stefan Welker and Paul Wohlhart and Chengda Wu and Markus Wulfmeier and Fei Xia and Ted Xiao and Annie Xie and Jinyu Xie and Peng Xu and Sichun Xu and Ying Xu and Zhuo Xu and Jimmy Yan and Sherry Yang and Skye Yang and Yuxiang Yang and Hiu Hong Yu and Wenhao Yu and Wentao Yuan and Yuan Yuan and Jingwei Zhang and Tingnan Zhang and Zhiyuan Zhang and Allan Zhou and Guangyao Zhou and Yuxiang Zhou},
  year = {2025},
  eprint = {2510.03342},
  archiveprefix = {arXiv},
  primaryclass = {cs.RO},
  url = {https://arxiv.org/abs/2510.03342},
  howpublished = {arXiv preprint arXiv:2510.03342},
}

@misc{rynnbrain2026,
  title = {{RynnBrain 1.1: Towards More Capable and Generalizable Embodied Foundation Model}},
  author = {Kehan Li and Bohan Hou and Minghao Zhu and Tianyi Zhang and Zesen Cheng and Zhikai Wang and Sicong Leng and Xin Li and Xiao Lin and Biying Yao and Minghua Zeng and Jiangpin Liu and Ronghao Dang and Jiayan Guo and Siteng Huang and Haoyu Zhao and Heng Ping and Yaxi Zhao and Tong Zhao and Kexiang Wang and Tong Lu and Shengke Xue and Jiahao Tang and Yulei Wang and Zejing Wang and Jianwei Gao and Shijian Lu and Chengju Liu and Jianfei Yang and Mingxiu Chen and Deli Zhao},
  year = {2026},
  eprint = {2607.17977},
  archiveprefix = {arXiv},
  primaryclass = {cs.RO},
  url = {https://arxiv.org/abs/2607.17977},
  howpublished = {arXiv preprint arXiv:2607.17977},
}

@misc{hyembodied2026,
  title = {{HY-Embodied-0.5: Embodied Foundation Models for Real-World Agents}},
  author = {{Tencent Robotics X} and {HY Vision Team} and Xumin Yu and Zuyan Liu and Ziyi Wang and He Zhang and Yongming Rao and Fangfu Liu and Yani Zhang and Ruowen Zhao and Oran Wang and Yves Liang and Haitao Lin and Minghui Wang and Yubo Dong and Kevin Cheng and Bolin Ni and Rui Huang and Han Hu and Zhengyou Zhang and Linus and Shunyu Yao},
  year = {2026},
  eprint = {2604.07430},
  archiveprefix = {arXiv},
  primaryclass = {cs.CV},
  url = {https://arxiv.org/abs/2604.07430},
  howpublished = {arXiv preprint arXiv:2604.07430},
}

@misc{embodiedr2026,
  title = {{Embodied-R1.5: Evolving Physical Intelligence via Embodied Foundation Models}},
  author = {Yifu Yuan and Yaoting Huang and Xianze Yao and Yutong Li and Shuoheng Zhang and Linqi Han and Pengyi Li and Jiangeng Sun and Wenting Jia and Zhao Zhang and Yuhao Liu and Ruihao Liao and Yucheng Hu and Qiyu Wu and Yuxiao Li and Zibin Dong and Fei Ni and Yan Zheng and Shuyang Gu and Yi Ma and Hongyao Tang and Han Hu and Jianye Hao},
  year = {2026},
  eprint = {2606.11324},
  archiveprefix = {arXiv},
  primaryclass = {cs.RO},
  url = {https://arxiv.org/abs/2606.11324},
  howpublished = {arXiv preprint arXiv:2606.11324},
}

@misc{physbrain2026,
  title = {{PhysBrain 1.5: From Vision-Language Models to Physical Foundation Models}},
  author = {{DeepCybo Team} and Yu Bin and Haipeng Cao and Zheng Chang and Kai Chen and Youning Chen and Kailin Deng and Yichao Du and Xiaotong Fu and Haoyang Ge and Yunlong Guo and Chenliu Hao and Jiyan He and Xuguo He and Yakun Hou and Kai Hu and Cong Huang and Tuopusen Huang and Yu Huang and Hong Li and Peize Li and Shijie Lian and Xiaopeng Lin and Yun Lin and Haibao Liu and Haochen Liu and Qiuzhi Liu and Shengcai Liu and Zhiqiang Liu and Tao Luo and Peng Ren and Shuo Ren and Chaoyi Ruan and Zhaolong Shen and Yukun Shi and Qiyuan Su and Yuxuan Tian and Yining Wang and Changti Wu and Hao Wu and Xueyin Xu and Ruoqi Yang and Zhaoyang Yang and Hang Yuan and Zhaoyang Zeng and Hanwen Zhang and Ruimeng Zhang and Yao Zhang and Yibo Zhang and Yuxiang Zhang and Zhirui Zhang and Ziyi Zhang and Zubin Zheng and Zishen Zhuang},
  year = {2026},
  eprint = {2609.14973},
  archiveprefix = {arXiv},
  primaryclass = {cs.CV},
  url = {https://arxiv.org/abs/2609.14973},
  howpublished = {arXiv preprint arXiv:2609.14973},
}

@misc{unexpected2026,
  title = {{An Unexpected Robot Policy: Early Evaluations of GPT-6 Astra on RoboDojo and Beyond}},
  author = {Wenbo Zhang and Kaixuan Wang and Yutao Ouyang and Xiaoyu Huang and Liyang Li and Kailun Su and Weiyang Jin and Wenhao Chai and Haotian Liang and Zhiyang Dou and Yue Chen and Tianxing Chen},
  year = {2026},
  eprint = {2609.24170},
  archiveprefix = {arXiv},
  primaryclass = {cs.CV},
  url = {https://arxiv.org/abs/2609.24170},
  howpublished = {arXiv preprint arXiv:2609.24170},
}

@misc{exploring2026,
  title = {{Systematically Exploring the Capabilities of GPT-6 Astra as Embodied Policies}},
  author = {{Galbot Team} and Xuchuan Chen and Xiaoqian Cheng and Yu Deng and Lihe Ding and Shaocong Dong and Xiangjun Gao and Haozhe Jia and Zekai Li and Zhoujian Li and Yunrui Lian and Sikai Liang and Chenghuai Lin and Dairu Liu and Jiahang Liu and Qingtao Liu and Yuxuan Ma and Zekun Qi and Jiayi Su and He Wang and Ruochen Xu and Tianyu Xu and Xudong Xu and Zhe Xu and Mi Yan and Siming Yan and Li Yi and Ruixi Yu and Jinlu Zhang and Yintianrun Zhang and Zhikai Zhang and Zhizheng Zhang and Yixin Zheng and Weiyi Zhu},
  year = {2026},
  eprint = {2609.38537},
  archiveprefix = {arXiv},
  primaryclass = {cs.RO},
  url = {https://arxiv.org/abs/2609.38537},
  howpublished = {arXiv preprint arXiv:2609.38537},
}

@misc{gptpolicy2026,
  title = {{In-Context Robot Learning with VLM Agents}},
  author = {Dongzhou Cheng and Taoran Yi and Ye Fang and Xingwu Zhang and Fan Feng and Yixuan Li and Gengxiong Zhuang and Rongze Wang and Shuai Yang and Wei Song and Weizhi Xue and Minyan Wu and Jie Gui and Jiaqi Wang and Tong Wu},
  year = {2026},
  eprint = {2609.19138},
  archiveprefix = {arXiv},
  primaryclass = {cs.CV},
  url = {https://arxiv.org/abs/2609.19138},
  howpublished = {arXiv preprint arXiv:2609.19138},
}

@misc{roboicl2026,
  title = {{RoboICL: Embodied In-Context Learning with GPT-6 Astra}},
  author = {Fangcheng Liu and Yeqing Shen and Anda Cheng and Weishi Mi and Chao Tang and Chenyuan Liu and Yushun Xiang and Tingguang Li and Yong-Lu Li and Yehui Tang},
  year = {2026},
  eprint = {2609.34261},
  archiveprefix = {arXiv},
  primaryclass = {cs.RO},
  url = {https://arxiv.org/abs/2609.34261},
  howpublished = {arXiv preprint arXiv:2609.34261},
}

@misc{agp2026,
  title = {{Agent as Policy for Robotic Manipulation}},
  author = {Mengzhao Jia and Yang Lin and Xixin Zhang and Zhihan Zhang and Xiaobai Liu and Meng Jiang},
  year = {2026},
  eprint = {2609.12541},
  archiveprefix = {arXiv},
  primaryclass = {cs.CL},
  url = {https://arxiv.org/abs/2609.12541},
  howpublished = {arXiv preprint arXiv:2609.12541},
}

@misc{roboharness2026,
  title = {{RoboHarness: A Simple Harness Outperforms VLA and World Action Models}},
  author = {Bince Qu and Wei Chen and Bo Zhang},
  year = {2026},
  eprint = {2610.14018},
  archiveprefix = {arXiv},
  primaryclass = {cs.RO},
  url = {https://arxiv.org/abs/2610.14018},
  howpublished = {arXiv preprint arXiv:2610.14018},
}

@misc{capx2026,
  title = {{CaP-X: A Framework for Benchmarking and Improving Coding Agents for Robot Manipulation}},
  author = {Letian Fu and Justin Yu and Karim El-Refai and Ethan Kou and Haoru Xue and Huang Huang and Wenli Xiao and Guanzhi Wang and Dantong Niu and Fei-Fei Li and Guanya Shi and Jiajun Wu and Shankar Sastry and Yuke Zhu and Ken Goldberg and Linxi "Jim" Fan},
  year = {2026},
  eprint = {2603.22435},
  archiveprefix = {arXiv},
  primaryclass = {cs.RO},
  url = {https://arxiv.org/abs/2603.22435},
  howpublished = {arXiv preprint arXiv:2603.22435},
}

@misc{fastharness2026,
  title = {{FastHarness: Experience-Driven Review Scheduling for Vision-Language-Action Policies}},
  author = {Tao Lin and Yuxin Du and Haoxiao Wang and Zile Zhong and Bo Zhao},
  year = {2026},
  eprint = {2610.31427},
  archiveprefix = {arXiv},
  primaryclass = {cs.RO},
  url = {https://arxiv.org/abs/2610.31427},
  howpublished = {arXiv preprint arXiv:2610.31427},
}

@misc{harnesspai2026,
  title = {{HarnessPAI: An Evolving Harness for Physical AI}},
  author = {Xin Wang and Wenhao Wu and Menghao Zhang and Zhi Wang and Kun Shao and Jian Luan and Yang Li and Qing Li and Shangding Gu and Huichi Zhou and Shuqing Shi and Fei Ni and Shuo Lu and Weicheng Meng and Kang Li and Jin Wu and Kang Zhao and Shangmin Guo and Gen Li and Yongqiang Tang and Zhizhong Zhang and Yuan Xie and Heng Qu},
  year = {2026},
  eprint = {2609.29166},
  archiveprefix = {arXiv},
  primaryclass = {cs.RO},
  url = {https://arxiv.org/abs/2609.29166},
  howpublished = {arXiv preprint arXiv:2609.29166},
}

@misc{dynaharness2026,
  title = {{DynaHarness: A Dynamic Physical Harness for Self-Evolving Robot Agents}},
  author = {Haoyuan Deng and Jiebin Liu and Tengxiao Zhang and Langning Yan and Hongye Cao and Ziwei Wang},
  year = {2026},
  eprint = {2609.40306},
  archiveprefix = {arXiv},
  primaryclass = {cs.RO},
  url = {https://arxiv.org/abs/2609.40306},
  howpublished = {arXiv preprint arXiv:2609.40306},
}

@misc{shaper2026,
  title = {{Self-Evolving Embodied Agents via Skill-Harness Evolution}},
  author = {Peidong Wang and Zhiming Ma and Ying Chang and Xufang Luo and Yiqun Zhang and Zihan Wang and Xiaocui Yang and Shi Feng and Yuqing Yang and Dongsheng Li},
  year = {2026},
  eprint = {2608.11350},
  archiveprefix = {arXiv},
  primaryclass = {cs.CL},
  url = {https://arxiv.org/abs/2608.11350},
  howpublished = {arXiv preprint arXiv:2608.11350},
}

@misc{embodiedrsi2026,
  title = {{EmbodiedRSI: Active Continual Robot Learning Through Hypothesis-Guided Co-Evolution}},
  author = {Python Song and Zhixuan Liang and Kelsey Fu and Mengdi Wang and Junfeng Yang and Shilong Liu},
  year = {2026},
  eprint = {2610.10498},
  archiveprefix = {arXiv},
  primaryclass = {cs.AI},
  url = {https://arxiv.org/abs/2610.10498},
  howpublished = {arXiv preprint arXiv:2610.10498},
}

@misc{scaleselection2026,
  title = {{Scale and Selection: What Makes Automatic Harness Evolution Work for Visual-Interface Robot Agents}},
  author = {Zhijie Wei and Ferris Tan and Jinghui Wang and Hao Zhou},
  year = {2026},
  eprint = {2609.39304},
  archiveprefix = {arXiv},
  primaryclass = {cs.RO},
  url = {https://arxiv.org/abs/2609.39304},
  howpublished = {arXiv preprint arXiv:2609.39304},
}

@misc{enpire2026,
  title = {{ENPIRE: Agentic Robot Policy Self-Improvement in the Real World}},
  author = {Wenli Xiao and Jia Xie and Tonghe Zhang and Haotian Lin and Letian "Max" Fu and Haoru Xue and Jalen Lu and Yi Yang and Cunxi Dai and Zi Wang and Jimmy Wu and Guanzhi Wang and S. Shankar Sastry and Ken Goldberg and Linxi "Jim" Fan and Yuke Zhu and Guanya Shi},
  year = {2026},
  eprint = {2606.19980},
  archiveprefix = {arXiv},
  primaryclass = {cs.AI},
  url = {https://arxiv.org/abs/2606.19980},
  howpublished = {arXiv preprint arXiv:2606.19980},
}

@misc{aspire2026,
  title = {{ASPIRE: Agentic /Skills Discovery for Robotics}},
  author = {Runyu Lu and Yubo Wu and Ethan Kou and Haotian Lin and Letian Fu and Wenli Xiao and Ajay Mandlekar and Yinzhen Xu and Guanya Shi and Ken Goldberg and Ang Chen and Mosharaf Chowdhury and Jan Kautz and Yuke Zhu and Linxi "Jim" Fan and Guanzhi Wang},
  year = {2026},
  eprint = {2607.00272},
  archiveprefix = {arXiv},
  primaryclass = {cs.RO},
  url = {https://arxiv.org/abs/2607.00272},
  howpublished = {arXiv preprint arXiv:2607.00272},
}

@misc{gap2026,
  title = {{GaP: A Graph-as-Policy Multi-Agent Self-Learning Harness For Variational Automation Tasks}},
  author = {Kaiyuan Chen and Shuangyu Xie and Letian Fu and Justin Yu and William Pacini and Sandeep Bajamahal and Hudson Kim and Jaimyn Drake and Daehwa Kim and Haoru Xue and Jonathan Francis and Christian Juette and Peter Schaldenbrand and Muhammet Yunus Seker and Ruwan Wickramarachchi and Uksang Yoo and Guanzhi Wang and Adithyavairavan Murali and Balakumar Sundaralingam and S. Shankar Sastry and Yuke Zhu and Linxi "Jim" Fan and Spencer Huang and Ken Goldberg},
  year = {2026},
  eprint = {2607.05369},
  archiveprefix = {arXiv},
  primaryclass = {cs.RO},
  url = {https://arxiv.org/abs/2607.05369},
  howpublished = {arXiv preprint arXiv:2607.05369},
}

@misc{rats2026,
  title = {{Playful Agentic Robot Learning}},
  author = {Junyi Zhang and Jiaxin Ge and Hanjun Yoo and Letian Fu and Zihan Yang and Yaowei Liu and Raj Saravanan and Shaofeng Yin and Justin Yu and Dantong Niu and Zirui Wang and Roei Herzig and Ken Goldberg and Yutong Bai and David M. Chan and Ion Stoica and Angjoo Kanazawa and Jiahui Lei and Haiwen Feng and Trevor Darrell},
  year = {2026},
  eprint = {2606.19419},
  archiveprefix = {arXiv},
  primaryclass = {cs.RO},
  url = {https://arxiv.org/abs/2606.19419},
  howpublished = {arXiv preprint arXiv:2606.19419},
}

@misc{teachgrow2026,
  title = {{Teach and Grow: An Agent-Centered Architecture for General Robot Learning}},
  author = {Chang Nie and Zhe Liu and Hesheng Wang},
  year = {2026},
  eprint = {2608.17209},
  archiveprefix = {arXiv},
  primaryclass = {cs.RO},
  url = {https://arxiv.org/abs/2608.17209},
  howpublished = {arXiv preprint arXiv:2608.17209},
}

@misc{real2gym2026,
  title = {{Real2Gym: Building Gyms from Videos, Bringing Skills to Robots}},
  author = {Kerui Ren and Yingxiang Xu and Kaiwen Song and Linning Xu and Bo Dai and Mulin Yu and Tao Lu},
  year = {2026},
  eprint = {2609.37089},
  archiveprefix = {arXiv},
  primaryclass = {cs.CV},
  url = {https://arxiv.org/abs/2609.37089},
  howpublished = {arXiv preprint arXiv:2609.37089},
}

@misc{pi072026,
  title = {{${\pi}_{0.7}$: a Steerable Generalist Robotic Foundation Model with Emergent Capabilities}},
  author = {{Physical Intelligence} and Bo Ai and Ali Amin and Raichelle Aniceto and Ashwin Balakrishna and Greg Balke and Kevin Black and George Bokinsky and Shihao Cao and Thomas Charbonnier and Vedant Choudhary and Foster Collins and Ken Conley and Grace Connors and James Darpinian and Karan Dhabalia and Maitrayee Dhaka and Jared DiCarlo and Danny Driess and Michael Equi and Adnan Esmail and Yunhao Fang and Chelsea Finn and Catherine Glossop and Thomas Godden and Ivan Goryachev and Lachlan Groom and Haroun Habeeb and Hunter Hancock and Karol Hausman and Gashon Hussein and Victor Hwang and Brian Ichter and Connor Jacobsen and Szymon Jakubczak and Rowan Jen and Tim Jones and Gregg Kammerer and Ben Katz and Liyiming Ke and Mairbek Khadikov and Chandra Kuchi and Marinda Lamb and Devin LeBlanc and Brendon LeCount and Sergey Levine and Xinyu Li and Adrian Li-Bell and Vladislav Lialin and Zhonglin Liang and Wallace Lim and Yao Lu and Enyu Luo and Vishnu Mano and Nandan Marwaha and Aikys Mongush and Liam Murphy and Suraj Nair and Tyler Patterson and Karl Pertsch and Allen Z. Ren and Gavin Schelske and Charvi Sharma and Baifeng Shi and Lucy Xiaoyang Shi and Laura Smith and Jost Tobias Springenberg and Kyle Stachowicz and Will Stoeckle and Jiaming Tang and Jimmy Tanner and Shalom Tekeste and Marcel Torne and Kyle Vedder and Quan Vuong and Anna Walling and Haohuan Wang and Jason Wang and XuDong Wang and Chris Whalen and Samuel Whitmore and Blake Williams and Charles Xu and Sukwon Yoo and Lili Yu and Wuming Zhang and Zhuoyang Zhang and Ury Zhilinsky},
  year = {2026},
  eprint = {2604.15483},
  archiveprefix = {arXiv},
  primaryclass = {cs.LG},
  url = {https://arxiv.org/abs/2604.15483},
  howpublished = {arXiv preprint arXiv:2604.15483},
}

@misc{lingbotvla2026,
  title = {{From Foundation to Application: Improving VLA Models in Practice}},
  author = {Wei Wu and Fangjing Wang and Fan Lu and He Sun and Shi Liu and Yunnan Wang and Yibin Yan and Yong Wang and Shuailei Ma and Xinyang Wang and Yibin Liu and Shuai Yang and Tianxiang Zhou and Kejia Zhang and Lei Zhou and Cheng Su and Nan Xue and Bin Tan and Han Zhang and Youchao Zhang and Fei Liao and Xing Zhu and Yujun Shen and Kecheng Zheng},
  year = {2026},
  eprint = {2607.06403},
  archiveprefix = {arXiv},
  primaryclass = {cs.RO},
  url = {https://arxiv.org/abs/2607.06403},
  howpublished = {arXiv preprint arXiv:2607.06403},
}

@misc{xiaomi2026,
  title = {{Xiaomi-Robotics-1: Scaling Vision-Language-Action Models with over 100K Hours of Real-World Trajectories}},
  author = {{Xiaomi Robotics Team} and Jun Guo and Piaopiao Jin and Jason Li and Peiyan Li and Yingyan Li and Futeng Liu and Wanli Peng and Optimus Qin and Yifei Su and Nan Sun and Qiao Sun and Runze Suo and Heyun Wang and Yunhong Wang and Rujie Wu and Caoyu Xia and Lina Zhang and Jack Zhao and Guoliang Chen and Wenlong Chen and Xinze He and Bin Li and Qing Li and Zhuorong Li and Heng Qu and Wenxuan Song and Diyun Xiang and Yifan Xie and Peiran Xu and Hangjun Ye and Wen Ye and Han Zhao and Quanyun Zhou},
  year = {2026},
  eprint = {2607.15330},
  archiveprefix = {arXiv},
  primaryclass = {cs.RO},
  url = {https://arxiv.org/abs/2607.15330},
  howpublished = {arXiv preprint arXiv:2607.15330},
}

@misc{qwenrobot2026,
  title = {{Qwen-RobotManip Technical Report: Alignment Unlocks Scale for Robotic Manipulation Foundation Models}},
  author = {Haoqi Yuan and Zhixuan Liang and Anzhe Chen and Ye Wang and Haoyang Li and Pei Lin and Yiyang Huang and Zixing Lei and Tong Zhang and Jiazhao Zhang and Jie Zhang and Jingyang Fan and Gengze Zhou and Qihang Peng and Chenxu Lv and Xiaoyue Chen and An Yang and Fei Huang and Junyang Lin and Dayiheng Liu and Jingren Zhou and Chenfei Wu and Xiong-Hui Chen},
  year = {2026},
  eprint = {2606.17846},
  archiveprefix = {arXiv},
  primaryclass = {cs.RO},
  url = {https://arxiv.org/abs/2606.17846},
  howpublished = {arXiv preprint arXiv:2606.17846},
}

@misc{pi062025,
  title = {{$\pi^{*}_{0.6}$: a VLA That Learns From Experience}},
  author = {{Physical Intelligence} and Ali Amin and Raichelle Aniceto and Ashwin Balakrishna and Kevin Black and Ken Conley and Grace Connors and James Darpinian and Karan Dhabalia and Jared DiCarlo and Danny Driess and Michael Equi and Adnan Esmail and Yunhao Fang and Chelsea Finn and Catherine Glossop and Thomas Godden and Ivan Goryachev and Lachy Groom and Hunter Hancock and Karol Hausman and Gashon Hussein and Brian Ichter and Szymon Jakubczak and Rowan Jen and Tim Jones and Ben Katz and Liyiming Ke and Chandra Kuchi and Marinda Lamb and Devin LeBlanc and Sergey Levine and Adrian Li-Bell and Yao Lu and Vishnu Mano and Mohith Mothukuri and Suraj Nair and Karl Pertsch and Allen Z. Ren and Charvi Sharma and Lucy Xiaoyang Shi and Laura Smith and Jost Tobias Springenberg and Kyle Stachowicz and Will Stoeckle and Alex Swerdlow and James Tanner and Marcel Torne and Quan Vuong and Anna Walling and Haohuan Wang and Blake Williams and Sukwon Yoo and Lili Yu and Ury Zhilinsky and Zhiyuan Zhou},
  year = {2025},
  eprint = {2511.14759},
  archiveprefix = {arXiv},
  primaryclass = {cs.LG},
  url = {https://arxiv.org/abs/2511.14759},
  howpublished = {arXiv preprint arXiv:2511.14759},
}

@misc{g052026,
  title = {{G0.5: One Autoregressive Stream for Robot Reasoning and Action}},
  author = {Yicheng Liu and Zibin Dong and Baijun Ye and Tianyuan Yuan and Tao Jiang and Anqi Yang and Shicheng Cao and Haonan Liu and Yue Sun and Zihan Guo and Xiao Liu and Dong Ke and Changxun Pan and Chenru Wu and Tailai Cheng and Xiaoshu Ren and Xinlei Zhang and Jianning Cui and Zijie Zhao and Haoyu Zhang and Kaiming Xu and Haodong Yang and Bowen Zhang and Jiahui Niu and Shaoting Zhu and Shiduo Zhang and Hang Zhao},
  year = {2026},
  eprint = {2608.11739},
  archiveprefix = {arXiv},
  primaryclass = {cs.RO},
  url = {https://arxiv.org/abs/2608.11739},
  howpublished = {arXiv preprint arXiv:2608.11739},
}

@misc{dreamzero2026,
  title = {{World Action Models are Zero-shot Policies}},
  author = {Seonghyeon Ye and Yunhao Ge and Kaiyuan Zheng and Shenyuan Gao and Sihyun Yu and George Kurian and Suneel Indupuru and You Liang Tan and Chuning Zhu and Jiannan Xiang and Ayaan Malik and Kyungmin Lee and William Liang and Nadun Ranawaka and Jiasheng Gu and Yinzhen Xu and Guanzhi Wang and Fengyuan Hu and Avnish Narayan and Johan Bjorck and Jing Wang and Gwanghyun Kim and Dantong Niu and Ruijie Zheng and Yuqi Xie and Jimmy Wu and Qi Wang and Ryan Julian and Danfei Xu and Yilun Du and Yevgen Chebotar and Scott Reed and Jan Kautz and Yuke Zhu and Linxi "Jim" Fan and Joel Jang},
  year = {2026},
  eprint = {2602.15922},
  archiveprefix = {arXiv},
  primaryclass = {cs.RO},
  url = {https://arxiv.org/abs/2602.15922},
  howpublished = {arXiv preprint arXiv:2602.15922},
}

@misc{cosmospolicy2026,
  title = {{Cosmos Policy: Fine-Tuning Video Models for Visuomotor Control and Planning}},
  author = {Moo Jin Kim and Yihuai Gao and Tsung-Yi Lin and Yen-Chen Lin and Yunhao Ge and Grace Lam and Percy Liang and Shuran Song and Ming-Yu Liu and Chelsea Finn and Jinwei Gu},
  year = {2026},
  eprint = {2601.16163},
  archiveprefix = {arXiv},
  primaryclass = {cs.AI},
  url = {https://arxiv.org/abs/2601.16163},
  howpublished = {arXiv preprint arXiv:2601.16163},
}

@misc{vpp22026,
  title = {{Video Prediction Policy 2: Predict Better, Act Better}},
  author = {Yanjiang Guo and Haodong Yan and Zhide Zhong and Zhongru Zhang and Qingyuan Yang and Qingzhou Lu and Xiaoyu Chen and Yen-Jen Wang and Shuying Deng and Chenghan Yang and Puzhen Yuan and Chenxin Liu and Tun Ban and Xiang Zhu and Yichen Liu and Kun Feng and Haoang Li and Jianyu Chen},
  year = {2026},
  eprint = {2610.10270},
  archiveprefix = {arXiv},
  primaryclass = {cs.CV},
  url = {https://arxiv.org/abs/2610.10270},
  howpublished = {arXiv preprint arXiv:2610.10270},
}

@misc{openwam2026,
  title = {{OpenWAM: An Open, Modular Exploration Towards Systematic World-Action Model Pretraining}},
  author = {Yuran Wang and Siqiao Huang and Mingleyang Li and Chenhao Zhang and Jiaqi Liang and Weiyang Jin and Yue Chen and Xuemin Chi and Donghao Zhou and Qize Yu and Yu-Kai Wang and Yuhan Rui and Shenzhe Yao and Zhen Yuan and Zhenhao Shen and Kefei Zhu and Zijie Zhu and Ning Gao and Xiaowei Chi and Guanqi He and Shanghang Zhang and Hao Dong and Lin Shao and Hang Zhao},
  year = {2026},
  eprint = {2609.07398},
  archiveprefix = {arXiv},
  primaryclass = {cs.RO},
  url = {https://arxiv.org/abs/2609.07398},
  howpublished = {arXiv preprint arXiv:2609.07398},
}

@misc{beingh2026,
  title = {{Being-H0.7: A Latent World-Action Model from Egocentric Videos}},
  author = {Hao Luo and Wanpeng Zhang and Yicheng Feng and Sipeng Zheng and Haiweng Xu and Chaoyi Xu and Ziheng Xi and Yuhui Fu and Zongqing Lu},
  year = {2026},
  eprint = {2605.00078},
  archiveprefix = {arXiv},
  primaryclass = {cs.RO},
  url = {https://arxiv.org/abs/2605.00078},
  howpublished = {arXiv preprint arXiv:2605.00078},
}

@misc{jepawam2026,
  title = {{JEPA-WAM: Learning Vision-Language-Action Policies with Joint-Embedding World Modeling}},
  author = {Yihan Lin and Jiawei He and Shifeng Bao and Chen Zhao and Yang Li and Xiaobo Wang and Yan Wang and Cheng Chi and Jing Zhang},
  year = {2026},
  eprint = {2608.09381},
  archiveprefix = {arXiv},
  primaryclass = {cs.RO},
  url = {https://arxiv.org/abs/2608.09381},
  howpublished = {arXiv preprint arXiv:2608.09381},
}

@misc{fastwam2026,
  title = {{Fast-WAM: Do World Action Models Need Test-time Future Imagination?}},
  author = {Tianyuan Yuan and Zibin Dong and Yicheng Liu and Hang Zhao},
  year = {2026},
  eprint = {2603.16666},
  archiveprefix = {arXiv},
  primaryclass = {cs.CV},
  url = {https://arxiv.org/abs/2603.16666},
  howpublished = {arXiv preprint arXiv:2603.16666},
}

@misc{longwam2026,
  title = {{Long-WAM: Scaling the Context of World-Action Models}},
  author = {Wei Huang and Bohan Zhang and Chenzhi Liu and Isabella Liu and Shuai Yang and Weian Mao and Luozhou Wang and Yicheng Xiao and Weifeng Lin and Qixin Hu and Bryan Chu and Sifei Liu and Linxi Fan and Xiaojuan Qi and Song Han and Yukang Chen},
  year = {2026},
  eprint = {2610.10528},
  archiveprefix = {arXiv},
  primaryclass = {cs.RO},
  url = {https://arxiv.org/abs/2610.10528},
  howpublished = {arXiv preprint arXiv:2610.10528},
}

@misc{hexashell2026,
  title = {{HexaShell: A Thin Harness for Exploring the Limits of VLM Agents as Robot Coding Agents}},
  author = {Hongcheng Gao and Alex Zhou and Jay Zhu and Zelin Zheng and Shijia Ge and Xiping Li and Di Wu and Yazhe Wang and Jianshu Zeng and Xuan Shangguan and Yongli Wang and Yexing Wan and Lingyu He and Zhiqi Jia and Sihang Wu and Xiao He},
  year = {2026},
  eprint = {2610.23479},
  archiveprefix = {arXiv},
  primaryclass = {cs.RO},
  url = {https://arxiv.org/abs/2610.23479},
  howpublished = {arXiv preprint arXiv:2610.23479},
}

@misc{embodiedevo2026,
  title = {{EmbodiedEvo: Self-Evolving Coding Agent For Physical Intelligence}},
  author = {Wenyi Li and Yizhe Chi and Youjie Zheng and Chunyu Liu and Xiaoqiu Wang and Mingju Gao and Zheng Jiang and Deyao Hong and Calvin Xiao and Kaisen Yang and Qinhuai Na},
  year = {2026},
  eprint = {2610.18867},
  archiveprefix = {arXiv},
  primaryclass = {cs.RO},
  url = {https://arxiv.org/abs/2610.18867},
  howpublished = {arXiv preprint arXiv:2610.18867},
}

@misc{internvla152026,
  title = {{InternVLA-A1.5: Unifying Understanding, Latent Foresight, and Action for Compositional Generalization}},
  author = {Haoxiang Ma and Junhao Cai and Xiaoxu Xu and Hao Li and Yuyin Yang and Yang Tian and Jiafei Cao and Hongrui Zhu and Zherui Qiu and Zhaxizhuoma and Yuqiang Yang and Jiaqi Peng and Xueyuan Wei and Yangkun Zhu and Jiahao Jiang and Xing Gao and Hanqing Wang and Feng Yuan and Kailin Li and Xueyue Zhu and Tai Wang and Yan Ding and Jiangmiao Pang and Jia Zeng and Jingjing Zhang and Bowen Zhou and Yao Mu and Chunhua Shen and Weinan Zhang},
  year = {2026},
  eprint = {2607.04988},
  archiveprefix = {arXiv},
  primaryclass = {cs.RO},
  url = {https://arxiv.org/abs/2607.04988},
  howpublished = {arXiv preprint arXiv:2607.04988},
}

@misc{xvla2025,
  title = {{X-VLA: Soft-Prompted Transformer as Scalable Cross-Embodiment Vision-Language-Action Model}},
  author = {Jinliang Zheng and Jianxiong Li and Zhihao Wang and Dongxiu Liu and Xirui Kang and Yuchun Feng and Yinan Zheng and Jiayin Zou and Yilun Chen and Jia Zeng and Ya-Qin Zhang and Jiangmiao Pang and Jingjing Liu and Tai Wang and Xianyuan Zhan},
  year = {2025},
  eprint = {2510.10274},
  archiveprefix = {arXiv},
  primaryclass = {cs.RO},
  url = {https://arxiv.org/abs/2510.10274},
  howpublished = {arXiv preprint arXiv:2510.10274},
}

@misc{pi052025,
  title = {{$\pi_{0.5}$: a Vision-Language-Action Model with Open-World Generalization}},
  author = {{Physical Intelligence} and Kevin Black and Noah Brown and James Darpinian and Karan Dhabalia and Danny Driess and Adnan Esmail and Michael Equi and Chelsea Finn and Niccolo Fusai and Manuel Y. Galliker and Dibya Ghosh and Lachy Groom and Karol Hausman and Brian Ichter and Szymon Jakubczak and Tim Jones and Liyiming Ke and Devin LeBlanc and Sergey Levine and Adrian Li-Bell and Mohith Mothukuri and Suraj Nair and Karl Pertsch and Allen Z. Ren and Lucy Xiaoyang Shi and Laura Smith and Jost Tobias Springenberg and Kyle Stachowicz and James Tanner and Quan Vuong and Homer Walke and Anna Walling and Haohuan Wang and Lili Yu and Ury Zhilinsky},
  year = {2025},
  eprint = {2504.16054},
  archiveprefix = {arXiv},
  primaryclass = {cs.LG},
  url = {https://arxiv.org/abs/2504.16054},
  howpublished = {arXiv preprint arXiv:2504.16054},
}

@misc{robodojo2026,
  title = {{RoboDojo: A Unified Sim-and-Real Benchmark for Comprehensive Evaluation of Generalist Robot Manipulation Policies}},
  author = {Tianxing Chen and Yue Chen and Zixuan Li and Junyuan Tang and Kailun Su and Haoran Lu and Weijie Wan and Baijun Chen and Songling Liu and Haowen Yan and Honghao Su and Zhiyang Dou and Kaixuan Wang and Dandan Zhang and Yunze Liu and Yan Qin and Qiwei Liang and Qiwei Wu and Zijian Lin and Wenwei Lin and Yuran Wang and Minghua He and Tianshu Wu and Ruihai Wu and Jingquan Zhou and Kai-Chong Lei and Haibao Yu and Yuanfeng Ji and Weiyang Jin and Guanyu Lin and Xiaofan Li and Qi Xiong and Renjing Xu and Zhongyu Li and Wenhao Chai and Enze Xie and Ziwei Wang and Yao Mu and Hao Dong and Wojciech Matusik and Mingyu Ding and Wenbo Ding and Ping Luo and Masayoshi Tomizuka},
  year = {2026},
  eprint = {2607.04434},
  archiveprefix = {arXiv},
  primaryclass = {cs.RO},
  url = {https://arxiv.org/abs/2607.04434},
  howpublished = {arXiv preprint arXiv:2607.04434},
}

@misc{internwdelta2026,
  title = {{InternW0-$\Delta$: A World Action Model Bridging Predictive Dynamics and Actions with 20K+ Hours of Open Data}},
  author = {Xingyu Miao and Zizun Li and Baole Fang and Kaiwen Song and Tenghui Wang and Hanxue Zhang and Yating Wang and Xudong Li and Yuping He and Xueyuan Wei and Chao Gao and Xijie Yang and Yingxiang Xu and Kerui Ren and Wenqi Guo and Jianjun Zhou and Xinzhe Wang and Weiguang Zhao and Ni Yang and Zetao Cai and Yufei Xue and Hengjie Li and Zeyu He and Yuanzhen Zhou and Rong Fu and Jianyang Zhang and Siwei Cui and Fuxian Huang and Yunsong Zhou and Xing Gao and Yifei Yao and Qiaojun Yu and Kailin Li and Ming Zhou and Mu Huang and Xinyue Li and Wenze Cui and Bingqi Jiang and Xueyue Zhu and Junting Dong and Haoyu Guo and Tao Lu and Mulin Yu and Bowen Zhou and Bin Zhao and Tianfan Xue and Weinan Zhang and Chunhua Shen},
  year = {2026},
  eprint = {2609.31394},
  archiveprefix = {arXiv},
  primaryclass = {cs.RO},
  url = {https://arxiv.org/abs/2609.31394},
  howpublished = {arXiv preprint arXiv:2609.31394},
}

@misc{xwam2026,
  title = {{Unified 4D World Action Modeling from Video Priors with Asynchronous Denoising}},
  author = {Jun Guo and Qiwei Li and Peiyan Li and Zilong Chen and Nan Sun and Yifei Su and Heyun Wang and Yuan Zhang and Xinghang Li and Huaping Liu},
  year = {2026},
  eprint = {2604.26694},
  archiveprefix = {arXiv},
  primaryclass = {cs.RO},
  url = {https://arxiv.org/abs/2604.26694},
  howpublished = {arXiv preprint arXiv:2604.26694},
}

@misc{physicalrsi2026,
  author = {{MMLab \& PhysicalRSI contributors}},
  title = {{Physical RSI 1.0: Recursive Self-Harness for Scaling Embodied Skills}},
  year = {2026},
  howpublished = {Project website and open-source software},
  url = {https://mmlab.hk/research/PhysicalRSI},
  note = {Preview release}
}

@techreport{r2si2026,
  author = {{Manifold AI Team}},
  title = {{$R^2$SI: Recursive Robot Self-Improvement via a Bi-Level Agentic System with Agent--Model Co-Evolution}},
  institution = {Manifold AI},
  year = {2026},
  month = oct,
  url = {https://r2si.worldscape.io/report.pdf}
}

@misc{liang2023code,
  title = {{Code as Policies: Language Model Programs for Embodied Control}},
  author = {Jacky Liang and Wenlong Huang and Fei Xia and Peng Xu and Karol Hausman and Brian Ichter and Pete Florence and Andy Zeng},
  year = {2023},
  eprint = {2209.07753},
  archiveprefix = {arXiv},
  primaryclass = {cs.RO},
  url = {https://arxiv.org/abs/2209.07753},
  howpublished = {arXiv preprint arXiv:2209.07753},
}

@misc{gptgear2026,
  title = {{Large Multimodal Models for Direct Robot Control: GPT-6 as a Beginning}},
  author = {Youwei Zhou and Bohan Hou and Yankai Fu and Yaoxu Lyu and Yifan Ye and Shilin Shan and Chen Guo and Jiaming Liu and Ran He and Caifeng Shan and Shanghang Zhang and Jianfei Yang and Zhihe Lu},
  year = {2026},
  eprint = {2610.28259},
  archiveprefix = {arXiv},
  primaryclass = {cs.RO},
  url = {https://arxiv.org/abs/2610.28259},
  howpublished = {arXiv preprint arXiv:2610.28259},
}

@misc{sensenova2026embodied,
  title = {{Has GPT-6-Astra Achieved Embodied Intelligence?}},
  author = {{SenseNova} and {Joint Research Team} and Zhongang Cai and Ziang Cao and Tailai Chen and Han Chong Chua and Linjun Dai and Hanming Deng and Hokin Deng and Haiwen Diao and Chenjing Ding and Weichen Fan and Xiao Fu and Zerui Gong and Chenyang Gu and Zexin He and Zhi Hou and Bo Li and Qinyue Li and Quanzhou Li and Xiaowan Li and Zhichao Liao and Dahua Lin and Zilong Lin and Bo Liu and Chunxiao Liu and Pengfei Liu and Tianqi Liu and Ziwei Liu and Chen Change Loy and Lewei Lu and Jiapeng Luo and Yuexiao Ma and Liang Pan and Xingang Pan and Oscar Qian and Ziyi Shang and Yuyao Shao and Dawei Su and Dacheng Tao and Qingyi Tao and Quan Wang and Ruisi Wang and Xiaogang Wang and Ziyi Wang and Beichen Wen and Donghao Wu and Jiaxu Wu and Silei Wu and Xiaoming Wu and Zhonghua Wu and Hongcan Xiao and Weiye Xu and Atsushi Yamashita and Lei Yang and Zhe Yang and Baiqiao Yin and Wanqi Yin and Hanfei Zhao and Tianbo Zhong and Long Zhuo and Jing Zuo},
  year = {2026},
  eprint = {2610.22471},
  archiveprefix = {arXiv},
  primaryclass = {cs.RO},
  url = {https://arxiv.org/abs/2610.22471},
  howpublished = {arXiv preprint arXiv:2610.22471},
}

@misc{pyrua2026,
  title = {{Fewer Tokens, Better Action: GPT-6 Astra Robot Agents with 14\% Higher Success Rate but 65\% Fewer Tokens}},
  author = {Ruiyang Si and Jianxin Bi and Shunyu Yang and Rui Ni and Wenbo Huang and Qiang Wang and Shulong Jiang and Yintianrun Zhang and Feixiang Ruan and Kewei Zhang and Duomin Wang and Xiuyu Li and Haiwen Feng and Zhen Dong and Daquan Zhou},
  year = {2026},
  eprint = {2610.01939},
  archiveprefix = {arXiv},
  primaryclass = {cs.CV},
  url = {https://arxiv.org/abs/2610.01939},
  howpublished = {arXiv preprint arXiv:2610.01939},
}

@misc{sear2026,
  title = {{SEAR: Self-Evolving Agentic Robotics}},
  author = {Xinyi Wang and Bangzheng Li and Hanyu Wang and Lin Zhang and Andrea Yaoyun Cui and Zhenggang Tang and Shyamal Buch and Dejia Xu},
  year = {2026},
  eprint = {2610.18792},
  archiveprefix = {arXiv},
  primaryclass = {cs.RO},
  url = {https://arxiv.org/abs/2610.18792},
  howpublished = {arXiv preprint arXiv:2610.18792},
}

@misc{wang2026vggt,
  title = {{VGGT-$\Omega$}},
  author = {Jianyuan Wang and Minghao Chen and Shangzhan Zhang and Nikita Karaev and Johannes Sch{\"o}nberger and Patrick Labatut and Piotr Bojanowski and David Novotny and Andrea Vedaldi and Christian Rupprecht},
  year = {2026},
  eprint = {2605.15195},
  archiveprefix = {arXiv},
  primaryclass = {cs.CV},
  url = {https://arxiv.org/abs/2605.15195},
  howpublished = {arXiv preprint arXiv:2605.15195},
}

@misc{carion2026sam,
  title = {{SAM 3: Segment Anything with Concepts}},
  author = {Nicolas Carion and Laura Gustafson and Yuan-Ting Hu and Shoubhik Debnath and Ronghang Hu and Didac Suris and Chaitanya Ryali and Kalyan Vasudev Alwala and Haitham Khedr and Andrew Huang and Jie Lei and Tengyu Ma and Baishan Guo and Arpit Kalla and Markus Marks and Joseph Greer and Meng Wang and Peize Sun and Roman R{\"a}dle and Triantafyllos Afouras and Effrosyni Mavroudi and Katherine Xu and Tsung-Han Wu and Yu Zhou and Liliane Momeni and Rishi Hazra and Shuangrui Ding and Sagar Vaze and Francois Porcher and Feng Li and Siyuan Li and Aishwarya Kamath and Ho Kei Cheng and Piotr Doll{\'a}r and Nikhila Ravi and Kate Saenko and Pengchuan Zhang and Christoph Feichtenhofer},
  year = {2026},
  eprint = {2511.16719},
  archiveprefix = {arXiv},
  primaryclass = {cs.CV},
  url = {https://arxiv.org/abs/2511.16719},
  howpublished = {arXiv preprint arXiv:2511.16719},
}

@misc{robocasa3652026,
  title = {{RoboCasa365: A Large-Scale Simulation Framework for Training and Benchmarking Generalist Robots}},
  author = {Soroush Nasiriany and Sepehr Nasiriany and Abhiram Maddukuri and Yuke Zhu},
  year = {2026},
  eprint = {2603.04356},
  archiveprefix = {arXiv},
  primaryclass = {cs.RO},
  url = {https://arxiv.org/abs/2603.04356},
  howpublished = {arXiv preprint arXiv:2603.04356},
}

@misc{harnessvla2026,
  title = {{Harness VLA: Steering Frozen VLAs into Reliable Manipulation Primitives via Memory-Guided Agents}},
  author = {Yixian Zhang and Huanming Zhang and Feng Gao and Xiao Li and Zhihao Liu and Yi Nie and Chunyang Zhu and Jiaxing Qiu and Yuchen Yan and Jiyuan Liu and Wenhao Tang and Jiaji Rao and Zhengru Fang and Changxu Wei and Yu Wang and Wenbo Ding and Chao Yu},
  year = {2026},
  eprint = {2607.08448},
  archiveprefix = {arXiv},
  primaryclass = {cs.RO},
  url = {https://arxiv.org/abs/2607.08448},
  howpublished = {arXiv preprint arXiv:2607.08448},
}

@misc{rldx2026,
  title = {{RLDX-1 Technical Report}},
  author = {Dongyoung Kim and Huiwon Jang and Myungkyu Koo and Suhyeok Jang and Taeyoung Kim and Beomjun Kim and Byungjun Yoon and Changsung Jang and Daewon Choi and Dongsu Han and Donguk Lee and Heeseung Kwon and Hojin Jeon and Jaehyun Kang and Jaekyoung Bae and Jihyuk Lee and Jimin Lee and John Won and Joonwoo Ahn and Junhyeong Park and Junyoung Sung and Kyungmin Lee and Minseong Han and Minsung Yoon and Sejune Joo and Seonil Son and Seungcheol Park and Seunggeun Cho and Seungjun Moon and Seungku Kim and Yonghoon Dong and Yongjin Cho and Youngchan Kim and Chang Hwan Kim and Dohyeon Kim and Heecheol Kim and Heewon Lee and Hensen Ahn and Hyungkyu Ryu and Hyunsoo Choi and Hyunsoo Shin and Jaeheon Jung and Jaewoo Kim and Jinwook Kim and Joochul Chang and Joonsoo Kim and Junghun Park and Jungwoo Park and Junho Cho and Junhyeok Park and Junwon Lee and Kangwook Lee and Kwanghoon Kim and Kyoungwhan Choe and Manoj Bhadu and Nayoung Oh and Sangjun Kim and Sangwoo Kim and Seunghoon Shim and Seunghyun Kim and Seungjun Lee and Seungyup Ka and Sungryol Yang and Wook Jung and Yashu Shukla and Yeonjae Lee and Yeonwoo Bae and Jinwoo Shin},
  year = {2026},
  eprint = {2605.03269},
  archiveprefix = {arXiv},
  primaryclass = {cs.RO},
  url = {https://arxiv.org/abs/2605.03269},
  howpublished = {arXiv preprint arXiv:2605.03269},
}

@misc{robocasaboard2026,
  title={{RoboCasa365} Leaderboard},
  author={{RoboCasa Team}},
  howpublished={\url{https://robocasa.ai/leaderboard.html}},
  note={Accessed 2026-10-30},
  year={2026}
}

@misc{duan2026lastai,
  title = {{The Last AI Built by Humans: Toward Genuine Recursive Self-Improvement}},
  author = {Yi Duan and Ying Liu and Zirui Tang and Haodong Chen and Jun Zhou and Yumou Liu and Bangrui Xu and Yukai Wu and Sidi Chen and Yuhan Zhou and Haoyu Wang and Xiaoyou Yu and Shaokun Han and Xuzhou Zhu and Le Zhou and Bolin Lu and Wei Zhou and Jiachen Liu and Nuozhou Fang and Jiaxin Tian and Ruoyu Chen and Yuxuan Li and Kai Zuo and Kaiyan Zhang and Qianyu Yang and Zijie Wang and Jiantao Qiu and Conghui He and Guoliang Li and Bowen Zhou and Zhiyuan Liu and Zhoufutu Wen and Jihua Kang and Xuanhe Zhou and Fan Wu},
  year = {2026},
  eprint = {2609.11873},
  archiveprefix = {arXiv},
  primaryclass = {cs.LG},
  url = {https://arxiv.org/abs/2609.11873},
  howpublished = {arXiv preprint arXiv:2609.11873},
}

@misc{zhou2026physicalrsi,
  title = {{Who Improves the Robot? Toward Physical Recursive Self-Improvement}},
  author = {Hai Zhou and Fan Yang and Zeyi Chen and Lejun Zhang and Chenyang Fu and Jianping Li and Jiayi Zhu and Xianyang Liu and Yingxuan Yang and Lili Meng and Lu Wang and Xinyang Han and Jiankai Sun and Weirui Ye and Guanchu Wang and Wanhua Li and Ming Zhou and Boyi Li and Ming Jin and Kun Shao and Muhao Chen and Guang Chen and Lei Ma and Joey Tianyi Zhou and Costas Spanos and Xia Hu and Mac Schwager and Weinan Zhang and Qian Feng and Shangding Gu},
  year = {2026},
  eprint = {2610.18842},
  archiveprefix = {arXiv},
  primaryclass = {cs.AI},
  url = {https://arxiv.org/abs/2610.18842},
  howpublished = {arXiv preprint arXiv:2610.18842},
}

@misc{robodojoboard2026,
  author = {{RoboDojo}},
  title = {{RoboDojo Leaderboard}},
  year = {2026},
  url = {https://robodojo-benchmark.com/leaderboard},
  note = {Accessed 2026-10-31}
}
